\pdfoutput=1
\documentclass[11pt]{article}
\usepackage[T1]{fontenc}
\usepackage[utf8]{inputenc}
\usepackage{lmodern}
\usepackage[margin=1in]{geometry}
\usepackage{microtype}
\usepackage{amsmath,amssymb}
\usepackage{graphicx}
\usepackage{booktabs}
\usepackage{multirow}
\usepackage{subcaption}
\usepackage{textcomp}
\usepackage{xcolor}
\usepackage{placeins}
\usepackage[font=small,labelfont=bf]{caption}
\usepackage[numbers,square,sort&compress]{natbib}
\usepackage{xurl}
\usepackage[colorlinks=true,allcolors=blue!50!black]{hyperref}
\usepackage{authblk}

\title{Speed-to-Power Dominates Energy Costs for AI Data Centers: Technoeconomic Analysis of Powering AI in the Continental United States}
\author[1,2]{Alex C. Newkirk\thanks{Corresponding author: \href{mailto:anewkirk@andrew.cmu.edu}{anewkirk@andrew.cmu.edu}}}
\author[2]{Daniel L. Gerber}
\author[1]{Erica R. H. Fuchs}
\author[2]{Alex Hubbard}
\author[1]{M. Granger Morgan}
\author[2]{Arman Shehabi}
\author[2]{Sarah J. Smith}
\author[1,3,4]{Constantine Samaras}
\affil[1]{Department of Engineering and Public Policy, Carnegie Mellon University, Pittsburgh, PA}
\affil[2]{University of California, Berkeley, Berkeley, CA}
\affil[3]{Department of Civil and Environmental Engineering, Carnegie Mellon University, Pittsburgh, PA}
\affil[4]{Wilton E. Scott Institute for Energy Innovation, Carnegie Mellon University, Pittsburgh, PA}
\date{}
\begin{document}
\maketitle

\begin{abstract}
The rapid growth of artificial intelligence is driving demand for new data centers at large scale and on short deployment timelines. To power them, firms connect to the commercial grid or contract generation directly, but interconnection and gas-turbine delays now incentivize on-site generation. We compare grid-connected and off-grid data center energy infrastructure (AC- and DC-coupled solar-plus-storage, islanded natural gas, and grid interconnection) across the continental U.S. Using an open-source technoeconomic framework built on empirically calibrated AI training loads, climate-driven cooling, stage-by-stage power conversion, and physics-based battery degradation, we model designs providing equivalent energy services. Our central contribution is a speed-adjusted cost of energy that values the compute unlocked by faster construction, making options with different deployment timelines directly comparable on one physically consistent basis. We find DC coupling reduces lifetime energy costs by 18\% (9¢/kWh) on average versus AC counterparts by eliminating inverters and improving efficiency. Though rarely cheapest on unadjusted energy cost, DC-coupled microgrids' faster deployment makes them the lowest-cost modeled option in 85\% of U.S. locations once foregone compute is valued, or 55\% under accelerated natural gas timelines. For AI infrastructure, deployment speed rather than unit energy cost determines the economically preferred path to power.
\end{abstract}

\section{Introduction}\label{sec1}

Expanding data center infrastructure to train and deploy cutting-edge artificial intelligence (AI) models is driving a surge in electricity demand in the United States (U.S.) \cite{shehabi_united_2024, iea_energy_2025}. This breaks a decade of equilibrium: data center energy consumption in the U.S. remained flat in the 2010s as operational efficiency improvements offset growing computational use \cite{shehabi_data_2018, masanet_recalibrating_2020}. The combination of growing AI use, increasing AI model size, and the empirical association between the computational resources allocated to AI models and their performance \cite{kaplan_scaling_2020, amodei_ai_2018} means operational improvements now fuel expanded capabilities rather than reduced energy consumption.

Specialized hardware performs the computations for AI training and inference in dedicated facilities. Given the performance benefits of high computational density, some of the largest data centers now operate at the scale of hundreds of megawatts of rated power capacity, and announced projects would exceed the energy demand of existing facilities by an order of magnitude \cite{cohn_land_2025}, on the scale of gigawatts \cite{ghaffary_oracle_2025}. To avoid falling behind competitors in capabilities and market share,  AI infrastructure firms prioritize deployment speed over other considerations  \cite{dorris_data_2025, morales_openai_2025, zeitlin_data_2025}. 

Currently this urgency is confronting a slower physical and policy reality in the legacy electricity system. The utility-planning and regulatory apparatus that reviews interconnection requests, plans capacity, and approves permitting, accustomed to nearly two decades of flat electricity consumption, struggles to meet the growing demands of not only data centers, but the current and future demand from building electrification, vehicle electrification, the growth of major industries, and the other drivers of load growth \cite{costa_samaras_ais_2025}. Along with these changes, grid interconnection queues (the delay while large loads or new generation await formal approval by local operators to connect to the electric grid and begin construction) are rising, exceeding 5 years in some regions \cite{rand_queued_2024, gorman_grid_2025, armstrong_can_2024, quint_assessment_2025, randQueued2025Edition2025}. Beyond administrative capacity, component manufacturing also bottlenecks energy infrastructure: grain-oriented electrical-steel \cite{mckenna_major_2024}, grid transformers \cite{moseman_transformer_nodate, kennedy_grid_2025}, high voltage circuit breakers \cite{boucher_challenge_2024}, and other electrical transmission and distribution equipment often have multi-year backlogs due to demand outpacing supply. Natural gas combustion turbines for power plants are similarly facing a shortage \cite{patel_gas_2025}, with new turbine waitlists quoted as ranging from four to seven years across turbine models \cite{clark_latest_2025, anderson_us_2025}. 

In an effort to bypass the slow pace of traditional grid interconnection and begin operations earlier, data centers are using exclusive power purchase agreements (PPA) \cite{elliott_exxon_2024, mandler_three_2024}. Other facilities are developing their own power infrastructure to operate partially \cite{judge_microsoft_2023} or fully islanded from local electric grids \cite{maxwell_data_2025, sayCrusoeRedwoodLaunch2026}. To better understand the physical and economic constraints on the planned expansion of data center infrastructure and associated energy demand, we analyze the technical and economic performance of different off-grid energy solutions for AI data centers across the continental U.S.

AI workloads involve large quantities of parallel operations performed by groups of specialized computer chips, generally termed accelerators, most commonly graphical processing units (GPUs). Accelerators require supervisory CPUs, integrated high-bandwidth memory, and high-performance networking and interconnection components. This bundle of hardware is referred to as a compute node. Nodes are analogous to a traditional enterprise server, though they draw 10-20 times the power, with a group of nodes referred to as a cluster, an 8-GPU configuration being a current commercial standard \cite{nvidia_nvidia_nodate}.

AI model development involves two phases: \textit{training}, during which  models learn from massive datasets through intense computation; and \textit{inference}, when trained models process queries to generate output. While both training and inference involve periods of high computational intensity, training workloads are especially challenging to energy systems.  Large-scale training for frontier models can require months of continuous operation, with individual nodes demanding 70-80\% on average of their manufacturer rated maximum power for the duration of the training \cite{shehabi_united_2024, latif_single-node_2025, patel_characterizing_2024}.

Electricity demand ``in-workload'' during training is flat at hourly or daily granularity, but locally, the power demand of individual nodes can fluctuate by thousands of watts in periods of just milliseconds \cite{newkirkEmpiricallycalibratedH100Node2025a,patel_characterizing_2024}, with fluctuations time correlated across all involved nodes. Transients at these frequencies on cluster scale strain electric grids \cite{choukse_power_2025} that serve commercial AI data centers \cite{li_ai_2025, li_unseen_2024} and high-performance scientific computing  facilities \cite{bates_electrical_2015, stewart_grid_2019}. This can reduce local power quality, both within the data center, disrupting other compute nodes, and beyond it, affecting sensitive electrical loads (such as hospitals and other data centers) as well as nearby residential utility customers. While infrastructure upgrades can address these transients, the cost allocation mechanism used to assure a minimum level of power quality can have broader impacts. Within our analysis, we bound the facility-level cost of managing these transients (Results, \textit{Robustness and Sensitivity Analysis}).

Electrical utilities charge customers both a capacity fee and a per-kilowatt-hour fee to recover generation costs and fixed transmission and distribution infrastructure costs. Even when there is sufficient existing generation capacity to meet data center loads \cite{norris_rethinking_2025, costa_samaras_ais_2025}, new data centers can drive up local electricity prices as ratepayers finance infrastructure upgrades to transmit and distribute power. A recent report by the Virginia legislature found typical residential customers could face \$37 per month higher electricity costs due to data center infrastructure expansion \cite{sarte_data_nodate}, and a national projection of electricity cost found increases by up to 57\% in some regions by 2030 \cite{Johnson_2026}.

Data center electricity consumption consists of IT equipment (servers, storage, networking) plus cooling and power distribution infrastructure. Cooling costs vary dramatically by climate: cold regions achieve near-optimal performance through free cooling, while hot, humid locations require energy-intensive mechanical chillers \cite{lei_climate-_2022}. Power Usage Effectiveness (PUE), the dimensionless ratio of total facility power to IT power, quantifies this overhead, with the highest performing facilities approaching the theoretical performance limit of 1.0.

Operational electricity generally comprises over 90\% of a data center's environmental footprint \cite{siddik_environmental_2021, schneider_life-cycle_2025}, with carbon intensity, water use \cite{mytton_data_2021}, and cost varying substantially by location. The performance of wind, solar, and geothermal assets varies according to local conditions. This spatial heterogeneity means no single microgrid architecture is likely to be best across all locations.

Previous work by Baranko et al. \cite{baranko_fast_2024} examined both the geographic and speed-to-deployment performance of solar and storage microgrids for AI data centers. The authors noted that while clean-firm generating technologies such as small-modular nuclear or advanced geothermal show promise for data center applications, their deployment supply chains are unlikely to mature in time to meet near-term AI demand. They found that the geographic potential for solar based microgrids in the United States was substantial \cite{cjhandmer_solar_2024} and that the estimated two-year combined permitting and construction timeline presented a key advantage for these technologies. 

The performance of solar and battery-energy-storage system (BESS) microgrids also varies according to system architecture: the generation and storage technology together with the AC or DC power-distribution scheme. Unlike thermal, hydroelectric, and wind power plants, solar photovoltaic (PV) panels generate direct current (DC) electricity. Because the U.S. power grid operates at 60 Hz alternating current (AC), connecting PV systems to the grid requires an inverter, a power-electronic device that converts this DC output into the 3-phase, 60 Hz AC and a power transformer to match grid voltage. Any voltage or current conversion step necessarily introduces some operational losses. Dispatchable storage resources like BESS and reversible fuel cells are notably charged by and discharge DC. To avoid conversion and clipping losses, storage and PV systems can interface prior to inversion, a microgrid architecture known as DC coupled. DC coupling improves system cost \cite{gerber_simulation-based_2018, ollas_energy_2023}, efficiency \cite{vossos_techno-economic_2018, lo_franco_efficiency_2021}, and reliability \cite{ sandelic_reliability_2019, gerber_comprehensive_2021}. 

While data centers typically require power at a variety of different voltage levels, they are unusually well suited to accommodate DC power distribution. An idealized data center contains only IT and cooling loads: chips already operate on DC power, with power supply (PSU) or power distribution (PDU) units typically performing the necessary voltage conversions. The primary electrical load for cooling is the motor powering fans, pumps, and compressors. Motor drives or electronic speed controllers perform the necessary voltage conversions, and are highly efficient. Although most often these are AC induction motors, the improvement in brushless DC motors shows potential for a fully DC facility. Further, electricity use is the primary operating cost of a data center, meaning the economic case for efficient DC power systems infrastructure in data centers is stronger than in most other facilities \cite{prabhala_overview_2018, gerber_energy-saving_2019, losalamosnationallab.lanllosalamosnmunitedstatesDCMicrogridsScoping2015}. For these reasons, previous research has explored all-DC data center designs \cite{emerge_alliance_380_2013, becker_dc_2011}, with pilot facilities in the United States \cite{tschudi_dc_2008}, Japan \cite{michael_keen_hokkaidos_2025}, and Europe \cite{miller_abb_2012}. 

There are gaps and limitations within this literature. Previous microgrid modeling for AI data centers has relied on static assumptions for cooling, power systems losses, or the off-grid natural gas case. Although separate research communities have investigated the advantages of DC power architectures for both solar+storage systems and data centers independently, to date, no work has explored the efficiency gains from integrating these approaches. While industry emphasis on deployment speed is well documented, no open framework exists for quantifying the tradeoffs between speed and cost. Each of these unknown quantities is decision-relevant and can potentially shift which architecture a developer should select.

\begin{figure}[tbp]
\centering
\includegraphics[width=0.62\linewidth]{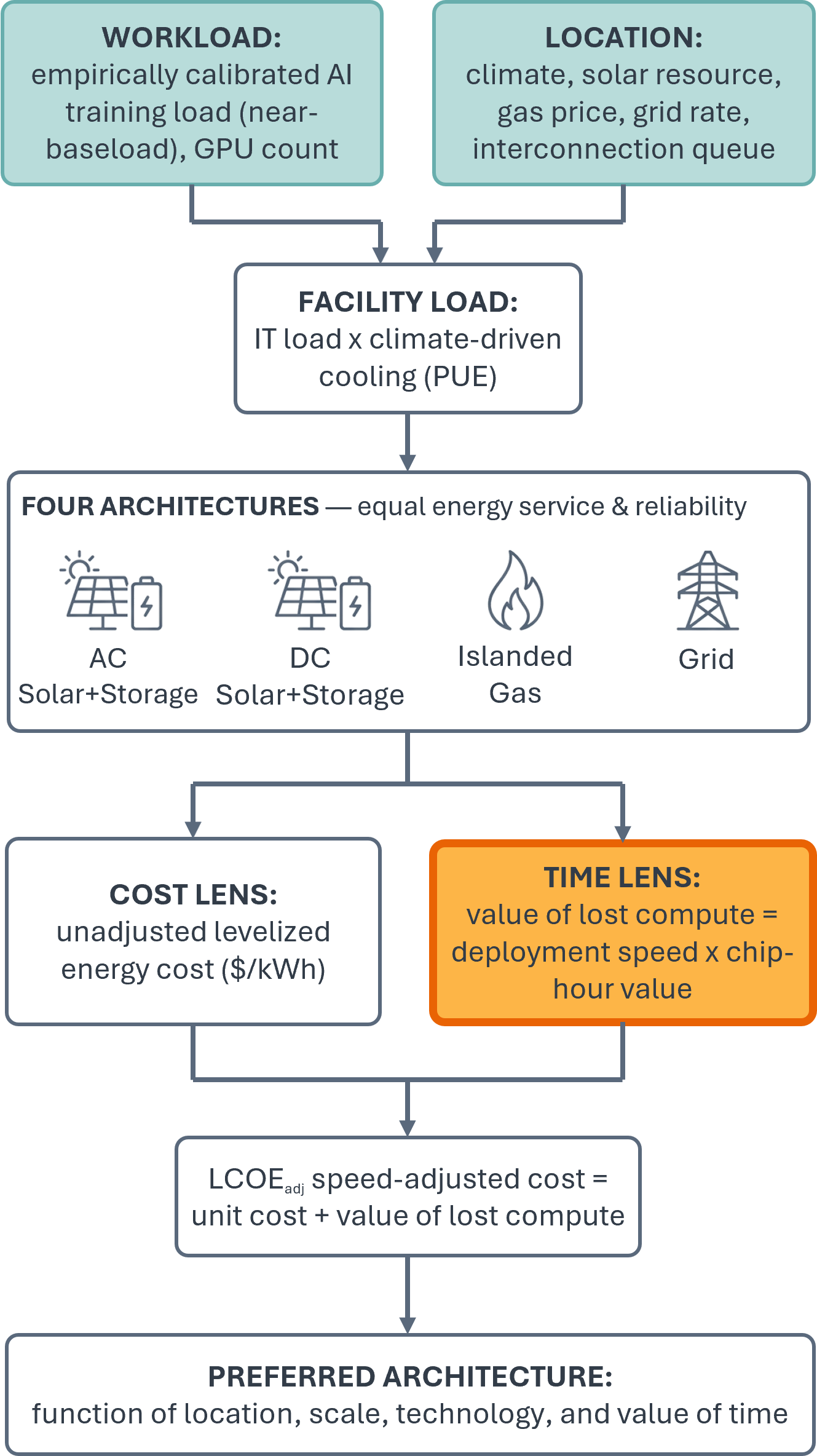}
\vspace{1em}
\caption{\textbf{Analytical framework.} For a given quantity of chips, workload (in this research empirically calibrated to AI training) and location (climate, solar resource, gas price, grid rate, and interconnection queue), we build an hourly facility load combining IT demand with climate-driven cooling (PUE). We then size and dispatch three islanded architectures, AC solar+storage, DC solar+storage, and islanded natural gas with diesel backup, and compare them against grid interconnection. Each deliver equivalent energy service at equivalent reliability through an explicitly modeled power-conversion path. Each option is assessed on two axes: its unadjusted levelized cost of energy (\$/kWh) and the value of compute forgone during construction (set by deployment speed and chip-hour value). Their sum, a speed-adjusted cost ($\mathrm{LCOE_{adj}}$), determines the preferred option, which varies with location, scale, technology, and the value of time.}
\label{fig:framework}
\end{figure}

In this work we develop an open-source, end-to-end technoeconomic model of self-supplied power for AI data centers (Fig.~\ref{fig:framework}). For a given scale and location, we construct an hourly facility load---empirically calibrated, near-baseload AI training demand plus weather-driven cooling performance. We design and dispatch three islanded architectures, AC and DC solar-plus-storage and islanded natural gas with diesel backup, comparing them against grid interconnection. Because each architecture must deliver equivalent energy service at equivalent reliability, and every shared input is held identical across cases, the resulting cost differences are attributable to the architectures themselves rather than to differences in assumptions between them. We evaluate each option on two axes: levelized cost of energy, and the time value of energized compute. We find that scale and location shift gas competitiveness enough to redraw the map of preferred architectures, and that DC coupling materially lowers lifetime energy cost relative to AC distribution. Quantifying the value of deployment speed through the time value of energized compute, we find that it dominates differences in unit energy cost in selecting the preferred architecture. The speed-premium concept generalizes to other time-sensitive capital-deployment decisions beyond data centers.

\section{Methods}
Our analysis proceeds in three stages. First, we develop the hourly facility load that any power system must serve, combining IT demand with location-specific cooling requirements. Second, we simulate candidate power distribution systems meeting this load profile, tracking hourly energy flows from generation through storage and conversion to final consumption. Third, we search each architecture's design space (varying design parameters such as battery capacity or turbine model within allowable configurations) to identify configurations that minimize levelized cost while achieving required reliability. Comprehensively detailed methods and a full inventory of modeling parameters are available in the Supplementary Information. 

We compare four power architectures (AC solar+storage, DC solar+storage, islanded natural gas with diesel backup, grid interconnection) serving equivalent IT and cooling loads at a particular location with identical reliability requirements. While cooling performance differs by locations, and power systems vary how efficiently they deliver energy from source to load, in a given location, for a given quantity of GPUs, all architectures must deliver equivalent energy services at equivalent reliability.

We model AI training specialized facilities, operating at near baseload utilization \cite{grahovacDataCenterIndustrial2026}, composed of a fleet of identical 8-H100 node equivalents: avg power 7.3 kW, max 8.5 kW inclusive of memory and interconnect components \cite{newkirkEmpiricallycalibratedH100Node2025a}. We calculate hourly PUE using Lei and Masanet's framework \cite{lei_climate-_2022}, selecting the minimum annual PUE option for the location. Power systems are configured sufficient to deliver the maximum IT load multiplied by the 99th percentile PUE and a 1.05 contingency factor. We index facility scale by GPU count and report the corresponding peak as a design load in megawatts (a 100{,}000-GPU facility, for example, has a design load of roughly 131~MW); average draw is lower, but power infrastructure is scoped to this peak. We compare each system in simulation year 27, end-of-life for the PV modules (25 year lifespan), with a base-case discount rate of 7\%. For scenarios with BESS, we assume one single end-of-life replacement of the battery modules occurring at the conclusion of operational year 13. All financial values are in 2022 dollars, and all discounting occurs mid-year.

To compare different microgrid systems, we develop candidate power-flow models of all paths between generation and facility loads, with the associated conversion losses. We model all power systems outward from the main low-voltage distribution bus, accumulating the efficiency of every voltage and conversion step to each load and from each source. The candidate power system architectures are shown in Figure~\ref{fig:powerflow}, and more comprehensive discussion of their design and efficiency is available in the Supplementary Information. Grid-connected and natural-gas facilities follow conventional AC data-center architecture \cite{chalise_data_2015, barroso_datacenter_2019, sartor_user_2021}, with the natural-gas generator terminal voltage (13-19 kV depending on turbine) stepped up to a facility medium-voltage distribution spine (34.5 kV) and then stepped down to 480 V AC in the data hall. 

\begin{figure}[tbp]
\centering
\includegraphics[width=\textwidth]{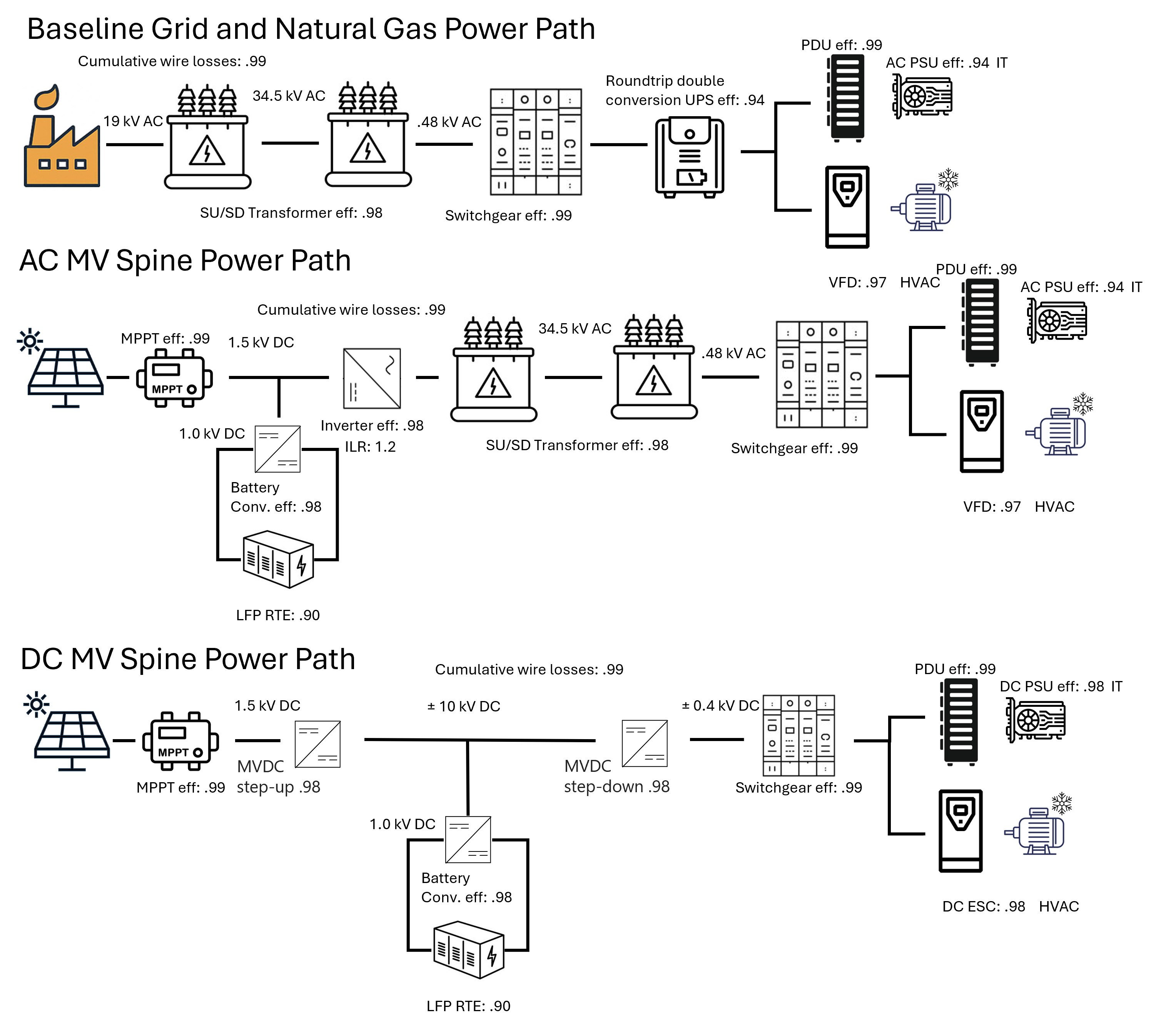}
\caption{\textbf{Candidate MV power flow models within the facility fence-line with modeled component efficiencies.} Power flows are idealized, without batching or correspondence to any particular facility scale. Grid power and islanded natural gas are assumed to have equivalent power systems architecture inside the facility fenceline. Note that while not shown on the plot, the diesel backup power for the grid comparison and islanded natural gas cases is modeled as entering the facility through the same main power distribution path. Fault protection components, monitoring components, or supercapacitor banks are not shown. BESS and UPS components are shown abstracted from the facility layout and composition, and may in practice be centralized for the facility, distributed at the sidecar, rack, or node level, or some combination. Microgrids that use BESS for backup substitute these assets for the UPS via a small share of capacity embedded in the LV data hall (see Supplementary Information). AC power systems distribute power at 480 V bus voltage, with embedded PSU or motor drive power electronics converting to the voltages required by IT and HVAC loads. Bidirectional grid-forming inverters convert the DC outputs to AC at the specified bus voltage and frequency. DC power systems operate at a ±400 V DC bus voltage, with DC-DC converters stepping up to bus voltage and PSUs and ESCs stepping down to deliver the required load voltages.  }
\label{fig:powerflow}
\end{figure}

Our PV+storage microgrids are based on typical centralized generation and storage designs, with panel collection strings wired into 1500 V DC blocks. We consider two distribution architectures: a small, modular layout with a single low-voltage bus serving co-located generation, storage, and load; and a medium-voltage layout in which an MV spine separates centralized generation from the data-hall loads and distributes power across the campus, as ampacity and PV gathering constraints require at scale. The medium-voltage topologies are the primary cases; we detail the LV architectures in the Supplementary Information (Supplementary Network Topology and Storage Placement; Supplementary Fig. 12). In the low-voltage AC case, storage is coupled to the data-hall switchgear through grid-forming inverters. In the medium-voltage case, power from the grid-forming inverters passes through a step-up transformer to the 34.5 kV AC distribution spine, then steps down to 480 V AC (3-phase) at the data hall.

In the AC architecture, the bulk of storage is DC-coupled to the PV field, sharing the PV inverter and charging upstream of it. In the all-DC architecture, it is instead centralized off the ±10 kV spine \cite{chenTechnoEconomicalAssessmentAC2022}. In both, a smaller increment of approximately the nameplate power of the load it protects (on the order of 2 MW of combined load and storage per LV spur \cite{hinzImpactOpportunitiesMediumvoltage2016, stienekerMVDCDistributionGrids2018, quPlanningAnalysisDemonstration2019}) is distributed to the data-hall LV buses for local outage protection, and in the AC case for grid-forming. Provisioning this distributed increment changes LCOE by 1\% or less compared to the purely centralized case. In the primary AC case, we assume these local grid-forming pods maintain power quality through training transients without full double-conversion isolation IT power. See the Supplementary Network Topology and Storage Placement section for more detailed discussion, including the cost performance of different topologies and headline cost of instead double-conversion conditioning the full load.

In the medium voltage all-DC microgrids, the 1500 V DC blocks are stepped up to $\pm$10 kV for distribution across the campus \cite{cuiDesignEvaluationSimulation2024, huMaximumPowerPoint2017, siemaszkoMVDCDistributionConcept2023, chenTechnoEconomicalComparisonMVAC2023}, and then stepped down to low distribution voltage within the data hall. Note that within the data hall ($\sim$30 MW), loads are containerized into $\sim$2MW \cite{hetzeneckerModelingApproachMVDC2023, quPlanningAnalysisDemonstration2019, siemaszkoMVDCDistributionConcept2023} low-voltage blocks to limit both continuous bus currents and prospective fault currents to within the interrupting capability of available DC protection. Within the data hall, the embedded battery’s DC/DC charge controller handles grid-forming through droop-voltage regulation, distributing power on a $\pm $400V bus \cite{DataCenterDC}. Server power supplies and cooling motor drives perform the final conversion to load voltage. This design is compliant with the emerging OCP standard \cite{NVIDIA800VDC}. We obtained our design logic, component efficiencies, and performance data from component datasheets, operational estimates from industry experts, existing all-DC pilots, and efficiency curve data \cite{beheshti_wide-bandgap_2020, pires_dc_2023, tschudi_dc_2008, gerber_cost_2023, gerber_comprehensive_2021, gerber_simulation-based_2018, hinzImpactOpportunitiesMediumvoltage2016, siemaszkoMVDCDistributionConcept2023, rothmund99Efficient102019}. 

To simulate PV+BESS microgrid systems, we adopt the vectorized hourly energy balance approach from \cite{james_offgridai-uscost-calculator_2025, baranko_fast_2024}, modified with a bus-centric architecture that explicitly models conversion losses through different power-flow paths. The PV generation model fetches TMY weather data (irradiance and temperature) for the specified coordinates from the National Solar Resource Database (NSRDB) \cite{senguptaNationalSolarRadiation2018} and feeds these into pvlib's ModelChain to calculate DC power output. We assume the battery system is composed of the underlying modules in 4-hour, 60 MW Lithium-Iron-Phosphate (LFP) packs, each rated to store a total of 240 MWh of usable energy \cite{ramasamy_us_2022}, though note that this system is comprised of 48 containerized 1.25 MW units. We chose these battery modules because they are well characterized in the literature and represent the current commercial standard. Note that minimum capacity increment for commercial designs is on the order of 1 MW of rated capacity, and we discuss sensitivity to step size and cost sensitivity of explicit incremental modeling as opposed to continuous optimization in the Supplementary Information. 

We select LFP as our battery chemistry due to its excellent durability and ramp rate \cite{goodenough_li-ion_2013} and current commercial deployment in grid-scale storage applications. The hourly dispatch algorithm prioritizes serving facility load from solar generation. Surplus solar charges the battery (subject to power rating and available headroom), with remainder curtailed. Deficits trigger battery discharge (subject to power rating and stored energy). Supplementary Fig. 1 traces this dispatch over a representative week. Component degradation follows established empirical models: solar modules degrade linearly (1\% first year from light-induced degradation, then 0.55\% annually) \cite{yang_comprehensive_2025}. Batteries follow BLAST-Lite's \cite{smith_lithium-ion_2021} empirically calibrated physics-based framework for LFP calendar aging and cycling stress as functions of temperature, state-of-charge, and charge-discharge patterns. We model mid-life battery replacement at year 13, reflecting industry practice of discrete infrastructure upgrades rather than continuous capacity additions. See Supplementary Methods for more details, with Supplementary Figs. 2-4 characterizing battery fade dynamics and the performance of our computational proxy, Supplementary Fig. 9 illustrating optimization stability and the Supplementary Sensitivity Analysis section discussing alternative battery replacement timing and augmentation strategy.

Natural gas combustion turbine power plant design differs from PV system optimization due to its discrete solution space: multiple interacting discontinuities that preclude simple scaling from reference designs. We generate candidate plant configurations using a library of commercially available turbines, apply engineering constraints derived from industry practice, and evaluate the full lifecycle economics of each specific configuration. Our design tool proceeds from a library of 16 reference natural gas turbine models classified as aeroderivative, F-class, or H-class. We model turbine performance accounting for part-load efficiency penalties, ambient temperature derating, and combined-cycle plant configurations where applicable. We adopt a configuration specific auxiliary load \cite{gulenImportanceAuxiliaryPower2010} empirically calibrated to EIA reference plant values \cite{mirletz_annual_2024}. We filter designs to those capable of delivering minimum facility power as defined in the facility load, operating within code-compliant load factor bounds, and passing design filters to reflect operational plant configurations (no mixed model fleets, etc.) We scope backup diesel systems to supply minimum facility power, and model annual diesel operations as matching the gas plant's expected annual unserved energy from planned maintenance and forced outages. We evaluate lifecycle costs for all filtered configurations and select the minimum-cost option. See Supplementary Methods for implementation details. 

We calculate LCOE for each candidate design as the ratio of the sum of the present value of costs to the sum of the present value of delivered energy. We source PV+storage capital costs from NLR's Annual Technology Baseline (ATB) \cite{mirletz_annual_2024}, disaggregated into component-level costs using detailed breakdowns from \cite{ramasamy_us_2023}, with land requirements parameterized at 4.17 acres-per-MW DC \cite{ong_land-use_2013, bolinger_land_2022}. Operating costs for PV came from \cite{ramasamy_us_2023} and battery storage O\&M from the PNNL energy storage cost database \cite{pacific_northwest_national_lab_esgc_2024}. Natural gas plant costs are from NLR ATB \cite{mirletz_annual_2024} for F-class and H-class turbines and EIA's Annual Energy Outlook \cite{larose_capital_2024} for aeroderivatives, with costs scaled to specific capacities using empirical scaling exponents of 0.67 for capital and 0.80 for fixed O\&M following established engineering practice \cite{tribe_scale_1986, james_quality_2022}. For a data center in a location, an optimization algorithm searches through the uptime-compliant designs within each architecture for the LCOE minimizing configuration.

We model project development timelines based on industry data and expert consultation. Solar+storage systems follow a 24-month timeline consistent with \cite{baranko_fast_2024} and validated by  industry experts. Natural gas plants use construction estimates from \cite{larose_capital_2024}: 40 months for simple cycle and 42 months for combined cycle configurations, with turbine delivery (conservatively estimated at 36 months based on fastest reported timeline by OEM) as the potential rate-limiting factor. We assume 30\% of construction is bottlenecked by turbine delivery; remaining work proceeds in parallel. For grid interconnection, we use supply-side queue data from \cite{rand_queued_2024} as a proxy, assigning each state the 2018-2023 median duration from initial request to commercial operation for its ISO/RTO region. See the Supplementary Information for implementation detail. 

Data centers contain GPUs that could generate revenue immediately if deployed elsewhere. Each month of construction represents foregone computational output: training runs not completed, models not deployed, competitive position ceded to faster rivals. This opportunity cost varies dramatically between architectures: grid interconnection requiring 5 years represents 3 years of lost operations compared to 2-year solar construction.

We quantify opportunity cost using GPU spot market rental rates. At \$2.40/GPU-hour (H100 rental index price in $y_{0}$ dollars \cite{bloomberg_lp_sdh100_2025}), an operator unable to extract at least this value should  rationally rent their compute to the market. This rental price thus represents the floor utility of the marginal GPU-hour committed to an internal workload. For hyperscalers balancing frontier model training (likely worth substantially more) against routine inference and fine-tuning (potentially worth less), this spot price provides a reasonable value proxy. For AI-specialized firms running exclusively proprietary training workloads, it likely understates value.

We calculate the value of lost compute in year $t$, $VLC_t$, as the idle GPU-hours multiplied by rental rate for all pre-operational years (Eq.~\ref{eq:vlc}):
\begin{equation}\label{eq:vlc}
\text{VLC}_{t} = \sum_{h=0}^{H_{\text{idle},t}} N_{\text{GPU}} \times p_{\text{spot}}
\end{equation}

We incorporate foregone compute value as an opportunity cost during construction years of any given system, creating a speed-adjusted LCOE that endogenizes deployment timeline (Eq.~\ref{eq:lcoe_adj}):
\begin{equation}\label{eq:lcoe_adj}
  \text{LCOE}_{\text{adj}} =  \frac{\sum_{t=0}^{n}\frac{(C_{\text{cap},t} + C_{\text{op},t} +
  \text{VLC}_t)}{(1+r)^{t+0.5}}}{\sum_{t=0}^{n}\frac{E_t}{(1+r)^{t+0.5}}}
\end{equation}
where $n$ is the evaluation year common across analyzed scenarios. 

We test sensitivity across GPU valuations from \$1.00-4.00/hr and construction timeline variations of $\pm$50\% to assess how these parameters affect architectural selection (see Results and the Supplementary Information).

\section{Results}

Here we outline how our modeled energy system performance varies both by location and by microgrid architecture. Unless otherwise stated, figures depict systems scoped for 99\% annual availability—that is, the islanded microgrid satisfies the facility's full IT-and-cooling load in at least 99\% of hours over the year (equivalently, 99\% GPU uptime), with the small remainder treated as unserved. 
\subsection{Architectural Comparison}

We find that all-DC architectures outperform equivalent AC systems on cost, with an average all-DC advantage of almost 9 cents per kWh (18\%), consistent with the energy performance observed in a pilot \cite{hirose_dc_2018}. At scale this returns material savings. Consider a 100,000-GPU, $\sim$131 MW design-load facility in the Permian Basin of West Texas. The all-DC microgrid delivers energy at \$0.27/kWh against \$0.33/kWh for the otherwise-identical AC system, a saving of \$0.062/kWh. Because both are PV+BESS systems, this gap is avoided hardware: eliminated inverters and higher path efficiency that let the all-DC system meet the same load with less installed generation and storage. This represents a \$570 million reduction in the net present value of energy costs. For systems of this scale, even minor improvements in efficiency return significant absolute savings. 

\begin{figure}[tbp]
\centering
\includegraphics[width=11.4cm]{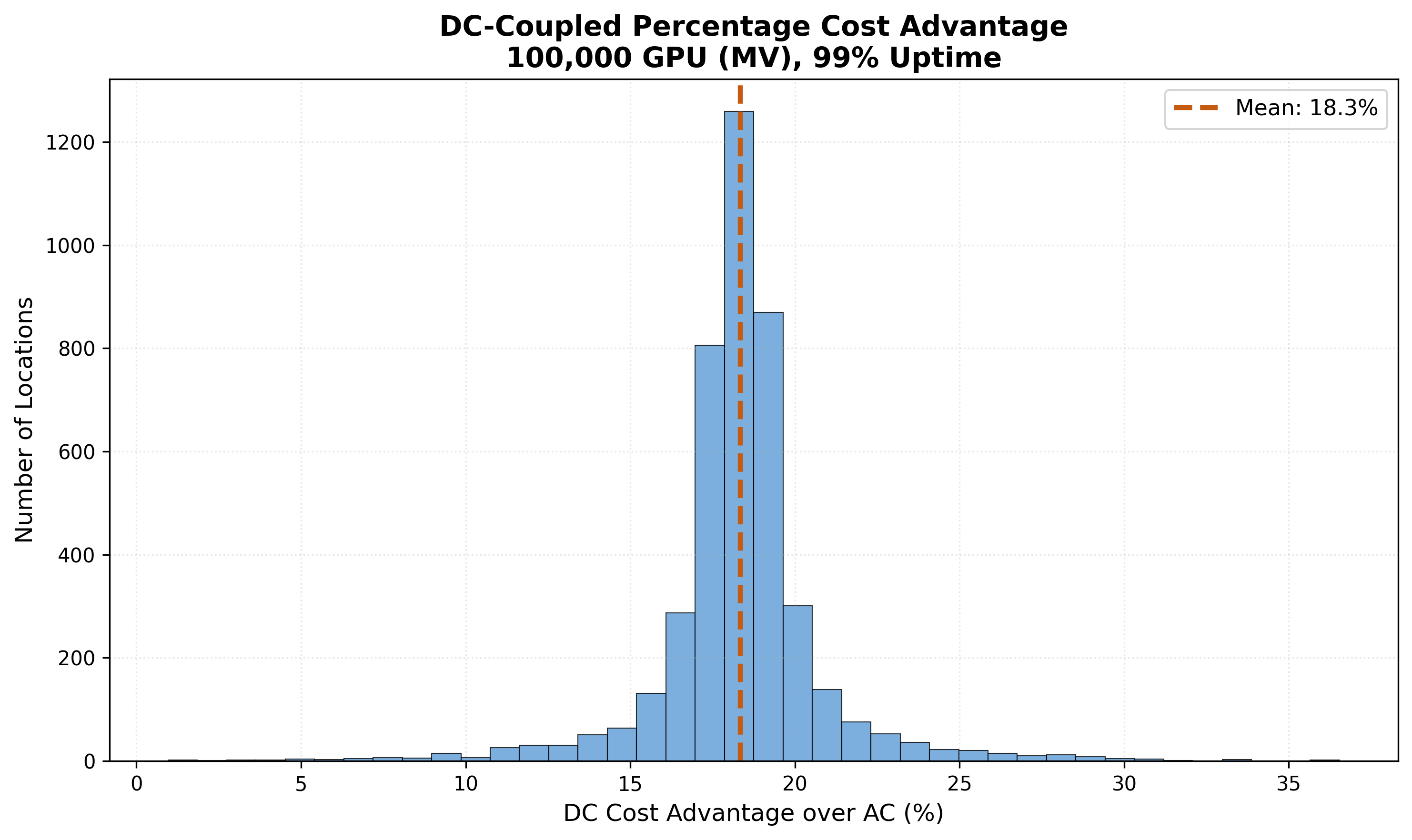}

\includegraphics[width=11.4cm]{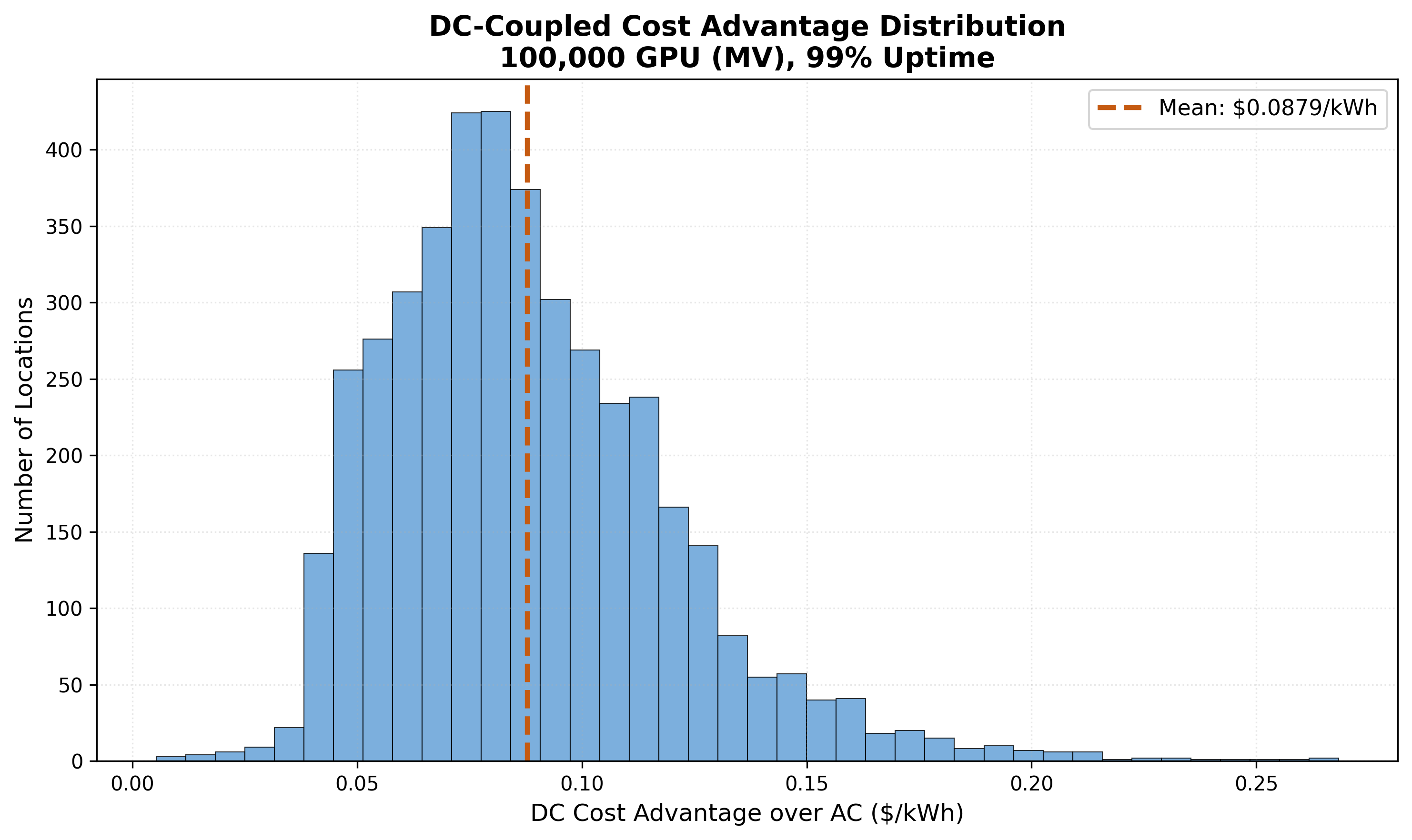}

\caption{\textbf{Cost advantage of DC-coupled systems over AC counterparts for 100,000 GPUs at 99\% uptime across all modeled continental U.S. locations.} Skew in the absolute distribution reflects variation in sensitivity to solar-to-load and battery-to-load path efficiency by location. Scale of DC advantage generally increases in locations with lower quality solar resources as marginal path efficiencies provide returns for highly overbuilt systems. In regions with substantial and consistent solar resources, both AC and all-DC microgrids can scope more aggressively, shrinking but not eliminating the DC cost advantage. 
\textit{(A)} Percentage cost advantage distribution. Mean DC advantage is 18\% of total cost of ownership for a given location. 
\textit{(B)} Absolute cost advantage distribution. Mean DC advantage is \$0.09 / kWh.}
\label{fig:cost_advantages}
\end{figure}

DC cost advantages are not uniform across the U.S., as shown in Figure~\ref{fig:cost_advantages}. Locations differ in the fraction of annual energy that flows directly from PV-to-load versus through the BESS-charge-discharge-load pathway. MV DC power distribution improves efficiency along both battery discharge and PV to load pathways, but it achieves largest performance gains on the battery discharge pathway and by avoiding inverters. See Supplementary Network Topology and Storage Placement for more detailed discussion of the cost drivers of different topologies. 

Across the continental United States, levelized cost varies within each PV architecture by roughly \$0.40/kWh over the central range of locations (5th–95th percentile) and double that spread between the cheapest and most expensive sites. This variation is set less by a location's average solar resource (a long-run climate feature) than by the depth of its worst sustained low-solar stretch (a within-year weather extreme). The system must ride through the longest cloudy spell, and system LCOE tracks the worst five-day insolation window far more closely than annual insolation. Two sites can receive essentially identical annual sunlight yet differ severalfold in cost. At roughly $5~\mathrm{kWh/m^2/day}$ of insolation, a dry interior-West Colorado site and a fog-prone northern-California site differ nearly threefold in cost (\$0.21 vs. \$0.61/kWh), because the latter's worst five-day power production window is dramatically worse. 

\begin{figure}[tbp]
\centering
\includegraphics[width=1\textwidth]{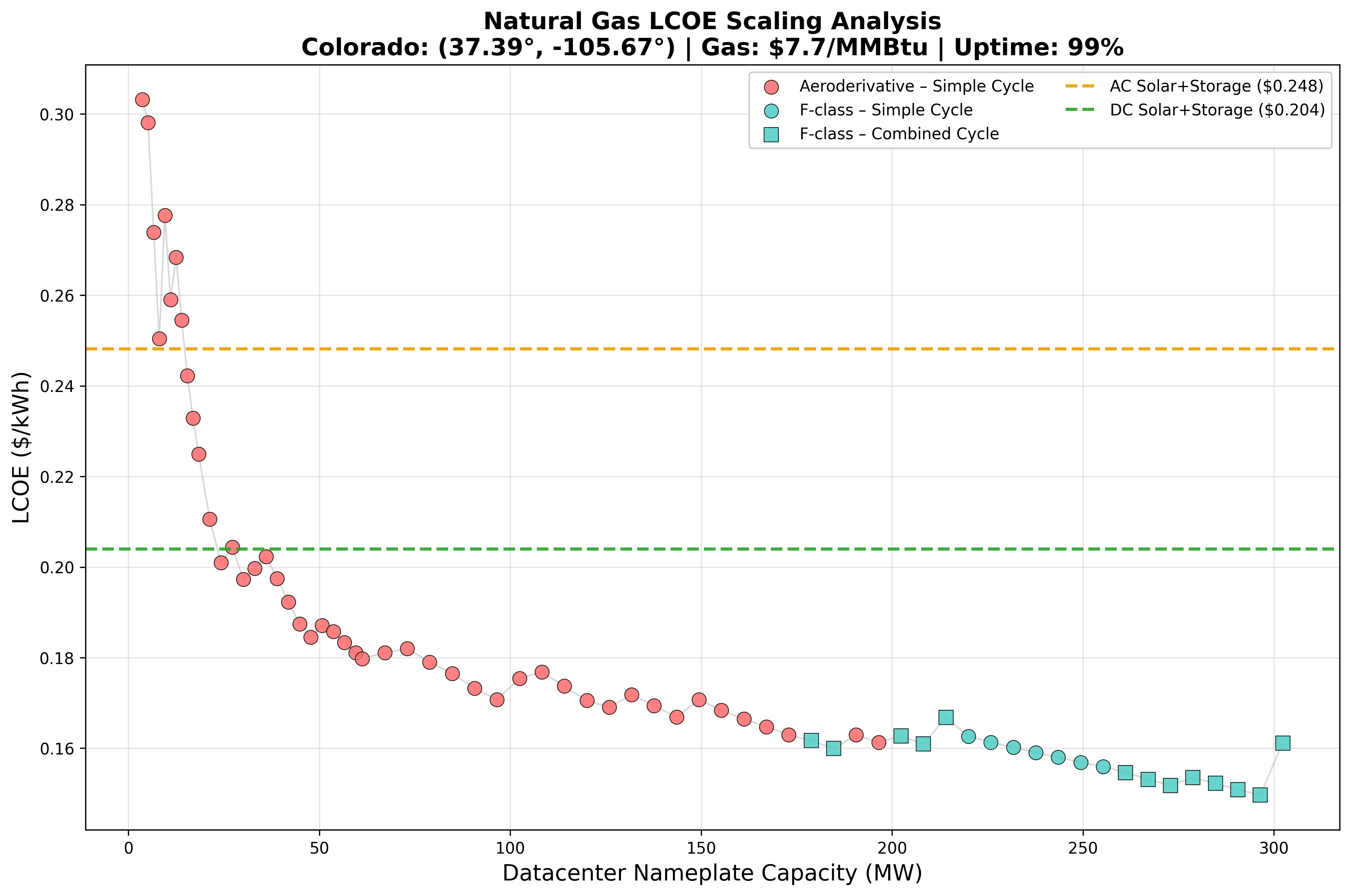}
\caption{\textbf{LCOE of the minimum cost natural gas configurations against different facility design loads at 99\% uptime in eastern Colorado; natural gas price of \$7.70 / MMBTU.} Points correspond to the LCOE for a given design, with color indicating turbine class (coral for aeroderivative, teal for F-class) and shape indicating single cycle (SC) or combined cycle (CC) configuration. Step-sizes between points are increments of 2 MW of design load up from 1-20 MW, 4 MW of design load from 20-64 MW, and 8 MW of design load from 64 MW to 320 MW. Dashed lines indicate the LCOE of medium-voltage DC (green) and AC (yellow) PV+BESS systems at the same location. Natural gas plants exhibit significant scale economies, dropping by $\$0.15$ / kWh over the range of system scales. PV+BESS systems showed minimal unit cost variation with nameplate capacity $(\pm\$0.002$ / kWh), we plot these costs as idealized average values without this noise for visual clarity. Jaggedness on the natural gas curve demonstrates transitions between turbine models and classes, and plant configurations as data center facility scale changes. }
\label{fig:natgas}
\end{figure}

While medium voltage PV-based microgrids are predictably sensitive to local geographic conditions, they are insensitive to facility scale. Modularity penalties remain negligible ($<0.3\%$ median) across realistic component sizes and decrease with system scale; see Supplementary Methods for more detail. When considering the low voltage architectures, either for a small facility, or modular designs comprised of multiple containers, costs decrease a median 2.4\% for DC and 3.7\% for AC compared to their respective medium-voltage counterparts. In contrast, natural gas plants display significant scale economies, as shown in Figure~\ref{fig:natgas}: costs decline sharply as facility capacity increases. Small aeroderivative plants operating at 10-20 MW achieve LCOEs of \$0.25-0.30/kWh, while large combined-cycle configurations decline towards marginal grid prices. The discontinuous structure in these cost curves reflects discrete turbine unit additions and transitions between designs as optimal configurations shift. 

Regional variation in fuel costs and solar resources significantly affects system costs. In the Colorado location shown in Figure~\ref{fig:natgas}, natural gas plants achieve cost parity with DC-coupled microgrids around 30 MW facility capacity. Elevated gas prices would shift this crossover point toward larger facilities, and cheaper gas or lower solar productivity locations shift the crossover towards smaller facilities. Natural gas prices and DC microgrid costs (see Supplementary Figs. 5 and 6) vary by more than a factor of three across U.S. locations, creating substantially different economic landscapes for off-grid AI data centers.

\begin{figure}[tbp]
\centering
\includegraphics[width=0.48\textwidth]{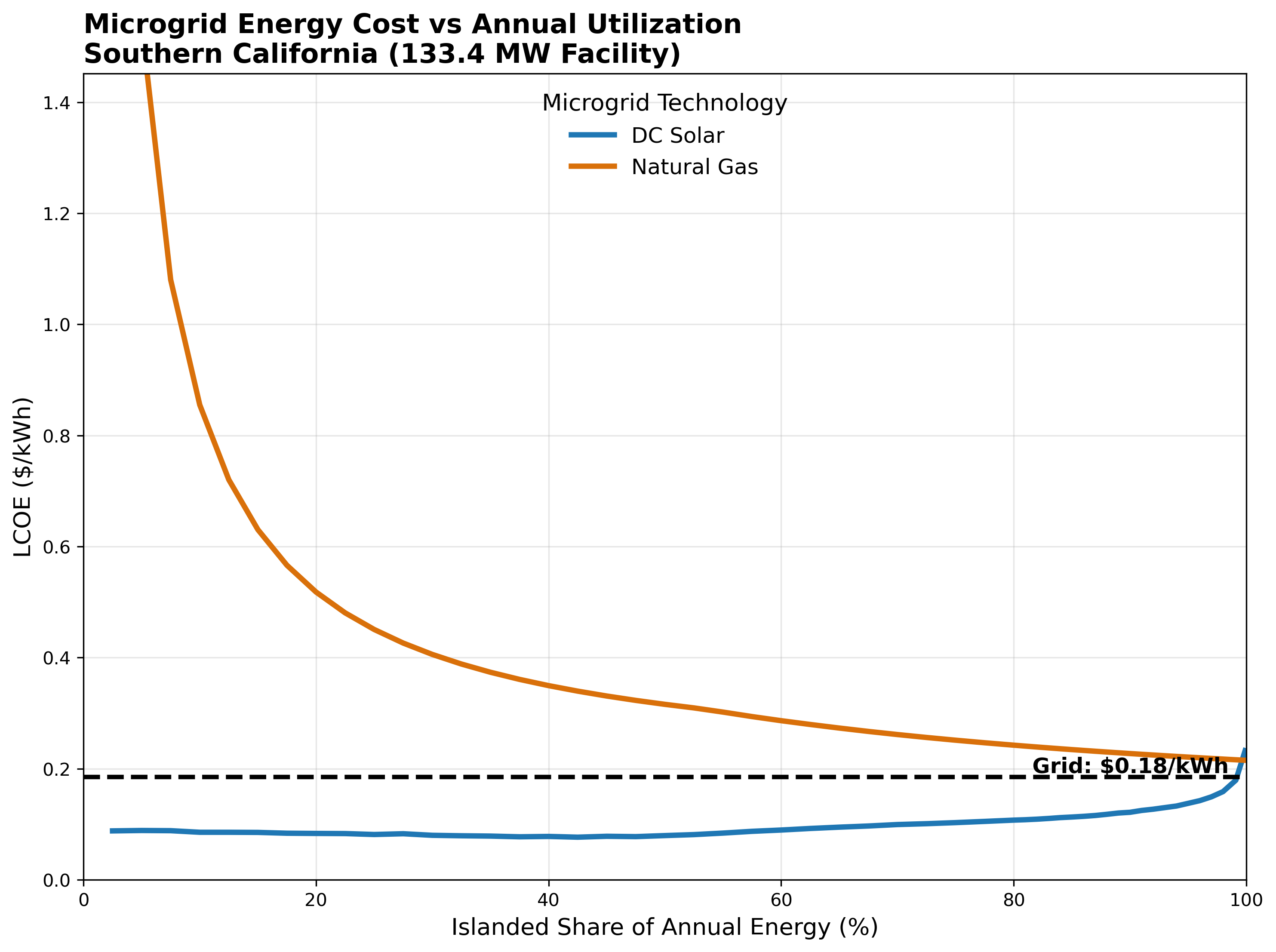}
\hfill
\includegraphics[width=0.48\textwidth]{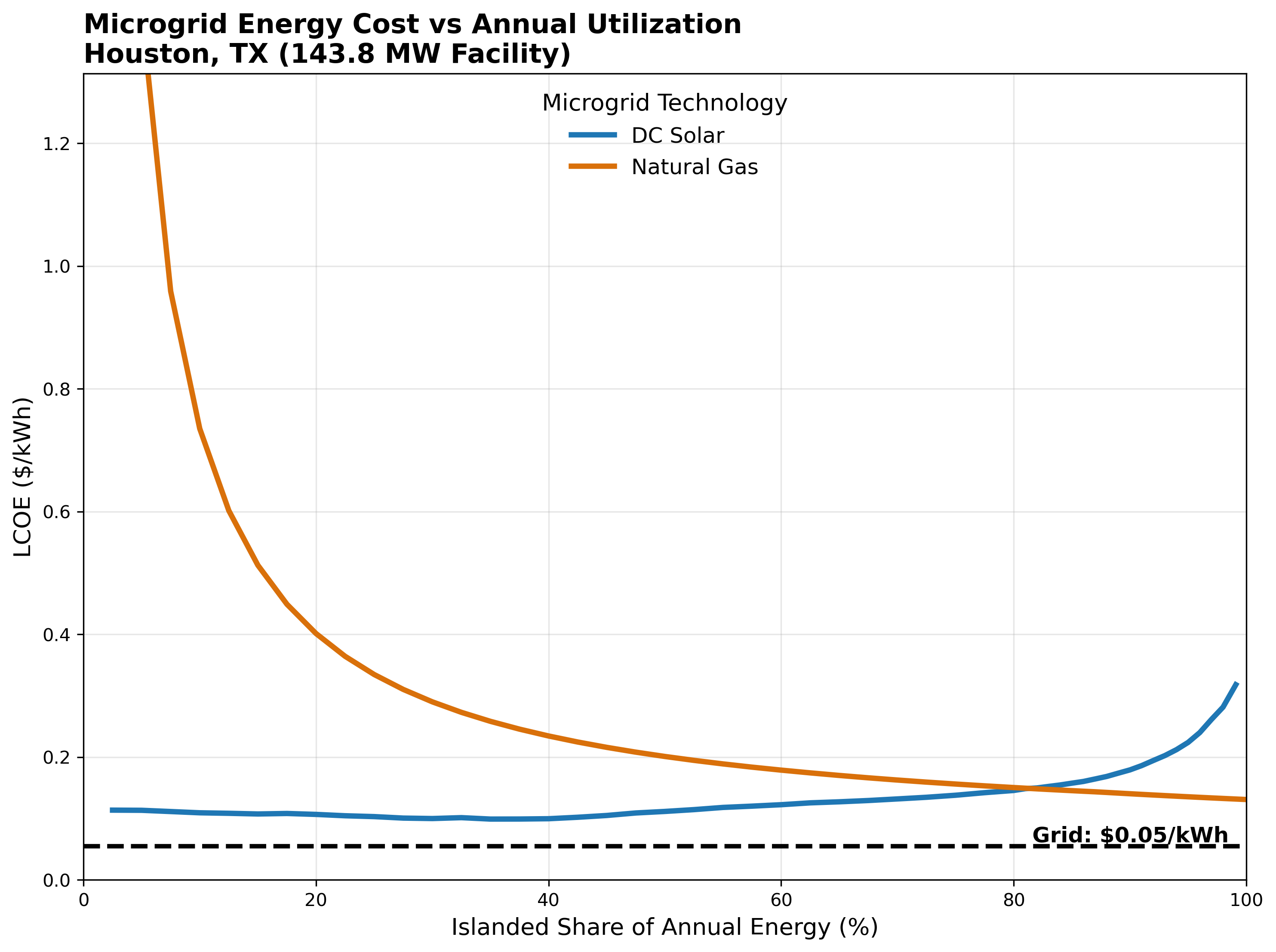}

\vspace{0.5em}

\includegraphics[width=0.48\textwidth]{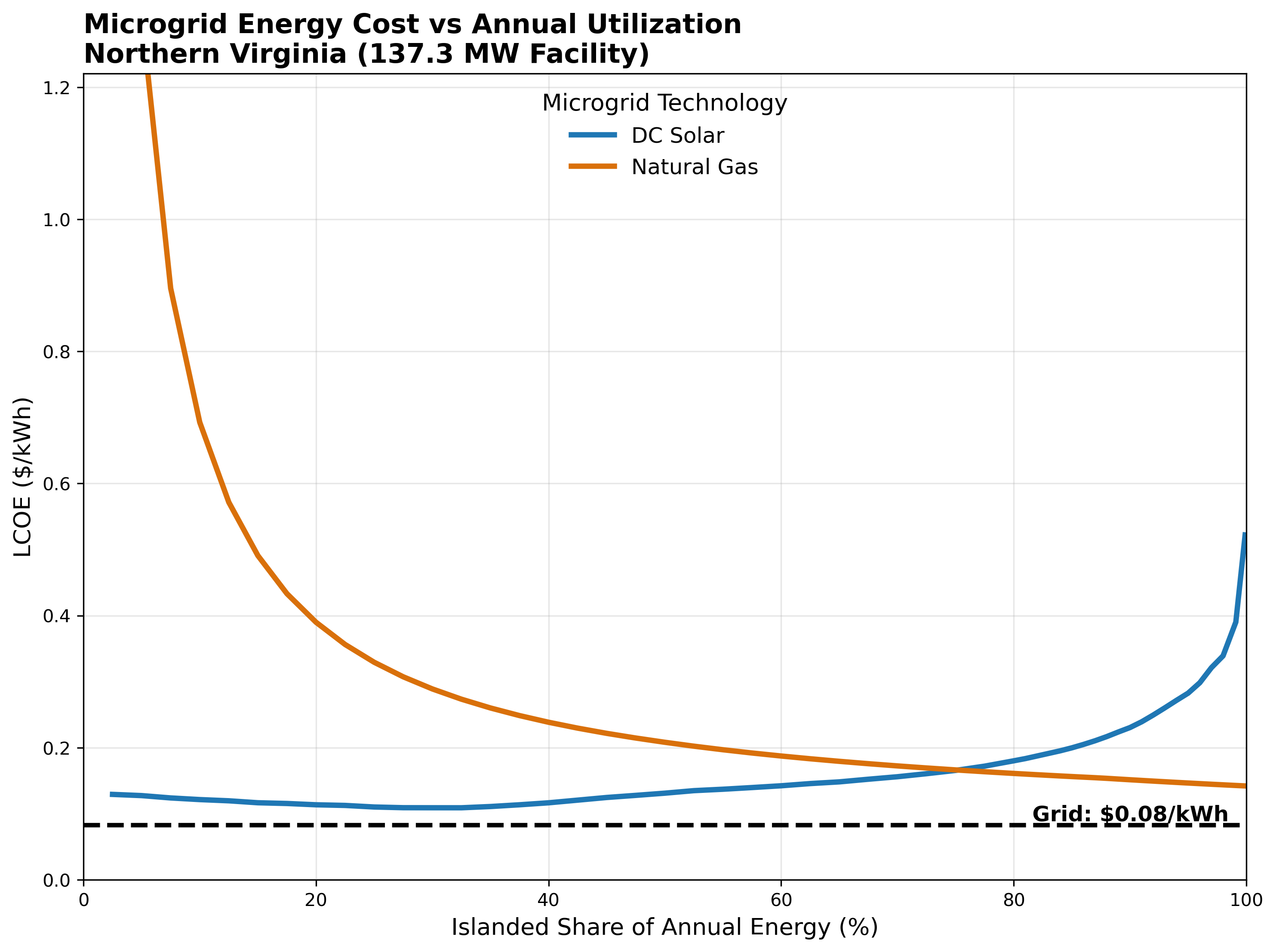}
\hfill
\includegraphics[width=0.48\textwidth]{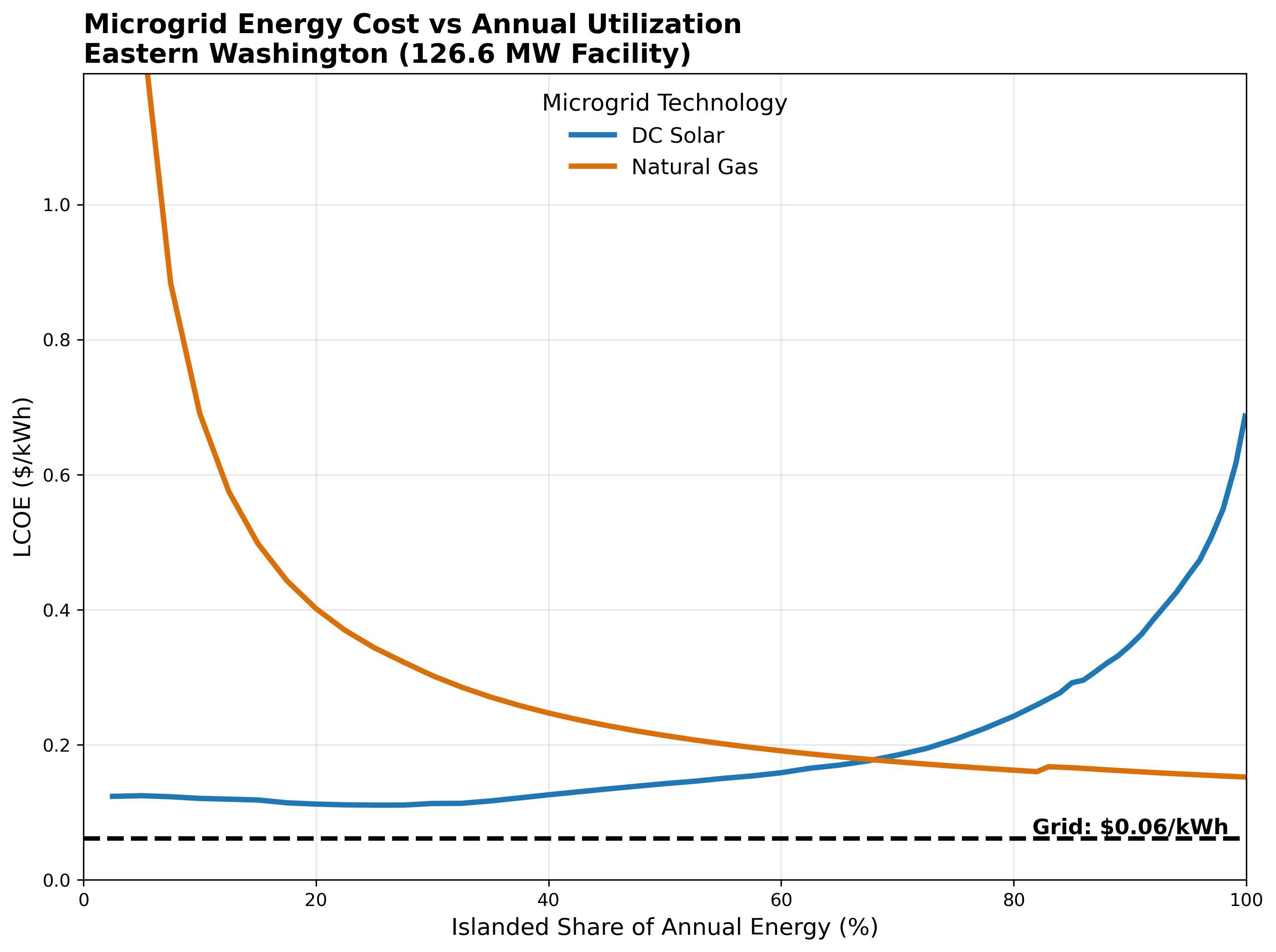}

\caption{\textbf{Annual utilization requirements determine optimal power architecture across diverse geographic and market conditions.} 
Each panel shows levelized cost of electricity versus annual share of energy-service-hours provided by the islanded microgrid for a 100,000 GPU datacenter, comparing MV DC-coupled solar (blue), natural gas with diesel backup (orange), and the grid comparison case (gray dashed). The horizontal axis reads two ways: the annual uptime of a fully islanded facility, with the remaining hours unserved; or the fraction of hours an otherwise grid-tied facility runs on its microgrid. Solar architectures at low utilization are constrained by power capacity rather than energy, meaning costs initially decline as utilization increases. Natural gas costs decline monotonically as fixed capital is amortized over more operating hours. Facility design loads vary by location due to location-specific cooling system performance.
(\textbf{A}) Southeastern California: Exceptional solar resources and high electricity prices (\$0.185/kWh) favor DC solar across all utilization levels. 
(\textbf{B}) Houston, Texas: Natural gas becomes less expensive than DC solar above 77\% annual uptime, though both exceed grid costs (\$0.055/kWh). 
(\textbf{C}) Northern Virginia: Data center alley. 
(\textbf{D}) Eastern Washington State: Solar resource inconsistency makes high reliability with PV+BESS systems expensive. The hitch in natural gas at higher utilization reflects transition to an alternative plant configuration at higher uptimes.  
}
\label{fig:uptime_effects}
\end{figure}

Our fully-islanded cases model operation entirely off-grid, but facilities might opt for a hybrid system. Such facilities would operate for some fraction of the year on islanded power, and would otherwise be grid supplied. Figure~\ref{fig:uptime_effects} shows the performance of microgrids of different architecture as we vary their expected annual usage. Solar-based microgrids require a disproportionate share of investment to increase reliability: each additional ``nine'' of availability—90\% (one nine), 99\% (two nines), 99.9\% (three nines)-requires more capacity. This mirrors macro electricity systems, where infrastructure must be scoped for the worst-case scenario. In the natural gas case, this contingency scoping is effectively priced-in to building a facility capable of delivering baseload power; higher utilization more effectively economizes that upfront investment.

LCOE varies by architecture, facility scale, and geography. Natural gas systems are more cost-effective for large scales across most locations. DC-coupled microgrids are cost-competitive for smaller facilities, and in regions with high-insolation, particularly the solar alley spanning southern California, Arizona, Colorado, Nevada and New Mexico as shown in Supplementary Figs. 5 and 6. This reflects fundamentally different cost structures: natural gas systems spread fixed costs more favorably across greater energy output, while both MV and LV PV+BESS costs scale near-linearly with capacity. Notably, when considering only unadjusted LCOE, microgrids of all architectures are generally dominated by grid interconnection, though our grid costs represent a lower bound (see discussion for details). 

\subsection{Speed-to-Power Premium}

\begin{figure}[tbp]
\centering

\includegraphics[width=0.9\textwidth]{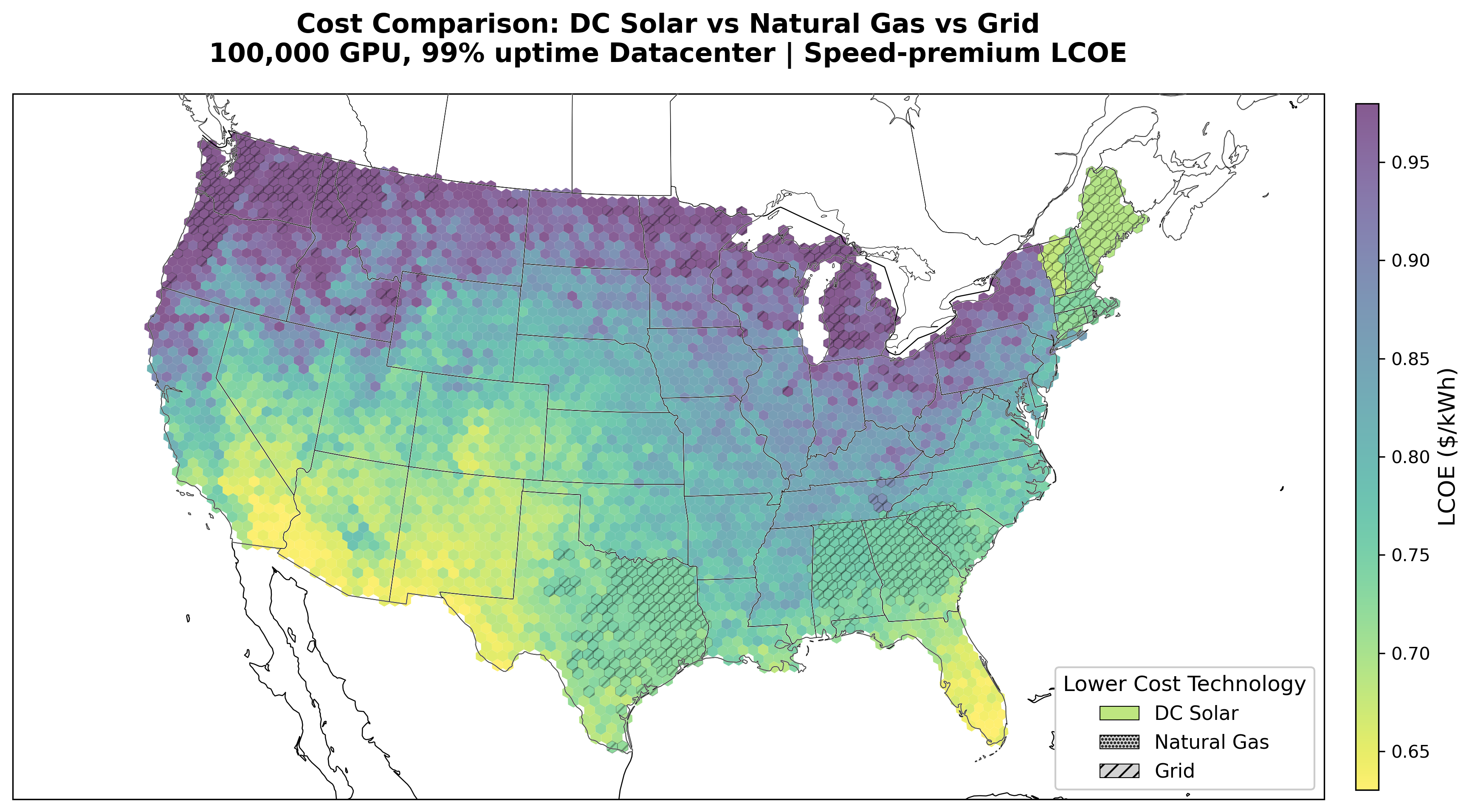}

\includegraphics[width=0.9\textwidth]{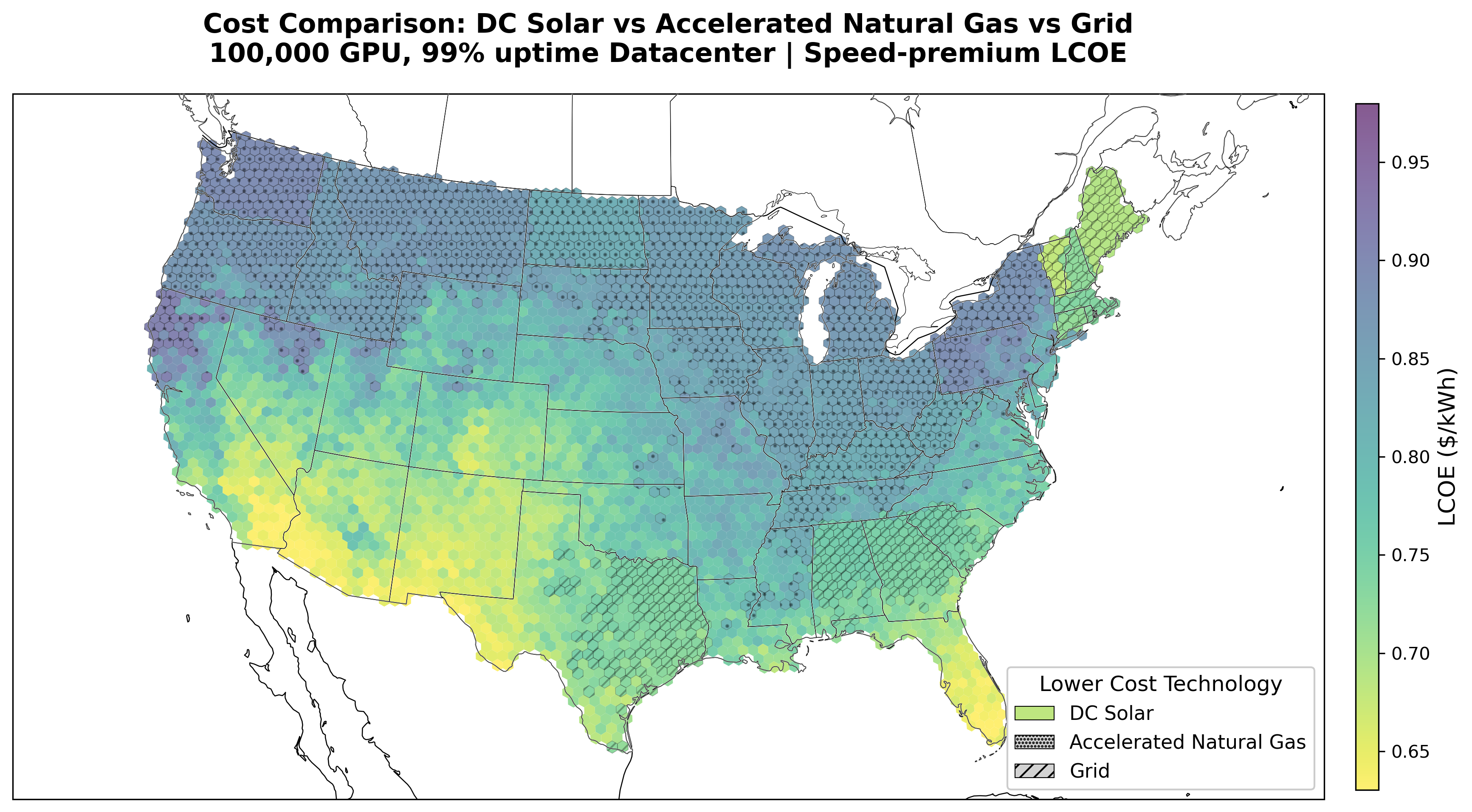}

\caption{\textbf{Minimum speed-premium LCOE for 100,000 GPU facilities across the continental United States.} Color indicates LCOE magnitude; hatching shows lowest-$LCOE_{\text{adj}}$ technology. Maps vary by natural gas deployment speed assumptions, with the top panel (A) showing the baseline case of 4 year combined turbine delivery and plant construction time, and the bottom panel (B) showing the accelerated 3 year case. 
\textit{(A)} Baseline construction timelines: DC solar is the lowest $LCOE_{\text{adj}}$ architecture in 85\% of locations due to rapid deployment despite higher unadjusted energy costs. Grid interconnection remains preferred in regions (15\% of locations) combining poor solar resources, low industrial electricity rates, and abbreviated approval timelines.  
\textit{(B)} Accelerated 36-month natural gas timelines: DC solar (solid) is preferred in $\sim$55\% of locations. Natural gas (dots) is preferred in $\sim$35\% of locations, forming coherent regional deployments in solar-poor areas (upper Midwest, Appalachian Basin, Pacific Northwest). Grid interconnection (diagonal hatching) remains optimal in $\sim$9\% of locations. Even under this aggressive scenario, DC solar microgrids remain the most widely preferred option.}
\label{fig:Speed_premium_comp}
\end{figure}

Incorporating a premium for the speed of deployment alters the architectural comparison results: DC-coupled microgrids prove lowest-cost among our modeled options across 84.9\% of the U.S. Grid interconnection out-competes where cheap electricity and fast interconnection align, as shown in Figure~\ref{fig:Speed_premium_comp}A. Deployment speed dominates other considerations as the value of lost compute ($VLC$) ultimately proves more consequential than unadjusted LCOE (Methods, Eqs.~\ref{eq:vlc}--\ref{eq:lcoe_adj}). This $VLC$ framework applies equally to operational reliability targets. Moving from 90\% to 99\% uptime is worth approximately \$0.20/kWh in avoided lost computing value, while the incremental value of each additional nine falls by roughly an order of magnitude. When we evaluate DC-coupled solar systems across each DOE climate zone, the marginal cost of moving from one to two nines falls below this threshold in all but one case (Seattle, WA), but no location justifies the incremental cost of moving to very high availability (+99.9\%) on compute value alone. 

A DC solar system's deployment speed advantage makes it the default choice across most locations, but where solar resources are sufficiently poor, an opening emerges for alternatives that outcompete on unit energy costs. Grid competitiveness depends on whether the combination of electricity prices and interconnection timelines can overcome the speed disadvantage.

For example, Maine and upstate New York illustrate this dynamic: both regions feature among the worst solar resources in the U.S. The grid is the lowest $LCOE_{adj}$ architecture in Maine despite moderate electricity prices due to the nation's fastest interconnection timeline. NYISO's longer interconnection queue makes DC solar preferable despite moderate utility rates and unimpressive PV performance.

In regions with adequate-to-good solar resources, grid connection succeeds only where inexpensive industrial electricity combines with short interconnection timelines. East Texas and Georgia exemplify this pathway: each has industrial electricity rates below 6¢ / kWh paired with three-year interconnection, making grid connection the lowest $LCOE_{\text{adj}}$ modeled option in these locations.

Natural gas combustion turbine systems are disadvantaged when considering $VLC$ due to correlated regional factors that eliminate the competitive niche. Local gas prices and grid electricity costs correlate because gas-fired generation sets marginal wholesale prices in most markets. Consequently, locations favorable for gas microgrids on engineering costs, such as East Texas, likely also feature inexpensive grid power. There are few locations where gas prices are low enough, interconnection queues long enough, and solar expensive enough that gas is preferred under our baseline assumptions.

\begin{figure}[tbp]
\centering
\includegraphics[width=1\textwidth]{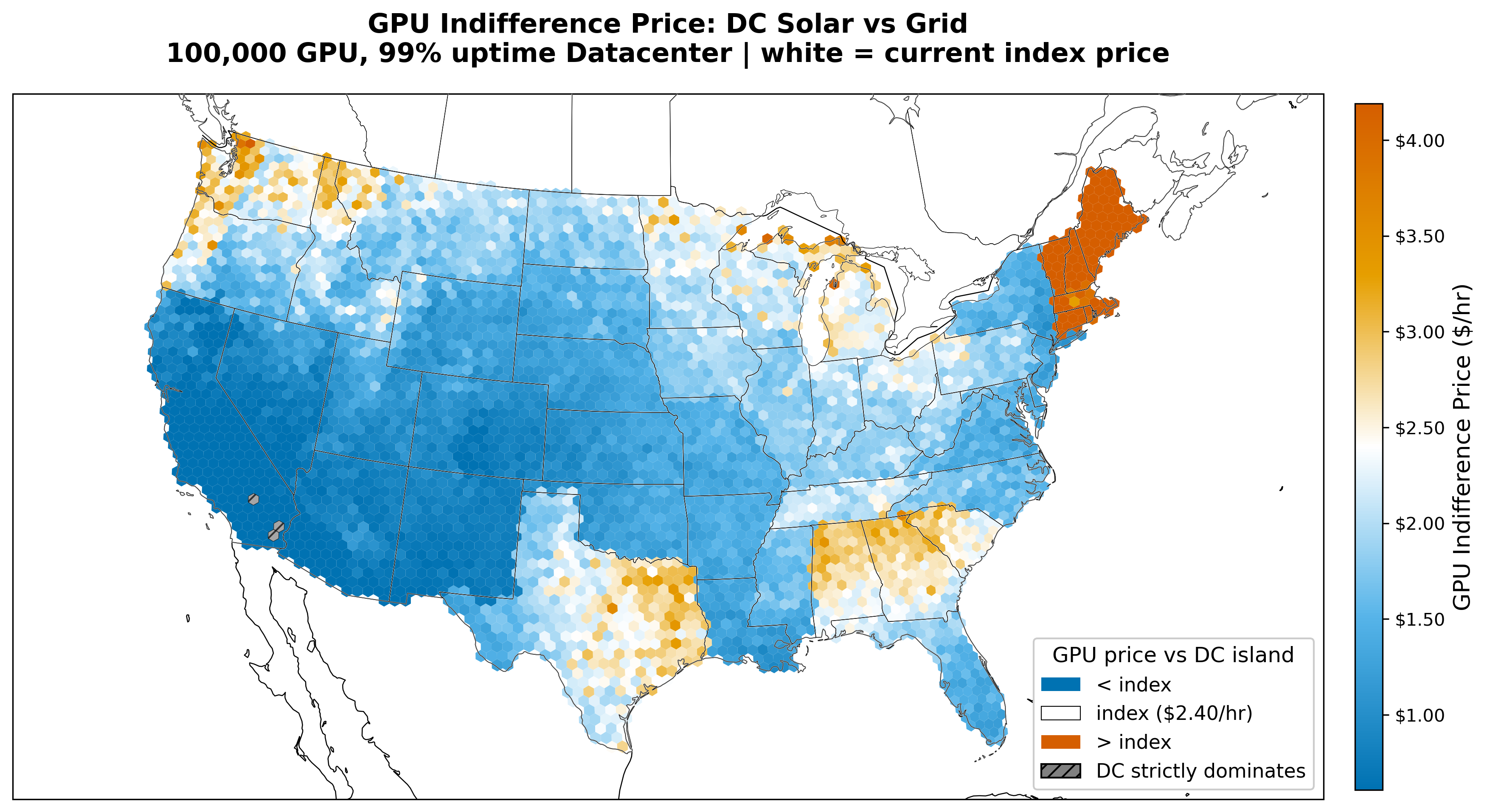}
\caption{\textbf{Value of a chip-hour at which firms would be indifferent between DC-coupled microgrids and grid interconnection.} Holding interconnection timelines at baseline (state-specific, 2.5-5 years). Color indicates indifference utility in \$/GPU-hour; white contour marks the baseline assumption (\$2.40/hr). Blue regions show where DC solar dominates even at low GPU values due to favorable solar resources and/or slow grid interconnection. The median indifference utility is \$1.61/hr. Orange/red regions (concentrated in New England and parts of Texas) show where grid remains competitive even at high GPU opportunity costs due to cheap electricity and fast interconnection. Most of continental U.S. lies below \$2.40/hr (blue spectrum), indicating DC microgrids are preferable to grid interconnection when valuing deployment speed even at lower GPU value.}
\label{fig:gpu_indif}
\end{figure}

To assess robustness of this finding, we parametrically vary the two main determinants of $VLC$: interconnection speed and the value of a chip-hour. Figure \ref{fig:gpu_indif} shows the internal value of a GPU-hour at which a firm would be indifferent between an all-DC microgrid and grid interconnection under baseline interconnection timelines. This threshold varies geographically from $<\$1.00$/hr in high-insolation regions to $>\$4.00$/hr in solar-poor locations (median indifference value: \$1.61 / chip-hour), indicating that architectural choice depends critically on both firm-specific valuations and location-specific resources. See Supplementary Fig. 7 for an analogous plot for the indifference interconnection speed holding $VLC$ fixed.

We also consider an accelerated natural gas deployment case, in which gas combustion turbine microgrids are operational after only three years. This would reflect either idealized plant construction parallelization or significantly ahead-of-schedule turbine delivery. Figure \ref{fig:Speed_premium_comp}B shows how the minimum $LCOE_{\text{adj}}$ for a 100,000 GPU facility changes under this accelerated natural gas timeline. DC solar remains the minimum cost modeled architecture in 55.6\% of U.S. locations. Natural gas microgrids rise to be the lowest $LCOE_{\text{adj}}$ of the modeled options in 35.2\% of locations, showing clear regional deployment patterns: the solar-poor and grid-sluggish areas of the Midwest, the Appalachian Basin, and Cascadia. Grid interconnection remains the $LCOE_{\text{adj}}$ preferred option in 9.2\% of analyzed locations: New England, East Texas, and the interior Southeast. Even under an optimistic natural gas construction timeline, DC solar retains majority deployment preference, and grid interconnection remains preferred in specific regions (New England, East Texas). See Supplementary Fig. 8 for a plot of the indifference natural gas plant construction speed.   

\subsection{Robustness and Sensitivity Analysis}

We evaluate how parameter uncertainty affects both system costs and architectural selection through two complementary approaches: varying inputs for fixed designs to bound cost uncertainty, and re-optimizing under modified inputs and modeling assumptions to test robustness. We present our summary findings of this analysis below, see the Supplementary Sensitivity Analysis section for more comprehensive discussion. 

For fixed designs, natural gas costs are dominated by fuel price: the continental-U.S. range (Oklahoma to California) creates \$0.08--0.11/kWh spreads that exceed the impact of discount-rate or capital-cost uncertainty. Solar+storage costs respond most strongly to the discount rate (a 27--43\% variation across 3--12\% rates); balance-of-system and battery costs have second-order impacts, and land cost is negligible---requiring a sixfold increase to match even battery O\&M, the next lowest cost influence parameter.

To gauge the robustness of the per-location architecture choice, we flag each tile in which a second architecture falls within a 1\textcent/kWh envelope of the lowest-$\text{LCOE}_{\text{adj}}$ option. Under baseline assumptions, 149 of 4{,}317 locations (3.5\%) sit in this near-indifferent band, all at the solar--grid boundary; under the accelerated three-year gas case the count rises to 348 (8.1\%), the additional count split roughly evenly between DC-solar (168 tiles) and gas-preferring (136 tiles) locations, with the remainder grid. These near-indifferent tiles remain a small minority and do not skew toward any one technology as runner-up.

Re-optimization leaves both the selected design and its cost stable. Across the plausible range of each technical performance input the re-optimized cost moves little: battery round-trip efficiency is the largest single lever, at $-2.6\%$ to $+3.2\%$ over 0.85--0.95, and power-conversion component efficiencies shift cost by $\pm$1--3\%. The selected design itself rarely changes: of 72 natural gas parameter variations across four scales, only five altered the chosen turbine configuration, each a power-electronics improvement that let a smaller plant clear the minimum-load filter. Across independent optimizer seeds, the recovered designs agree within $\sim$1\% in LCOE (median over fourteen sites), while component size varies more (5--7\%), indicating a near-flat cost basin in which solar and battery capacity substitute on the margin (see Supplementary Information Optimization Stability for more detail).

Two modeling choices warrant focused treatment. First, in the base case, we replace the battery once, at year 13. Re-anchoring the degradation schedule to a year-10 or year-16 replacement and re-optimizing shifts LCOE by $-0.6\%$ to $+5.1\%$ (AC) and $-1.1\%$ to $+3.5\%$ (DC), and moves the optimal battery capacity itself by a few percent, as later replacement requires somewhat more initial storage to hold reliability through the more-degraded pre-replacement year. Replacing only faded capacity through fixed-period augmentation (installing additional modules to replace faded capacity but keeping the existing fleet in use), rather than swapping the full system, reduces LCOE by roughly 8\%. These results are consistent with the high state-of-health ($\approx$89\%) of the modules in the modeled replacement year. We retain full replacement in our baseline case as a deliberately conservative operational assumption.

Second, we size each system against a single typical meteorological year (TMY). Re-optimized against each historical year from 2010 to 2023 at 28 sites spanning all 14 DOE climate zones, TMY proves a mildly optimistic design: it sits a median 4.5\% (\$0.012/kWh) below the historical-year mean (Supplementary Fig. 11). This is a right-skewed tail effect, not a uniform offset: the reliability constraint makes cost a nonlinear function of the worst sustained multi-day cloudy stretch. Stitching together a composite of broadly typical months necessarily tends to trim the worst-in-a-decade stretches. As an estimator of the across-year mean cost, TMY is nonetheless about twice as precise as a single randomly drawn year (See Supplementary Information Design-year weather sensitivity for more detail).

Fuel-price volatility enters only the natural gas case, and its magnitude is fixed at design time. Smaller facilities are limited to fuel-exposed simple-cycle plants (a committed elasticity near 0.0116~\$/kWh per \$/MMBtu, nearly independent of location); only sufficiently large facilities ($\gtrsim$180~MW, in high-fuel-price regions) adopt the lower-heat-rate combined cycle (Supplementary Fig. 10).  A large facility that is averse to fuel-price risk, or expects prices to rise, can therefore choose combined-cycle configuration partly as a fuel price hedge, exchanging higher capital cost for a more favorable heat rate. This option is not economically available to small turbine plants. Propagating each state's historical price dispersion through the committed plant moves LCOE within a 5th--95th-percentile band of roughly \$0.005--0.024/kWh (median $\sim$\$0.011/kWh) with no shift in the central estimate. 

Choices internal to the power system change the results less than the architecture itself. Reallocating storage among the PV field, the medium-voltage spine, and the data hall changes LCOE within $\pm$1.5\%. The one exception is power conditioning: routing the entire IT load through online double-conversion, rather than maintaining quality through grid-forming alone, adds $\sim$5.9\% to base LCOE ($\sim$6.4\% when also conditioning cooling). These values reflect our modeled topologies under full conditioning; how much of the load a facility must condition, and how aggressively, varies with hardware \cite{choukse_power_2025}, load profile, and the tolerance of co-located loads. The all-DC advantage itself decomposes into a cumulative-distribution-path-efficiency component of $\sim$7.9\% and a roughly equivalent avoided-capital component from the eliminated inverter and AC collection equipment. This delta persists unless DC-side balance-of-system costs approach double our central estimate (see Supplementary Information Sensitivity Analysis).

\section{Discussion}

For planned AI data centers in the continental U.S., we find the value of deployment speed overwhelms differences in energy costs. This result highlights an economic advantage for solar+storage microgrids, which can operate after 24-month construction timelines. Conversely, grid interconnection can take 4-5 years, creating substantial opportunity costs in lost computing revenue as a result of the delay in beginning operations. Islanded microgrids avoid interconnection studies and associated regulatory delays, though quantifying geographic variation in permitting timelines remains beyond this analysis. Under optimistic natural gas combustion turbine deployment scenarios (36 months), DC-coupled solar remains lowest-cost among our modeled options in 55\% of continental U.S. locations. 

\subsection{Limitations}

We adopt several simplifying assumptions that could affect our results. Most consequentially, we model grid interconnection as purchasing industrial electricity at prevailing rates after an interconnection delay \cite{noauthor_electric_2025}. This omits the cost of infrastructure upgrades that large loads may trigger: substation expansion, transformer procurement, transmission reinforcement, and reconductoring. While utilities nominally bear these costs, they recover them through rates, effectively creating cross-subsidies to data centers from other ratepayers. Accurately modeling such infrastructure costs would increase the grid LCOE by an uncertain but potentially substantial amount. Our comparison therefore represents a lower bound on grid costs, strengthening our finding that microgrids are competitive even in the face of optimistically modeled grid scenarios.

We estimate grid interconnection timelines from the regional median duration of the supply-side queue from \cite{rand_queued_2024}, assigned at the ISO/RTO or state level. This proxy has known limitations: demand-side interconnection processes may differ systematically from supply-side requests, and state-level aggregation obscures any intrastate variation in queue congestion. Supply-side queues represent the best-available empirical benchmarks for large-load interconnection timelines. 

We design systems using typical meteorological year (TMY) data, but industry-leading uptime will require systems to survive worst-case, not average years. Re-optimizing against each historical year from 2010--2023 confirms TMY to be a mildly optimistic sizing basis, a median 4.5\% (\$0.012/kWh) below the historical-year mean. Our hourly simulation therefore captures average conditions well but may modestly undersize solar+BESS systems relative to what firms would deploy to manage weather extremes. Additionally, our use of single-location TMY datasets means adjacent locations with similar long-term climates may show diverging results if their TMY datasets captured different realizations of weather, particularly in regions with high spatial climate heterogeneity.

We assume mature supply chains for DC power systems that in some cases do not yet exist at scale. Design standards remain uncodified, some DC fault protection equipment is precommercial, and the skilled workforce overwhelmingly specializes in AC systems. Power electronics datasheets provide our component values, but integrated system performance at 100+ MW requires empirical validation. The 9¢/kWh average DC advantage represents expected savings; early deployments may capture only partial gains while encountering integration challenges and supply chain constraints we do not model. Trade or geopolitical disruptions to the supply of PV modules or LFP cathodes would delay deployment. 

Our natural gas modeling uses state-level industrial gas prices as uniform proxies, obscuring intrastate heterogeneity. Facilities near pipeline hubs may face lower costs than modeled, while remote locations could face premiums. The natural gas turbines we modeled  span aeroderivative, F-class, and H-class units, with representative performance. However this does not exhaustively catalog all generator models, primarily affecting discrete optimization space rather than fundamental cost trends.

Our central estimates treat fuel and grid electricity prices as fixed regional averages. We bound the effect of natural-gas fuel-price volatility in a sensitivity analysis (Results, \textit{Robustness and Sensitivity Analysis}), finding it modest, but we do not model grid electricity price fluctuations, nor do we carry these risk premia through into our central LCOE. Solar+storage microgrids offer the strongest cost certainty: after initial capital recovery, operational costs are predictable, and more importantly, very low. In contrast, natural gas systems face commodity price risk throughout their operational life, though long-term supply contracts can partially hedge this exposure. Grid-connected facilities face both fuel price passthrough and regulatory uncertainty in rate structures. These omitted certainty premia likely favor solar microgrids on average, as they eliminate exposure to both fuel and regulatory risk, while potentially improving gas systems’ relative attractiveness over grid connection in markets where capacity contracts lock in fuel costs but retail electricity prices remain volatile.

Our analysis assumes frictionless land acquisition for all generating technologies. Solar installations at modeled scales require hundreds to thousands of acres, while natural gas facilities require substantially smaller operational footprints. However, all energy infrastructure faces siting challenges including environmental assessments, zoning approvals, and community opposition. Our sensitivity analysis shows land costs themselves have negligible impact on LCOE. If acquisition challenges manifest primarily as higher costs rather than timeline delays, this limitation has minimal effect on our findings. However, if land acquisition significantly extends deployment timelines asymmetrically across technologies, it could alter the speed premium dynamics that drive our comparative findings.

\subsection{Future Work}
Future work should empirically assess the performance of the DC power systems for AI data centers. DC facilities use bus voltage as their primary control signal, and the ability to maintain that voltage under transients depends on the slew rate and response time of the individual components. If the BESS lack sufficient ramping capabilities, or if the duty-cycle would dramatically alter the system stressor, facilities might require additional components for fast droop voltage control (e.g. supercapacitors). Converter or microgrid-energy-management-system performance may also represent a bottleneck. Notably, LFP cells are capable of rapid ramping \cite{peterson_lithium-ion_2010}, and routinely interface with PV installations that exhibit comparable transients \cite{curtright_character_2008}, meaning LFP batteries show technical promise as the regulator of bus voltage in these systems.

Modeling heterogeneous compute fleets and varied workloads (e.g., training and inference, different node types) would require only minor modifications. This would enable analysis of more complex facility operations, including fleet dispatching and distributed computing. While not all workloads are amenable to this kind of distribution and demand response (memory throughput and interconnect latency can make this infeasible in training, and some inference is time sensitive), both distributed training \cite{duan_efficient_2024} and the scheduling optimization \cite{stojkovic_tapas_2025, stojkovic_dynamollm_2024} are ongoing areas of research. Future work could expand to model facilities serving more diverse workloads, and compare the energy systems costs of concentrating or distributing those computational capabilities between single or multiple facilities. 

Future research should analyze microgrids based on other technologies, including nuclear (both conventional light-water reactors and emerging advanced nuclear designs), advanced geothermal, wind, reciprocating engine, and mixed portfolios. Wind resources are a fuel-free option plentiful in the solar-poor region of the upper Midwest \cite{draxl_wind_2015}.  Mixed architectures could leverage the complementarities between wind and solar generation \cite{lopez_prol_wind-solar_2024}, and natural gas fuel cells could solve the "last mile" of reliability for an otherwise PV and battery-based all DC system. This framework could identify the deployment speed at which these emerging generation technologies can economically compete with mature generation-and-storage systems.

\subsection{Implications}

AI training workloads align well with solar+storage technical characteristics. AI training runs last months with significant post-processing, enabling facility siting in high-insolation regions without performance penalties. High-reliability microgrid requirements (99\%+ uptime) necessitate substantial battery capacity, but the resulting steady daily discharge pattern creates benign duty cycles that maintain storage lifespans beyond modeled replacement intervals. 

Beyond energy supply, battery storage enables proactive facility management. Storage provides hours of response time to address potential supply shortfalls, enabling operators to curtail lower-priority workloads or migrate jobs between facilities \cite{colangelo_turning_2025, chien_reducing_2023}. The uptime target modeled in this analysis assumes a naively dispatched, non-reactive load, and should therefore be interpreted as a conservative lower bound. Facilities leveraging the operational flexibility afforded by storage would in practice achieve higher effective reliability. Battery modules at mid-scenario replacement retain a high state-of-health (89\% on average) suitable for redeployment in grid services, arbitrage, or power quality applications, though we omit this residual value from our analysis.

The economic case for microgrids arises from the speed premium, which results from the high value of AI compute. Our baseline value is likely a floor for the internal utility of a chip-hour. It also reflects contemporary hardware -- more computationally efficient and higher-performance chips provide more services per unit time and energy. As computing and model capabilities increase, so too does the speed premium. 

Data center firms demonstrate revealed preference for deployment speed, co-locating with power generation and acquiring PPAs for aging or retired power plants. However, the number of idled nuclear reactors is finite. While firms can rely on spare capacity and rental gas turbines in the near-term \cite{morales_openai_2025, zeitlin_data_2025}, continued buildout will eventually exhaust these resources. Previous operational commitments\cite{smith_microsoft_2020,google_247_2020,amazon_climate_nodate} may further constrain their choice of power source.

Energy demand to serve new data centers will require new grid supply and delivery infrastructure, and without new policy processes to allocate costs to data centers, this new infrastructure will likely be financed by the base of ratepayers. Given the high value of speed, data centers that supply their own power and recapitalize local infrastructure may merit an expedited process for the remaining permitting \cite{samaras_speed_2025}. Because firms often turn to self-supply in response to interconnection delays, deployment timelines and policy will shape whether AI demand stays on the shared grid or shifts to a parallel, self-supplied system. Such a shift may carry longer-term consequences for the cost, reliability, and environmental footprint of the broader power system that extend beyond the facility-level economics we model here.

\section*{Data Availability}

All input data derive from publicly available sources: NSRDB PSM4 solar resource data \cite{senguptaNationalSolarRadiation2018}, NLR Annual Technology Baseline cost data \cite{mirletz_annual_2024}, NLR solar and storage cost breakdowns \cite{ramasamy_us_2023}, the PNNL energy storage cost database \cite{pacific_northwest_national_lab_esgc_2024}, EIA Annual Energy Outlook turbine costs \cite{larose_capital_2024}, and grid interconnection queue durations \cite{rand_queued_2024}. Processed per-location result tables and the configuration files specifying all model parameters are available in the public repository: \url{https://github.com/acnewkirk/AI-Datacenter-Microgrid-Analysis}. 

\section*{Code Availability}

The technoeconomic model, including all power-flow, dispatch, degradation, and optimization code used in this study, is publicly available at \url{https://github.com/acnewkirk/AI-Datacenter-Microgrid-Analysis} under an MIT License. The repository includes a notebook (\texttt{notebooks/reproduce\_paper\_figures.ipynb}) that reproduces the main-text figures from the shipped result tables. Reproducing the full per-location sweep requires a no-cost API key from the NLR Solar Resource API to obtain typical meteorological year data for arbitrary U.S. locations.

\section*{Acknowledgements}

The authors would like to thank Gerald Robinson, Natalie Bates, Christine Cao, Kevin Watson, Alex McEachern, Jay Whitacre, Hannah Morin, and Mary Anne Piette for their helpful discussions about this work. An earlier version of this manuscript was posted as a preprint under the title ``Technoeconomic Analysis of Microgrids for AI Data Centers in the Continental United States'' (\href{https://doi.org/10.21203/rs.3.rs-8272920/v1}{doi:10.21203/rs.3.rs-8272920/v1}).

\subsection*{Declaration of generative AI and AI-assisted technologies in the manuscript preparation process}
During this research, the author(s) used \textit{Claude Opus} to assist in the creation of software tools, which were checked and run independently. During the preparation of this manuscript, the author(s) used \textit{Claude Opus} and \textit{Claude Sonnet} to improve language, clarity, and typesetting. After using this tool or service, the author(s) reviewed and edited the software and text as needed and take(s) full responsibility for the content of the publication.

\section*{Author Contributions}

A.C.N.: Conceptualization, Methodology, Software, Formal analysis, Investigation, Data curation, Visualization, Validation, Writing -- original draft, Writing -- review \& editing. \\ 
D.L.G.: Conceptualization, Methodology, Data curation, Validation, Writing -- original draft, Writing -- review \& editing. \\ 
E.R.H.F.: Conceptualization, Validation, Writing -- review \& editing. \\ 
A.H.: Methodology, Validation, Writing -- review \& editing. \\ 
M.G.M.: Methodology, Validation, Writing -- review \& editing. \\ 
A.S.: Conceptualization, Funding acquisition, Project administration, Supervision, Writing -- review \& editing. \\ 
S.J.S.: Methodology, Software, Validation, Writing -- review \& editing. \\ 
C.S.: Conceptualization, Methodology, Project administration, Supervision, Writing -- original draft, Writing -- review \& editing. \\ 

\section*{Competing Interests}

The authors declare no competing interests.

\FloatBarrier
\bibliographystyle{unsrtnat}
\bibliography{refs}

\clearpage
\FloatBarrier
\setcounter{section}{0}
\setcounter{figure}{0}
\setcounter{table}{0}
\setcounter{equation}{0}
\renewcommand{\thesection}{S\arabic{section}}
\renewcommand{\thefigure}{S\arabic{figure}}
\renewcommand{\thetable}{S\arabic{table}}
\renewcommand{\theequation}{S\arabic{equation}}
\renewcommand{\theHsection}{S\arabic{section}}
\renewcommand{\theHfigure}{S\arabic{figure}}
\renewcommand{\theHtable}{S\arabic{table}}
\renewcommand{\theHequation}{S\arabic{equation}}
\begin{center}
{\LARGE\bfseries Supplementary Information}\\[1ex]
{\large Speed-to-Power Dominates Energy Costs for AI Data Centers: Technoeconomic Analysis of Powering AI in the Continental United States}
\end{center}
\vspace{1em}
\section{Comprehensive SI Methods}

We evaluate microgrid feasibility through technoeconomic simulation of complete power systems serving identical datacenter loads. We apply a common operational scenario: a specific number of GPUs requiring continuous power at a target location, then determine the minimum-levelized-cost configuration \text{within} each power architecture class that can meet this demand at required minimum annual uptime across the lifespan of the facility.

The analysis proceeds in three stages. First, we establish the hourly facility load that any power system must serve, combining IT demand with location-specific cooling requirements. Second, we simulate candidate power systems meeting this load profile, tracking hourly energy flows from generation through storage and conversion to final consumption. Third, we search each architecture's design space to identify configurations that minimize levelized cost while delivering required reliability.

This framework enforces scenario equivalence: all power systems face identical hourly demands across the same 25-year evaluation period at the same geographic location. We evaluate four candidate power systems: AC-coupled solar+storage, DC-coupled solar+storage, natural gas with backup diesel, and interconnection to the local electric grid as a baseline.

To establish this scenario, we accept location (lat/long coordinates), required annual facility uptime, and quantity of GPUs as specified inputs. Location-specific climate data are taken from typical meteorological year (TMY) sourced from the National Solar Resource Database (NSRDB) public API. TMY datasets are statistically representative hourly weather sequences derived from multi-decade observations that capture seasonal and diurnal patterns while excluding extreme outliers.

We compare each system in simulation year 27, end-of-life for the PV modules (25 year lifespan), with a base-case discount rate of 7\%. For scenarios with BESS, we assume one single end-of-life replacement of the battery modules occurring at the conclusion of operational year 13 (see the battery degradation subsection for more detail). All financial values are in 2022 dollars, and all discounting occurs mid year. 

All software is implemented in Python, and all tools are available from the public github repository at https://github.com/acnewkirk/AI-Datacenter-Microgrid-Analysis/tree/main. Our work builds upon two existing open-source tools: NLR's BLAST-Lite optimization framework \cite{gasper_nrelblast-lite_2025} and the cost calculator developed by OffGrid AI \cite{james_offgridai-uscost-calculator_2025}. Each provided critical starting points for the development of this work. To conduct the geographic analysis, we generated a hexgrid tiling of the continental United States, encompassing 4,317 individual hexagons. For each hexagon, we performed a full simulation at the center point of the hexagon for each microgrid architecture, assigning the resulting values to all points within that tile.

The following section begins with how we construct the IT facility load profile. We then detail how we design and simulate each of our candidate power system. We conclude with the techno-economic assessment and system optimization approaches.

\subsection{Facility Modeling}

Our facility model builds from the IT out, following established bottom-up datacenter modeling approaches \cite{koomey_estimating_2007, shehabi_2024}. We base IT demand on 8-GPU Nvidia H100 DGX nodes, drawing from empirical characterization of these systems during training workloads \cite{newkirk_empirically-calibrated_2025}. Each node exhibits average power draw of 7.3 kW and maximum draw of 8.5 kW, inclusive of memory and interconnect components. The load encompasses compute nodes, embedded networking, and management systems, but excludes cooling and power conversion losses addressed in subsequent sections.

We model aggregate hourly AI training load as baseload demand with minimal variation (±8\% random noise around unity), representing 8,760 hours of continuous operation \cite{grahovacDataCenterIndustrial2026}. This scenario can be conceptualized as one continuous training run involving all facility nodes, or alternatively as a facility with mixed workloads performing jobs under optimal scheduling and saturating demand. This training load shape is an idealization of practical facility operations, but serves as a conservative limiting case.  The framework accepts alternative load profiles and hardware specifications through modular inputs: node-level power parameters (average and maximum draw), hourly demand patterns, and total GPU count. While our current implementation assumes uniform load distribution across all nodes, the modular architecture readily supports extensions for heterogeneous fleet modeling.

Facilities with a more strongly diurnal usage pattern or greater time flexibility in execution have shown potential in recent work to employ demand response for cost savings and effective capacity expansion \cite{norris_rethinking_2025, colangelo_turning_2025}. Our modeled scenario provides a conservative bound on microgrid viability. Systems that perform well under inflexible baseload will show improved economic performance with operational flexibility under relaxed constraints.

We allocate the chip-quantity input into 8-GPU servers, that each have the power ranges described above. We construct an expected facility IT load by taking the product of our baseload demand time series, the average node power, and the quantity of nodes. This represents the expected power that must be delivered to the IT hardware in hour. We define the facility modeled maximum IT power demand as the sum of the node maxima. Datacenter operators typically contract with utilities for capacity that reflects the maximum theoretical facility demand. Similarly, power delivery infrastructure, both within and outside the fenceline, is scoped to accommodate this upper-bound value, even though average demand is much lower. This is a common feature of power systems design; infrastructure is built for peak demand, but then amortized over the typically lower level of average utilization.

The peak-provisioning principle fundamentally shapes the economics of both grid-connected and islanded facilities. To meet peak loads, operators must invest in infrastructure capable of handling maximum theoretical demand, even if this capacity sits idle during normal operations. We therefore define a facility design load representing the maximum instantaneous power the infrastructure must deliver:
\begin{equation}
P_{\text{design}} = N_{\text{nodes}} \times P_{\text{node,max}} \times \text{PUE}_{99} \times \alpha_{\text{contingency}}
\end{equation}
where $N_{\text{nodes}}$ is the number of compute nodes, $P_{\text{node,max}}$ is the maximum node power (8.5 kW), $\text{PUE}_{99}$ is the 99th percentile PUE value, and $\alpha_{\text{contingency}} = 1.05$ represents an engineering margin for component failures. This design load, distinct from average operational load, establishes power system sizing requirements that drive capital costs across all architectures.

\subsubsection{Cooling}

Datacenter cooling systems must reject heat equal to IT power consumption plus infrastructure losses, with efficiency varying substantially by technology and climate. Lei and Masanet found that the energy consumption of a given cooling technology can vary by up to 60\% location to location\cite{lei_climate_2022}. We model location-specific cooling efficiency using hourly PUE values, capturing how ambient conditions affect the availability of free cooling and mechanical system performance throughout the year.

Cooling system performance varies with climate: mechanical-only systems show low geographic variation while free-cooling systems exhibit PUE ranges from 1.05 to 1.4 depending on ambient conditions. The rising heat density of AI chips increasingly favors liquid cooling systems, both direct-to-chip and immersion \cite{narayanan_investigation_2024,kong_enhancing_2024,beberide_performance_2024,say_meta_2023}, with future server configurations requiring such systems \cite{connatser_nvidia_2024}. Following industry practice for high-density deployments, we exclude configurations with insufficient cooling capacity or prohibitively costly low performance.

We consider six candidate cooling configurations for our modeled facility: (1) IT liquid cooling with waterside economizer; (2) IT liquid cooling with adiabatic-assist air-cooled chiller; (3) IT liquid cooling with dry-cooler air-cooled chiller; (4) air-cooled IT with air-cooled chiller, airside economizer, and adiabatic cooling; (5) air-cooled IT with water-cooled chiller, airside economizer, and adiabatic cooling; and (6) air-cooled IT with air-cooled chiller and airside economizer. These systems align with the cooling distribution modeled for AI-specialized facilities by \cite{shehabi_2024}. Each system's cooling efficiency and capacity depends on local climatological conditions, specifically dry-bulb temperature and ambient humidity.

We construct an hourly time-series of location-specific climate and weather data relevant to cooling performance (dry-bulb temperature and ambient humidity) from the  TMY for the input location. Using Lei and Masanet's empirically calibrated datacenter cooling system model \cite{lei_climate_2022}, we pre-compute PUE values for each cooling technology across the range of ambient conditions, creating lookup tables indexed by dry-bulb temperature and humidity. We then build hourly PUE performance profiles of the candidate configurations for the specified location. We assume firms select the cooling configuration with the lowest average PUE value over the year because operational energy use dominates total-cost-of-ownership in cooling infrastructure \cite{cho_economic_2017}.. We acknowledge that in practice, firms select cooling according to a variety of decision variables, including familiarity, vendor relationships, and compatibility with existing staff expertise and capital equipment. We then combine the hourly IT-load with this hourly PUE to construct an hourly facility load.

The integration of cooling load illustrates the core design principle of \textit{scenario equivalence}. For a specified quantity of chips and location, the amount of power that must be delivered to the IT and cooling load, and the minimum uptime that design must deliver that power, are held constant in each modeled scenario. Differences in power delivery efficiency affect how much power must be generated at the source to meet that load, and the associated economics, but \textit{the energy required by the IT and cooling} are constant. 

Power demanded by the IT load is the hourly value defined in the previous section, and infrastructure power demand is that value multiplied by the difference between the hourly PUE value and 1. Recall that power infrastructure, both in generation and delivery, is scoped not for the annual average, but for the theoretical maximum. Once we've selected a cooling system, we select a design PUE value equal to the 99th percentile PUE produced by the selected system in the modeled year, consistent with industry practice. We then take our maximum IT load, multiply it by this design PUE, and then multiply this product by a scalar design contingency factor $\alpha_{\text{contingency}}$ , 1.05 in our base case. This factor corresponds to an engineering margin accounting for unanticipated component failure or malfunctions. This facility design load is the power delivery minimum of our candidate designs for a given facility; any generating system must be able to deliver at least this much power to IT and cooling system. 

\subsubsection{Power Systems Architecture}

To compare different microgrid systems, we develop candidate power flow models of all paths between generation and facility loads. We address a gap in the existing data center literature by modeling the explicit power distribution system losses of different  facility and microgrid architectures. A scenario assumption is colocated generating assets, though the power flow models could be extended to include transmission. We model the individual voltage and current conversion steps within the facility fenceline. Baseline candidate designs are shown in main-text Fig.~\ref{fig:powerflow}, with the LV topologies shown in \ref{fig:lv_powerflow}. Note that we assume islanded natural gas facilities adopt the same power systems architecture as conventional grid-connected facilities, and that the diesel backup generators connect to the facility through the same main transformer. Sensitivity analysis shows that the power-flow pathway of the backup gensets is not significant to facility economic performance due to low annual utilization, so we universally adopt this simplifying assumption.

We model all power systems out from the main LV distribution bus, with the cumulative efficiency of all voltage and conversion steps necessary to each load and from each source. We calculate path efficiency as the aggregate product of all component efficiencies between the bus and the source or load. The component efficiencies, displayed in main-text Fig.~\ref{fig:powerflow}, are based on component datasheets \cite{beheshti_wide-bandgap_2020} and operational estimates from industry experts \cite{pires_dc_2023}. We discuss optimum-design levelized-cost sensitivity to our choice of component efficiency in the technical parameter sensitivity section of the SI.

Our proposed power-flow paths are consistent with conventional data center design practice \cite{chalise_data_2015, barroso_datacenter_2019} or existing all-DC pilots \cite{tschudi_dc_2008, hiroseDCPoweredData2018}. We assume that the grid baseline and natural gas microgrid follow identical power delivery paths, and adopt the same power-delivery architecture for each, considered from within the facility fenceline. In this baseline power-path, the backup diesel generators are modeled as entering the facility through the same distribution transformer as primary power, and both cooling and IT load are supplied through double-conversion UPS to ensure consistent power quality, an architecture consistent with critical operation baseload facilities \cite{sartor_user_2021}. Even if physically collocated, the natural gas turbines do not naively output facility bus voltage, and the power must be gathered and transmit to loads. Exact output voltage is turbine and configuration specific, but is most commonly 13.8 kV for 60 Hz AC.  We model a utility step-up transformer which changes voltage for distribution within the on a MV spine at 34.5 kV AC, and then a step-down transformer which then switches either grid or natural gas power down to data hall distribution voltage. 

We model PV+Storage systems as using battery storage to provide both facility-scale backup power and uninterruptible power services. These architectures replace diesel backup generators and standalone UPS components with grid-scale Lithium-Iron-Phosphate (LFP) battery modules. Our power system designs follow typical microgrid models that utilize centralized generation and storage. Distributed generation and storage is also viable alternative proven in other applications \cite{gerber_cost_2023}, but is not considered in our baseline case. Component efficiencies are based on the averages of efficiency curve data from \cite{gerber_comprehensive_2021, gerber_simulation-based_2018}. We consider both low-voltage (LV) power distribution and medium-voltage (MV) distribution spine with individual data halls as LV spurs off of this MV bus.

The most efficient distribution topology keeps all assets on a single low-voltage bus: PV, BESS, IT, and cooling co-located, with no MV gathering stage and its attendant step-up/step-down conversions. We model this as a modular containerized design. This LV module would be ampacity bound: the maximum current a single LV bus and its protection can carry, and the maximum wiring length over which PV strings can be gathered at acceptable resistive loss. Taking the  800~V~DC data-hall consistent with the emerging OCP standard and a representative per-feeder continuous ampacity of 1{,}250~A (values consistent with commercial off the shelf cabling), a single LV bus carries $800~\text{V} \times 1{,}250~\text{A} \approx 1~\text{MW}$ of combined IT and cooling load with its proportionate share of co-located PV and BESS. In the MV cases, the embedded battery sits on its own busway off the switchgear, sized to deliver up to 1 MW of discharge power to the load it backs up; charging is routed through the main switchgear, whose solid-state protection is rated to 2 MW (consistent with the 2.5 kA interrupting capability of available DC breakers). No individual LV channel carries more than 1 MW.

In the low-voltage AC microgrid system, we model a ground-mount solar array with outputs combined through 500~kW central inverters that output three-phase 480~V~AC. The storage is a set of centralized battery banks within close proximity, AC-coupled to the switchgear though grid-forming 500~kW battery inverters. Wires are sized to minimize wire loss within reasonable cost. In the medium voltage case, we model 4 MW centralized inverters that feed the step-up transformer. The AC power supply unit (PSU) is a Platinum-rated 10.2~kW 240~V AC to 48~V DC power-factor correction (PFC) rectifier, integrated into each Nvidia DGX H100 node as four plus two redundant 3300~W rectifier modules. The integrated chip-level power electronics then switch the PSU voltage down to the internal chip voltages The cooling load (HVAC, chiller, and pumps) all high performance AC induction or brushless DC motors, powered by an AC variable speed drive (VSD) or a DC electronic speed controller (ESC). Here, we only model the losses of the AC/DC PFC rectifier stage within the VSD, as the DC/AC motor inverter stage is present in all topologies. Power quality and any training induced power transients in the AC systems are managed by the UPS or the BESS components. 

The low-voltage DC microgrid topology mirrors the LV AC microgrid. Here we model the solar panels connected with high-power DC/DC power optimizers to DC switchgear. The battery's DC/DC charge controller handles grid-forming \cite{gerber_comprehensive_2021} through droop-voltage regulation and coordinates with the microgrid energy management system (MEMS). It distributes power on a $\pm$400~V~DC bus. Note that some DC topologies directly connect the battery to the DC bus, but these topologies are out of scope for this study. DC/DC PSUs are required to step the 400~V distribution voltage down to the 48~V of the GPU node. They do not need a rack-mount UPS, and DC VSDs can connect directly to the 400~V bus without need for DC/DC conversion.

A modular campus composed of tiles of these LV busses electrically isolated but interconnected through fiber optic and cooling distribution networks is physically realizable. Such a facility, however, would require hundreds of independently-managed pods, carrying an operational complexity burden. The layout of the campus, dispatching and monitoring across facility, and operations and maintenance would be a substantial undertaking over and above the design challenge. As wide bandgap power electronics continue to mature, we view purely LV systems as viable options for relatively small facilities composed of a few such pods. In light of operational complexity, we consider topologies with MV distribution spines as our baseline architectures. 

Given a MV facility, the BESS can occupy distinct positions in the network, trading conversion efficiency against equipment cost and service capability. In our baseline cases, we model centralized storage at a single node in the topology. Note this is an idealization: at facility scale these assets would not sit on one large bus but would be containerized as the scale demands and as power electronics allow. For the purpose of understanding their interface with the network, however, they can be conceptualized as dispatching from a single node. In the AC system the bulk of storage is DC-coupled at the PV field (sharing the PV inverter and charging upstream of it); in the all-DC system it is centralized on the MV-DC spine. In both, a smaller increment of order the nameplate of the load it protects ($\sim$1~MW per LV bus)-is distributed to the data-hall LV buses for local outage ride-through, and in the AC case for grid-forming and load reliability (the effect of this split on modeled cost is low, see sensitivity analysis). Physical factors bound how storage may be redistributed. In the AC topologies, collocating storage and generation to share the inverter requires the battery to be electrically adjacent to PV on the same DC bus, so a distributed PV field could face gathering ceiling on how much storage can be field-coupled. Other storage assets must sit off a spur or at the hall, each forfeiting the shared inverter and buying its own. LV distribution busses have little ampacity headroom for co-located charge/discharge power, and single solid-state DC breakers presently cap at $\sim$2.5~kA ($\sim$2~MW), so hall-sited storage is constrained to a small share of an islanded fleet that is itself many times facility load. 

Storage placed at the data hall provides services: local uninterruptible ride-through, fast transient and fault response local to the load, and, specific to the AC systems, grid-forming voltage and frequency support that must be held physically close to the IT to be effective. These services are real but hard to amortize, since their value accrues as avoided downtime and power-quality assurance rather than energy. We therefore treat the distributed hall increment as a modeled architectural feature whose cost we quantify in the sensitivity analysis while noting its benefits qualitatively. 

Our baseline MV cases are detailed in the maintext of the manuscript, but they are repeated here for convenience. Each use a medium-voltage spine to separate centralized generation and storage from the data-hall loads, with individual halls connected as low-voltage spurs; they differ in whether distribution is AC or DC. In the baseline AC microgrid, PV is gathered through central inverters with the bulk of storage DC-coupled at the field (sharing the PV inverter). Power is gathered from the panel strings at 1.5 kV DC, and switched to 1 kV to charge the battery through a bidirectional DC/DC converter. On the shared bus  power passes through grid-forming inverters and a step-up transformer onto a 34.5 kV AC spine before stepping down to 480 V three-phase AC at the hall. In the baseline all-DC microgrid, the 1500 V DC PV blocks are stepped up to ±10 kV for campus distribution with the bulk of storage centralized on the MV-DC spine, then stepped down to a ±400 V data-hall bus where the embedded battery's DC/DC controller provides grid-forming via droop regulation. In both cases a load-matched storage increment ($\sim$1 MW per LV bus) is sited at the data-hall buses for local ride-through, and the centralized storage is modeled as concentrated at a single node for the purpose of characterizing its network interface.

Here we provide some high-level design heuristics for each PV+BESS architecture, though note the value of ancillary services are not valorized in our core cost model. In the MV DC case, operators should place as much of their storage capacity within the LV data hall as ampacity limits allow. The fewer intermediating components between storage assets and critical loads within the distribution network, the more reliably that node can operate, as there are fewer points of failure to prevent delivery of backup power. Further, DC systems don't incur the marginal inverter cost penalty (or more accurately, battery converter and management system costs are borne irrespective of site within the topology). Given that the performance of these systems are more sensitive cumulative storage-to-load path efficiency than to PV-to-storage efficiency (see sensitivity subsection), and given that the batteries can provide droop control in the LV hall, DC systems improve in performance as share of batteries on the LV bus increases. 

In the AC case, heuristics differ due to marginal inverter costs. PV and BESS are preferably collocated on the same DC bus to share the same inverter. This represents the maximally PV-to-storage path as it avoids any switching losses. It can also avoid inverter clipping losses, allowing otherwise clipped power to be routed to battery charging. In our modeled designs, such savings are marginal to trivial in terms of total system performance. the primary cost lever is thus inverter requirements. While large scale utility inverters on the order of 4 MW can be jointly shared by the panels and storage, once located elsewhere in the network standalone BESS require their own marginal inverters on top of those still required by the generating strings. The grid forming, LV hall level reliability, and power quality benefits of storage assets in the LV hall are non-trivial, so the trade off with inverter costs will be facility specific. Facilities confident that field-side grid-forming holds across the MV spine, and in the distribution network's reliability for delivering backup power, should site as much storage as possible behind the main inverters; otherwise they should distribute storage into the data hall as needed to provide local grid-forming, reliability, and power-conditioning services.

For each pathway, we calculate cumulative efficiency as the product of all component efficiencies along that path. To determine hourly power requirement at each bus, we divide the IT and cooling loads by their respective bus-to-load efficiencies, accounting for conversion and distribution losses. Similarly, we calculate required generation by dividing bus power by the source-to-bus efficiency for each generation or storage asset. This power flow modeling establishes the core of the facility-specific design constraints: user-specified minimum uptime, location-aware hourly expected IT and cooling demand on the main distribution bus, facility design load for infrastructure scoping, and pathway-specific power delivery efficiencies. Note that we don't explicitly model supercapacitor banks to maintain bus voltage or any required fault management infrastructure like high efficiency circuit breakers. Full characterization of the costs and performance characteristics of these systems can be found in the sensitivity analysis section of the SI.

\subsection{Generation, Storage, and Degradation}

Having established facility demand and power-flow paths, we next model how each generation technology supplies that demand over the project lifetime. Renewable systems must buffer resource intermittency with storage, whereas thermal plants must ride through mechanical failures and maintenance outages. All technologies degrade over time, requiring accurate modeling throughout the evaluation period.

We compare all architectures at year 27,the planned decommissioning point of the solar systems, so that capital recovery periods are aligned even though batteries undergo mid-life replacement and gas turbines operate through continuous maintenance cycles. Using location-specific TMY data, we model hourly solar irradiance and ambient temperature effects on all equipment: PV generation, battery efficiency, and gas turbine thermal derating. Solar-storage systems are simulated with hourly dispatch to capture battery cycling and multi-day weather events. We model natural-gas plants with a probabilistic availability model that reflects forced outages and scheduled maintenance. The following subsections describe the generation, storage, and degradation models in detail.

\subsubsection{PV and BESS}

We adopt the vectorized hourly energy balance approach from \cite{james_offgridai-uscost-calculator_2025, baranko_fast_2024}, implemented with a bus-centric architecture that explicitly models conversion losses through different power-flow paths. The PV generation model fetches TMY weather data (irradiance and temperature) for the specified coordinates and feeds these into pvlib's ModelChain to calculate DC power output. The ModelChain simulates a reference PV module equipped with single-axis tracking, incorporating the CEC temperature model for thermal derating and accounting for 14\% system losses from soiling, shading, mismatch, wiring, and light-induced degradation. This simulation runs across all 8760 hours of the year to produce raw DC power values. We then normalize these values by dividing each hourly output by the annual peak DC output, yielding a capacity factor profile ranging from 0 to 1. This can then be rescaled to any specified nameplate capacity. 

PV module output is translated to power delivered to the distribution bus with architecture-specific efficiency chains defined in the power-systems subsection.  For AC-coupled layouts the chain includes inverter clipping: the effective AC power output from the inverter, $P_\text{effective}$, is defined as the minimum of the efficiency-adjusted DC power produced by the PV modules and the inverter's nameplate AC output:
$
P_\text{effective} = \min(P_\text{DC} \times \eta_\text{inv}, P_\text{inverter})
$ where $P_\text{DC}$ is the instantaneous DC panel output, $\eta_\text{inv}$ is the inverter's conversion efficiency, and $P_\text{inverter}$ is the inverter's rated AC capacity. DC-coupled layouts bypass the inverter and use their own conversion stages. 

We assume the battery system is composed of 4-hour, 60 MW LFP modules, each rated to store 240 MWh of usable energy \cite{ramasamy_us_2022}. We selected LFP as our storage chemistry due to its excellent durability and ramp rate \cite{goodenough_li-ion_2013} and current commercial deployment in grid-scale storage applications. While alternate chemistries provide superior energy-density, in the stationary application we model space and weight are not key performance sensitivities. We adopt a conservative LFP round-trip-efficiency (RTE) of 90\% \cite{rehm_comparing_2025}, and initial battery state-of-energy as 75\% nameplate capacity. 

Battery capacity fades over time due to calendar and cycling effects, with charge-discharge patterns and depth of discharge affecting fade severity \cite{gasper_degradation_2023, crawford_lifecycle_2018, xiong_multi-factor_2025}. Battery management systems protect cell longevity by restricting operation to safe voltage ranges, typically utilizing 80-90\% of theoretical cell capacity. Our model treats the manufacturer's nameplate capacity as representing this BEMS-accessible energy window; when we model a 240 MWh battery at 100\% state-of-charge, this corresponds to the maximum energy the BEMS permits, not the theoretical cell maximum. This nameplate capacity degrades over time as discussed in subsequent sections.

The hourly dispatch algorithm prioritizes serving facility load from solar generation. Surplus solar charges the battery (subject to power rating and available headroom), with remainder curtailed. Deficits trigger battery discharge (subject to power rating and stored energy). The simulation implements:
\begin{equation}
\Delta SOE_t = \begin{cases}
P_{\text{surplus},t} \cdot \Delta t / (\eta_{\text{charge}} \cdot E_{\text{nom}}) & \text{if solar exceeds load} \\
-P_{\text{deficit},t} \cdot \Delta t \cdot \eta_{\text{discharge}} / E_{\text{nom}} & \text{otherwise}
\end{cases}
\end{equation}
Subject to $0 \leq SOE_t \leq 1$ and $|P_{\text{batt}}| \leq P_{\text{batt,max}}$. The simulation tracks annual cycles, throughput, curtailment, and unmet load for economic analysis.

\begin{figure*}[t]
\centering
\includegraphics[width=\textwidth]{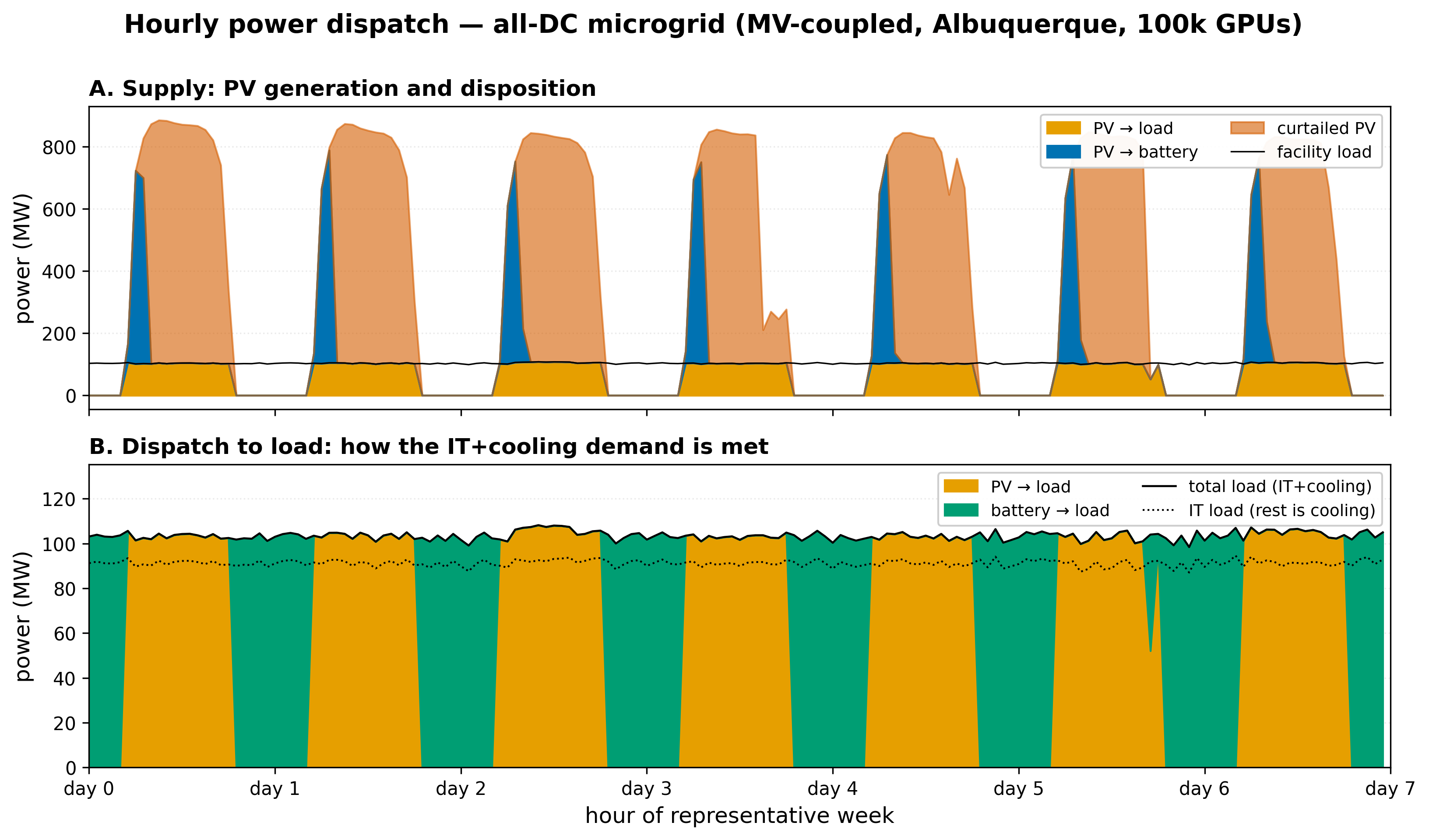}
\caption{\textbf{Hourly power dispatch for the all-DC microgrid over a representative week} (MV-coupled, Albuquerque, 100{,}000 GPUs). \textit{(A)} PV generation and its disposition: power delivered directly to load (amber), used to charge the battery (blue), and curtailed (vermilion), with the facility load drawn for scale. \textit{(B)} How the IT-plus-cooling load is met each hour, by direct PV (amber) and battery discharge (green); the solid line is total load and the dotted line the IT component, the difference being cooling. The eight-fold gap between peak PV and facility load, the heavy midday curtailment, and the full overnight reliance on storage are characteristic of an energy- rather than power-constrained islanded design, and underlie the low-stress battery duty cycle documented in the degradation subsection.}
\label{fig:dispatch}
\end{figure*}

Figure~\ref{fig:dispatch} traces this dispatch over a representative summer week for the optimized all-DC design at our 100{,}000-GPU Albuquerque reference site, and makes the energy-constrained character of the islanded design concrete. Peak PV generation exceeds the facility load roughly eight-fold; the battery charges to full within the first hours of sun, the bulk of midday generation is curtailed, and overnight the battery alone carries the load. The gap between the total-load and IT-load traces is the cooling overhead.

\subsubsection{Natural Gas}

Natural gas power plant design fundamentally differs from renewable energy system optimization due to its non-linear solution space: multiple interacting discontinuities that preclude simple scaling from reference designs. A gigawatt-scale combined cycle plant represents a fundamentally different engineering solution than what a 100 MW datacenter requires—not merely smaller, but potentially employing a different turbine class, cycle configuration, and redundancy architecture altogether. Turbine selection is inherently discrete, with aeroderivative, F-class, and H-class machines each occupying distinct niches in the capacity-efficiency-reliability design space. The number of units introduces step changes in redundancy and operational flexibility, while the choice between simple and combined cycle configurations fundamentally alters project economics: combined cycle plants approach 60\% thermodynamic efficiency but require higher upfront capital investment and extended construction timelines. Our approach addresses these interdependencies by systematically generating candidate configurations from commercially available turbines, applying engineering constraints derived from industry practice, and evaluating the full lifecycle economics of each specific configuration.

Our design tool proceeds from a library of reference natural gas turbines, comprised of 16 individual turbine models, subset according to their manufacturer classification: aeroderivative, F-class, and H-class. These classifications correspond to turbine characteristics: Aeroderivative turbines are flexible but low-capacity and comparatively inefficient; class H turbines have the highest capacity and fuel-efficiency, but are the most complex to manufacture with high upfront costs; F-class turbines are medium-weight, medium capacity frames most often used in industrial applications \cite{darrow_catalogue_2017}. For each turbine model, we source nameplate capacity, combustion efficiency, and annual availability from manufacturer specifications and turbine data sheets. 

We model the operations of these turbines according to the following formulations. Note that specific parameter values are available in the configuration file of the codebase and the SI. The effective power output and efficiency of any given gas turbine configuration are determined by its class, $c$, from the set $C = \{ \text{aeroderivative, F-class, H-class} \}$. Performance is a function of scheduled maintenance, operational load, and ambient environmental conditions, which we model as follows:

Periodic maintenance is modeled as a deterministic reduction in a plant's availability. The Maintenance planned Outage, $M_{c}$, for a turbine of class $c \in C$ is defined as the ratio of its required annual maintenance period, $H_{c}$ (in hours), to the total hours in a year, $y_{year}$. 

\begin{equation}
M_{c} = \frac{H_{c}}{y_{year}}
\label{eq:pof}
\end{equation}

A turbine's thermodynamic efficiency degrades when operating below its nameplate capacity. The adjusted operational efficiency, $\eta_{op}$, is the product of the turbine's rated efficiency, $\eta_{rated}$, and a class-dependent part-load multiplier, $f_{\text{part-load}}(\beta_{op}, c)$. This multiplier is a function of the plant's operational load factor, $\beta_{op}$, which is the ratio of average instantaneous power to the plant's total nameplate capacity. 

\begin{equation}
\eta_{op} = \eta_{rated} \times f_{\text{part-load}}(\beta, c)
\label{eq:part_load}
\end{equation}

The part-load multiplier function is based on empirical analysis of turbine performance curves \cite{munoz_de_escalona_part-load_2012} and is defined as:
\begin{equation}
f_{\text{part-load}}(\beta, c) =
\begin{cases}
1.0 - 0.10(1 - \beta) & \text{if } c = \text{aeroderivative} \\
1.0 - 0.15(1 - \beta) & \text{if } c = \text{F-class} \\
1.0 - 0.20(1 - \beta) & \text{if } c = \text{H-class}
\end{cases}
\label{eq:part_load_cases}
\end{equation}

A turbine's maximum power output degrades with higher intake air temperature due to lower density. The available power, $P_{avail}$, is determined by derating the turbine's nameplate capacity, $P_{rated}$, based on the ambient temperature, $T_{amb}$. The derating is governed by a class-dependent factor \cite{de_sa_gas_2011}, $r_{\text{derate}}(c)$, and applies only for temperatures above the ISO standard of $T_{ISO} = 15^\circ\text{C}$.

\begin{equation}
P_{avail} = P_{rated} \times \left[ 1 - r_{\text{derate}}(c) \times \max(0, T_{amb} - T_{ISO}) \right]
\label{eq:temp_derate}
\end{equation}

For both capacity derating and part-load efficiency penalty, we selected conservative estimates, and varied the values in sensitivity analysis. 

The total power of a combined cycle plant, $P_{CC}$, is calculated by augmenting the gas turbine's power, $P_{GT}$, with additional power generated from its waste heat. The power gain depends on the heat of the exhaust gas, which depends on the gas turbine's efficiency, $\eta_{GT}$, the effectiveness of the heat recovery steam generator, $\epsilon_{HRSG}$ (assumed to be 80\%), and the efficiency of the steam turbine cycle, $\eta_{steam}$ (assumed to be 34\%) \cite{ganapathy_heat-recovery_1996}.

\begin{equation}
P_{CC} = P_{GT} \left[ 1 + \left(\frac{1}{\eta_{GT}} - 1\right) \times \epsilon_{HRSG} \times \eta_{steam} \right]
\label{eq:cc_power}
\end{equation}

Natural gas power plants require a variety of active electrical components to operate, varying from motors to run compressors or cooling water pumps to facility lighting. These loads, collectively termed "auxiliary loads" or less commonly "house loads", draw parasitic power generated by the gas turbine. Importantly what components are necessary for power operation and their cumulative share of nominal turbine output differs by plant configuration \cite{gulenImportanceAuxiliaryPower2010} . We adopt a configuration specific auxiliary load empirically calibrated to EIA reference plant values \cite{mirletz_annual_2024}

We derive plant performance according to these relationships, with a candidate configurations completely and uniquely determined by: turbine model, fleet size, and single-or-combined cycle. We restrict the design space using industry derived engineering filters to exclude infeasible and implausible configurations.

The most basic is capacity filtering: a plant must be able to meet the facility's peak demand $P_{req}$, defined as $P_{\text{design}}$ scaled by cumulative power system losses, even under adverse environmental conditions. We enforce this by requiring that the plant's total capacity, after being derated for the site's 99th percentile design temperature, be greater than or equal to $P_{req}$.

Beyond this physical requirement, we constrain the plant's annual expected operational load factor  ($\beta_{op}$) to a window of 30-90\%. The 30\% floor reflects a conservative minimum stable output level where turbine manufacturer specifications remain reliable. The 90\% ceiling enforces an \textasciitilde11\% planning reserve margin, which provides the necessary headroom to handle demand fluctuations or unexpected outages. This margin is a slight relaxation of the 15\% standard common in utility planning, reflecting the stable demand profile of the modeled data center. 

We impose architectural constraints to ensure that designs reflect commercially viable and operationally practical power plants. First, each plant configuration is limited to a single turbine model, precluding mixed-turbine fleets to reflect the significant logistical and operational challenges of mixed-asset plants. Second, aeroderivative turbines are restricted to simple cycle configurations due to the unfavorable economies of scale of the required heat recovery systems. Finally, to reflect practical limits on plant complexity, the model constrains the number of units in a fleet. This heuristic, based on industry consultation, allows for a greater number of units in fleets of smaller turbines, such as aeroderivatives and small F-class models, than in fleets composed of large industrial frames.

Finally, a more stringent \textit{N-1 redundancy} filter can be optionally enabled. This standard reliability constraint requires that the plant can still meet the peak demand $P_{req}$ even with its largest single unit offline. This check is performed using the temperature-derated capacity to ensure N-1 sufficiency during hot weather conditions.

While the engineering filters ensure each plant configuration is robust, we calculate the Expected Unserved Energy (EUE) to quantify the remaining reliability risk and to appropriately size the backup power system. EUE is a standard industry metric representing the expected annual energy shortfall from a given plant design due to both random (forced) and scheduled (planned) outages.

The total EUE is the sum of the forced and planned outage components, $EUE_{total} = EUE_{forced} + EUE_{planned}$. We calculate the forced outage component, $EUE_{forced}$, using a binomial probability model based on the single-unit forced availability, $A_f$. The expected annual shortfall is the sum of all potential outage states weighted by their probabilities:
\begin{equation}
EUE_{forced} = H_{year} \sum_{k=0}^{N-1} \binom{N}{k} A_f^k (1-A_f)^{N-k} \times \max(0, P_{avg} - k \cdot P_{unit})
\end{equation}

The planned outage component, $EUE_{planned}$, is the deterministic energy shortfall that arises when units are deliberately taken out of service for maintenance. For both forced and planned outages we compute expected unserved energy using the facility’s average demand. For planned outages this implicitly assumes maintenance is scheduled outside the hottest weeks. 

Note that EUE is not used as a direct optimization target in our methodology. The primary application of the calculated EUE value is to inform the sizing of the backup diesel system's fuel reserves. The fuel storage capacity must be sufficient to generate the energy equivalent of the gas plant's expected annual shortfall, plus a safety margin. This ensures the integrated system can meet its overall reliability target. EUE is an annual statistical expectation, not a chronological simulation. It does not provide information on the timing, frequency, or duration of individual outage events. It serves to size the total volume of backup energy required per year, but not the power capacity of the backup system, which is sized to meet the facility's peak demand ($P_{req}$) .

Unlike the probabilistic treatment of gas plant reliability, diesel backup sizing follows deterministic criteria standard in mission-critical facilities. The diesel system must be able to deliver the full facility demand $P_\text{req}$ regardless of the gas plant configuration, as any shortfall would violate the data center's uptime requirement. We enforce an N+1 redundancy architecture where the total installed fleet includes one additional unit for redundancy. This N+1 configuration ensures the diesel plant can meet full load even with one generator offline for maintenance or failure. Diesel backup system design rules reflect guidance for critical backup power infrastructure \cite{ericson_comparison_2019, mott_best_2021}, with those sources also providing our diesel system financial parameters.

While diesel generators provide the power capacity for backup, their runtime is constrained by on-site fuel storage. We size fuel reserves based on the gas plant's Expected Unserved Energy, applying a safety factor to account for uncertainty in outage patterns:
\begin{equation}
V_\text{fuel} = \max\left( \lambda \cdot \text{EUE}_\text{total} \cdot \text{SFC}, V_\text{min} \right)
\end{equation}
where $V_\text{fuel}$ is the fuel storage volume (gallons), $\lambda$ is a safety factor (typically 2.0), $\text{SFC}$ is the specific fuel consumption (gallons/MWh), and $V_\text{min}$ represents the minimum code-required storage. The minimum storage ensures 72 hours of runtime at full load, a standard requirement for critical facilities. This dual criterion ensures adequate fuel for both statistical expectations and exogenous scenarios such as extended natural disasters or fuel delivery disruptions. 

Note that the tool includes another optional engineering filter that we do not enforce in our base analysis: maximum diesel turbine runtime. Operators are typically only permitted to operate backup power turbines up to some annual limit (typically 200-500 hours, varying by region) per annum due to air quality standards. Above this threshold, operators need to register as prime power providers, with higher operational scrutiny. We elect to not enforce this filter in our base case as the facility in our scenario is a primary power provider, and would thus likely undergo full permitting and regulatory review. If they anticipate their design would require operating the diesel gensets above annual maximum, they would obtain permitting at that initial stage of the process. Users of the tool suite can enforce the flag to filter out designs that would exceed some specified annual max, with the practical impact of favoring multi-turbine configurations.  

The diesel backup system we do not independently optimize but rather size deterministically based on the selected gas plant configuration. This coupling means that gas plant reliability directly affects total system costs: a less reliable gas configuration reduces gas plant capital costs but increases both diesel fuel storage requirements and expected diesel consumption. The life-cycle cost optimization described in design optimization subsection captures these tradeoffs, selecting the gas-diesel combination that minimizes total cost while meeting reliability specification.

\subsubsection{Degradation}

Power-generating and energy-storage assets degrade over their useful life. Accurate degradation modeling is essential for economic comparison, as systems that appear cost-competitive at commissioning may diverge significantly in delivered energy costs over a 25-year horizon. We implement degradation models calibrated to empirical data where available, with conservative assumptions where data is limited.

For solar PV, we apply a linear degradation model with 1\% first-year loss, primarily from light-induced degradation, followed by 0.55\% annual degradation thereafter \cite{yang_comprehensive_2025}. These degradation values were validated through consultation with a utility solar system design expert. While module-specific and climate-dependent variations exist, the total uncertainty band over 25 years has minimal impact on LCOE due to heavy discounting of later years in financial analysis.

Natural gas turbines experience both capacity and efficiency degradation, with rates differentiated by turbine class to reflect their distinct operational characteristics and maintenance requirements. Properly maintained plants see dramatically lower degradation rates. Our model assumes realistic operating environments with regular maintenance interventions such as compressor washing, resulting in annualized degradation rates that align with industry experience of 0.5-1.5\% loss between major overhauls \cite{kurz_degradation_2000}. We differentiate by turbine class, with aeroderivative turbines experiencing higher degradation rates due to their higher pressure ratios and sensitivity to fouling, while F-class and H-class heavy-duty frames exhibit greater durability owing to their robust design and longer service intervals. Importantly, efficiency degradations in natural gas plants affect cost performance but not the system's ability to deliver SLA-guaranteed power, and capacity degradations remain much smaller than the capacity headroom required by our design rules. This distinction allows us to adopt simplified linear calendar approaches for both PV and gas turbine degradation modeling. See the sensitivity analysis for the discussion of the sensitivity to our choice of degradation factors. 

Lithium-ion battery degradation requires more sophisticated treatment. Battery capacity fade results from complex, interdependent electrochemical processes \cite{xiong_multi-factor_2025} that vary nonlinearly with operating conditions\cite{morin_quantifying_2025}. Depth of discharge (the fraction of battery capacity used in each cycle), C-rate (the charge/discharge rate relative to battery capacity), cell temperature, and state-of-charge (the battery's current energy level as a percentage of capacity) all influence degradation trajectories. Further, batteries experience both calendar aging (degradation over time regardless of use) and cycling stress (degradation from charge-discharge cycles), and these factors can mutually interact, introducing significant non-linearity into battery fade dynamics.

Battery capacity fade modeling employs BLAST-Lite, an open-source tool implementing semi-empirical degradation models developed by Smith et al. \cite{smith_lithium-ion_2021}. The tool's LFP model derives from multi-year cycling data of grid-scale batteries under representative duty cycles, making it directly applicable to our use case. BLAST-Lite tracks degradation through both pathways: calendar aging (time-dependent SEI growth and electrolyte decomposition) and cycle aging (mechanical stress and lithium inventory loss from charge-discharge cycles). The model accepts hourly time-series of power throughput, state-of-charge, and temperature, then computes capacity fade using stress-weighted composite of these mechanisms.

Blast-lite summarizes complex charge-discharge patterns into equivalent stress cycles. For our LFP chemistry, the model represents key empirical behaviors: minimal sensitivity to depth-of-discharge below 80\%, accelerated degradation at extreme SOC ranges (>95\% and <5\%), and Arrhenius-type temperature dependence with roughly doubled degradation rate per 10°C increase. These path-dependent calculations produce more accurate lifetime predictions than simplified equivalent-cycle or calendar-only models.

Our 25-year facility operating lifespan exceeds typical 15-year battery system operating lives, making replacement timing and strategy critical design decisions. While unconstrained optimization might minimize initial capacity and add cells annually to offset degradation, this approach misrepresents actual operational practices, as firms value operational simplicity over theoretical optimality. In theory, an operator could dramatically oversize the initial battery system capacity to provide, in expectation with modeled duty cycle, 25 years of operation above minimum uptime. Such a strategy would be maximally costly, in essence buying idle battery capacity that is undergoing calendar fade that will only see utilization in later years of facility operation. Alternatively, a facility might install the minimum scale battery system capable of achieving y0 uptime, with steady augmentation each year to replace faded capacity. Such a strategy, termed augmentation \cite{HowDecideOversizing}, does not align with data center industry prioritization of reliability and turnkey power systems operation. Existing deployments of large scale energy storage by data centers have been large scale installations with capacity overhang \cite{sayCrusoeSignsTwo2026, sayCrusoeRedwoodLaunch2026, WorldsLargestGrid}. 

\begin{figure}
    \centering
    \includegraphics[width=1\linewidth]{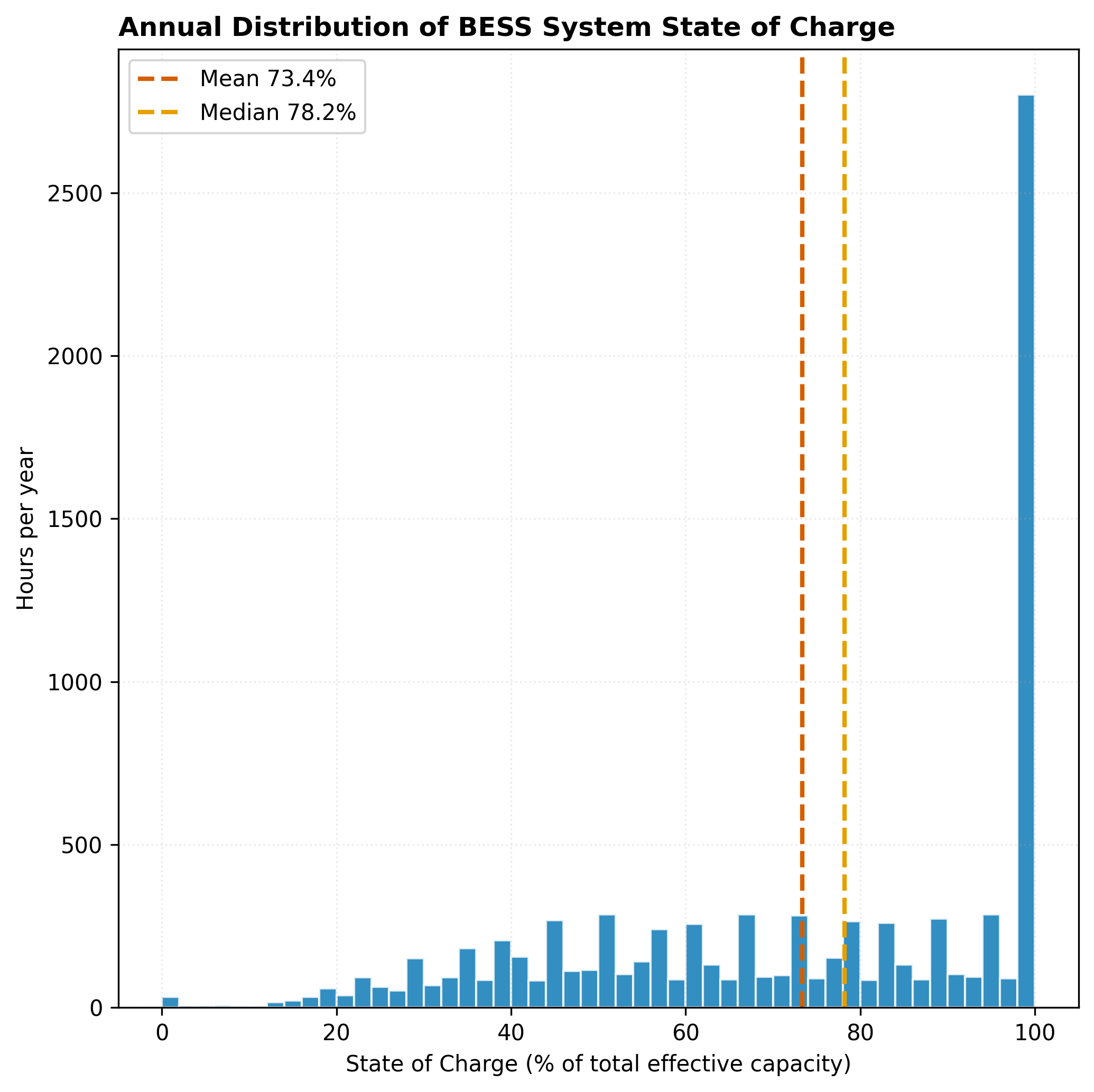}
    \caption{Distribution of BESS state of energy over the hours in simulation year 0 for a candidate DC coupled design. The modeled design is in southern California (34.47 N, -116.16 W), would supply 50,000 GPUs (facility design load of 65.3 MW) on a MVDC architecture though 447 MW solar PV and 264 MW of BESS capacity. Note the strong skew towards high states of energy, as storage systems are scoped for outlier events. In practice, alongside the practically modest discharge rate for steady overnight operations, configurations capable of high uptime exhibit comparatively low-stress cycling patterns.}
    \label{fig:soe}
\end{figure}

Overbuilt battery systems still require module replacement to ensure lifetime capacity at reasonable cost. Too early or late a refresh point and the firm is commits to low total usage of the storage assets for one period and greater overbuild for the other. We therefore implement a single full-scale battery replacement in year 13 as an approximation of system midlife, to reflect practical procurement strategies in which facilities execute major infrastructure upgrades as planned capital projects, not continuous optimization exercises. Note that we evaluate the sensitivity to the selection of replacement year in the technical parameter stability sensitivity analysis subsection. 

Battery thermal management follows a simplified physics-based approach that tracks temperature evolution throughout the year. We model heat generation from round-trip inefficiency as $Q_{gen} = P_{throughput} \times (1 - \sqrt{\eta_{RTE}})$ where throughput power and round-trip efficiency determine waste heat. The battery's thermal mass (2.3 kWh/°C per MWh capacity) buffers these temperature fluctuations. The thermal mass parameter was derived from a representative 3.9 MWh, 38-ton containerized BESS \cite{reber_role_2022}, calculating the system's total thermal capacitance using mass-weighted specific heat values of its components: 57\% LFP cells, 20\% steel, and 23\% other materials \cite{tang_review_2019, neupane_heat_2018, morganti_multi-scale_2019}.

Active cooling engages when cell temperature exceeds 30°C, with power requirements calculated as the excess heat divided by the cooling system coefficient of performance (derived from the facility PUE). We account for this parasitic load by reducing the facility's net energy delivery by an equivalent value to the cooling energy use. Though thermal loads could theoretically be scheduled to coincide with solar curtailment periods using batteries, we naively dispatch climate control any hour outside of thermal envelope. The model includes provisions for heating when batteries are idle in cold conditions (below 15°C), this rarely engages in practice. This approach captures the essential thermal dynamics and their impact on both degradation rates and parasitic loads without requiring more detailed modeling. 

\begin{figure}
    \centering
    \includegraphics[width=1\linewidth]{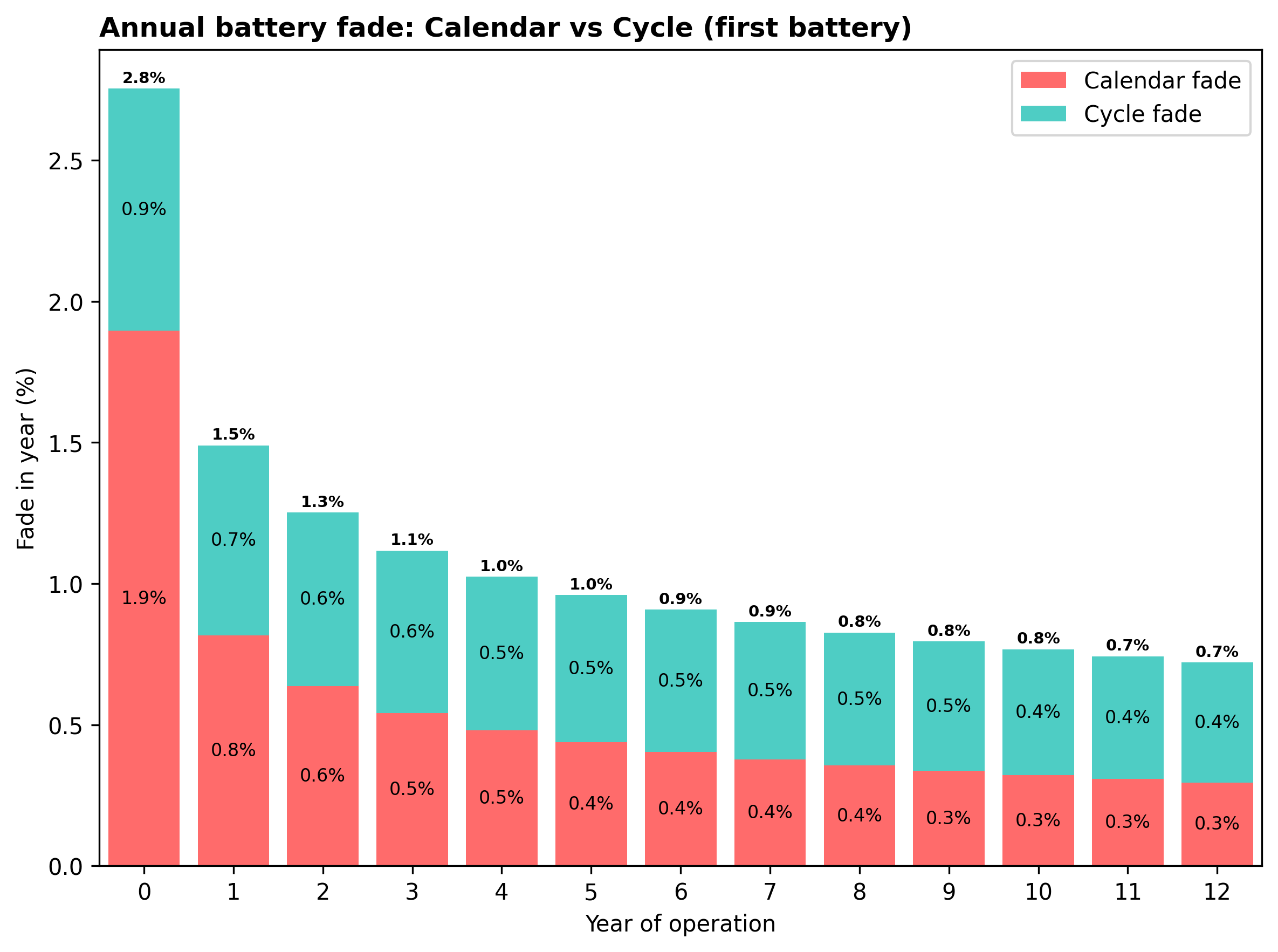}
    \caption{Cumulative fade from cycling and calendar effects over modelled lifespan of LFP battery system. The modeled design is in southern California (34.47 N, -116.16 W), would supply 50,000 GPUs (facility design load of 65.3 MW) on a MVDC architecture though 447 MW solar PV and 264 MW of BESS capacity. Cumulative fade is 12\% at replacement, with calendar fade contributing 51\% and cycle fade contributing 49\%. Note that cumulative fade and fade composition are both design specific so will differ system-to-system. }
    \label{fig:fades}
\end{figure}

We conducted initial exploratory data analysis on candidate systems using Blast-lite, revealing characteristics of our usage pattern. Batteries in systems that deliver greater than 90\% annualized uptime show only modest fade at their parameterized retirement year, averaging 11\% cumulative fade with roughly even total cycle and calendar fade, as shown in \ref{fig:fades}. Because batteries are sufficient for overnight operations, and because solar panels are sufficient to both deliver facility load during daylight hours and charge the battery for expected overnight energy demand, battery state-of-energy is strongly skewed, with more than a quarter of typical hours spent at 100\% SoE, as shown in \ref{fig:soe}. Typical depth-of-discharge is relatedly modest.

The low battery health degradation we find is consistent with underlying theoretical and empirical results on LFP cycle fade \cite{lam_decade_2025, peterson_lithium-ion_2010}. Islanded PV-plus-BESS systems require significant nominal power overbuild to achieve 90\% annual uptime (to say nothing of 2 or 4 9s of reliability) because the sun does not shine at night. The design region is thus energy constrained, not power constrained. Beyond even the storage for a typical overnight load, the battery system must also be scoped to accommodate seasonal ebbs and outlier events.

Systems of this size charge and discharge only a small fraction of their total energy capacity in any given time step. Though the exact value varies design-to-design, the maximum nominal C-rate in a given year is approximately .2, with the average C-rate a third of that. This gentle nominal C-rate is consistent with a low-stress battery operating regime \cite{peterson_lithium-ion_2010}.This stressor pattern is also fairly stable year to year: annual throughput cycling increases on average only 11\% cumulatively over the life of the battery, mean SoC reduces by 4\% on average, and annual thermal load increasing less than 1\% on average.

A full Blast-lite call to evaluate every candidate design in the optimization loop would have been computationally intractable. The relatively stable fade regime of our batteries allowed us to address this through gray-box surrogate distillation. The surrogate follows a physics-informed machine-learning strategy: a low-order analytic “scaffold" expresses the dominant aging mechanisms, and a data-driven term learns only the small residual that remains between the scaffold and the simulator output.

We split degradation into calendar and cycle terms:
$$\Delta q_{\text{cal}} =
    A\,t^{\beta}
    \exp\!\Bigl(-\frac{E_{\text{cal}}}{RT}\Bigr)
    \bigl[1+\gamma\lvert\mathrm{SOC}-0.5\rvert\bigr],\qquad
\Delta q_{\text{cyc}} =
    B\,\mathrm{EFC}^{\alpha}
    \exp\!\Bigl(-\frac{E_{\text{cyc}}}{RT}\Bigr),$$
where $t$ is elapsed time in years, $\mathrm{EFC}$ is cumulative equivalent full cycles, $T$ is mean cell temperature in kelvin, and $R$ is the universal gas constant.  We locked the Arrhenius energies at the BLAST-Lite values $E_{\text{cal}} = 43.3\;\text{kJ mol}^{-1}$ and $E_{\text{cyc}} = -18.2\;\text{kJ mol}^{-1}$.  Starting from the BLAST-Lite exponents $\beta_{0}=0.526$ and $\alpha_{0}=0.828$, we let the data move them within ±0.25 and fit the scale factors $A, B$ plus the SOC-curvature coefficient $\gamma$.
To capture path dependence, we simulated every design through a complete life cycle  and recorded each year’s incremental fade.  After fitting the scaffold to those increments, we modeled the residuals
$$r = q_{\text{BLAST}} - \Delta q_{\text{scaffold}}$$
with a GP that uses a Matérn ($\nu=2.5$) kernel and a small white-noise term.  We assessed accuracy with five-fold grouped cross-validation, treating each design as one group to prevent data leakage.

We constructed a training set out of detailed Blast-lite degradation simulations of candidate systems in 13 locations over a full battery life cycle. These locations included at least one entry in every DOE climate zone in the continental United States, and we simulated 4 designs at each location. Each year's simulation provided stress features: state-of-charge statistics, temperature, and throughput metrics, creating a path dependent stressor history for each design. Group-aware 5-fold cross-validation ensured robustness across designs and allowed for hyperparameter optimization. We then evaluated the final distilled model performance on a new set of randomly generated designs from 10 locations explicitly not within the training set. 

\begin{figure}
    \centering
    \includegraphics[width=1\linewidth]{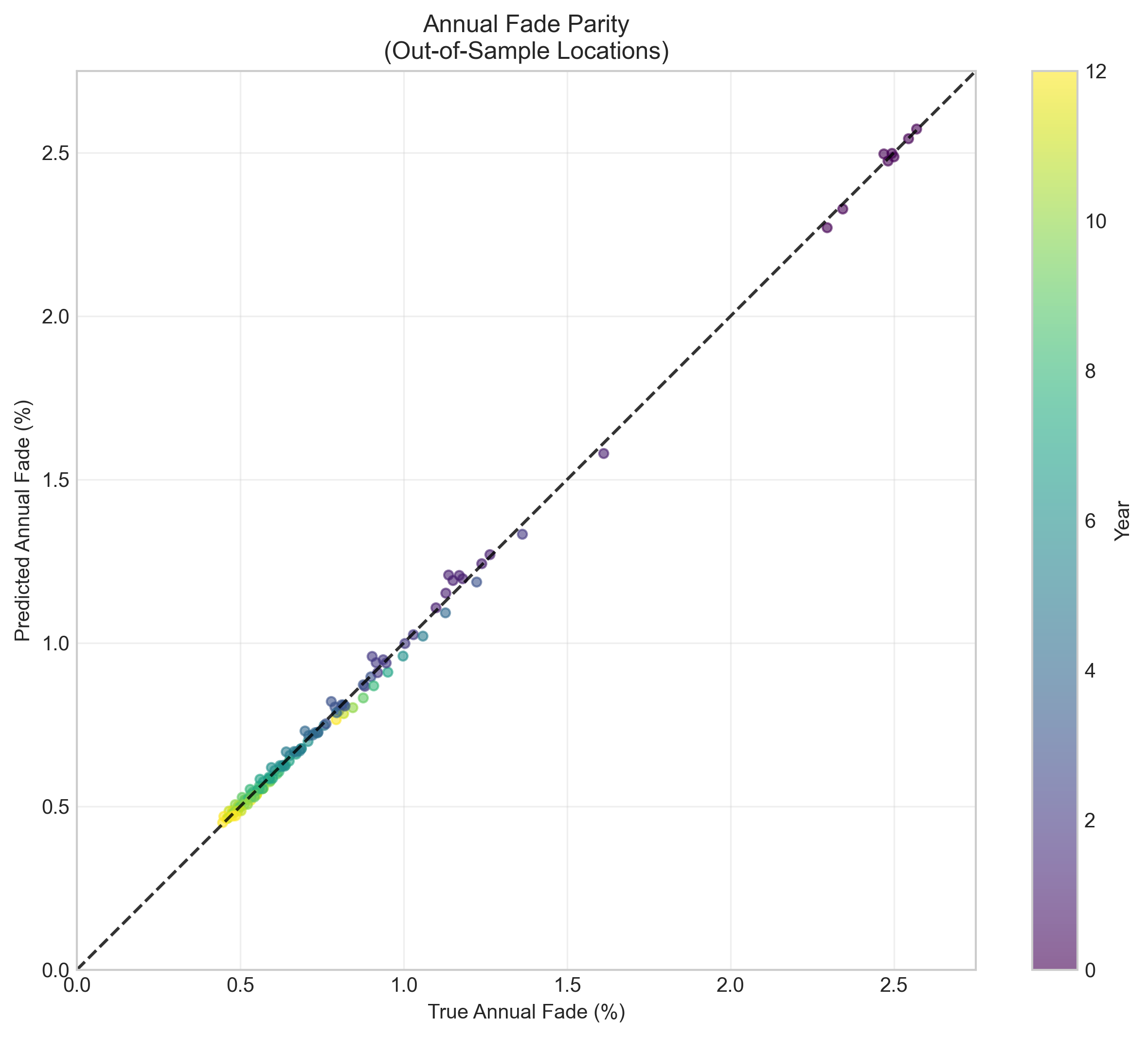}
    \caption{Out-of-sample validation of the gray-box battery fade surrogate model. The parity plot compares predicted annual fade rates (vertical axis) against ground-truth BLAST-Lite simulations (horizontal axis) for 10 geographic locations not included in the training data. Each point represents one operational year (0-12, color-coded) for one of the validation designs. The tight clustering along the diagonal (dashed line) demonstrates that the surrogate maintains predictive accuracy when extrapolating to new locations.}
    \label{fig:oos}
\end{figure}

The proxy showed high fidelity to the underlying Blast-Lite simulation: out-of-fold (OOF) mean absolute errors for calendar and cycle fade are 0.0016 and 0.0097 respectively, and each  had an R\textsuperscript{2} greater than 0.99. This proxy maintained predictive performance in our validation data, as shown in \ref{fig:oos}.  The mean cumulative state-of-health error at battery end-of-life between Blast-lite and this proxy is 0.03\% for the out-of-sample designs.

We implemented battery fade using static stressor proxies for computational tractability. Rather than modeling the full evolution of stress factors over the battery lifecycle, we approximated degradation using simplified metrics: mean state-of-energy from year zero, and cumulative throughput calculated as year-zero equivalent full cycles multiplied by operating years. While this simplification doesn't account for evolution in stressors, validation against dynamic stressor models showed minimal accuracy loss: a mean absolute error of only 0.14\% state-of-health at end of life. While we initially intended to develop a more complex stressor evolution model, exploratory testing of the $y_{0}$ stressors as proxy to provide a performance baseline proved sufficient for accuracy. This static approximation captures the dominant degradation drivers while keeping the optimization computationally manageable, making it well-suited for our large-scale system design exploration.

\subsection{Technoeconomic Analysis}

We employ technoeconomic assessment (TEA) to evaluate the financial performance of candidate designs. Using the net power flow models for each system detailed in the previous sections, we associate costs with the construction and operational parameters of each system, discounting future costs into $y_{0}$ dollars. We then divide these costs by the cumulative discounted energy production of the system to calculate the levelized cost of energy (LCOE) as shown in \ref{eq:lcoe}:

 \begin{equation}
      \mathrm{LCOE} = \frac{\sum_{t=0}^{n}\frac{Cap_t+Op_t}{(1+r)^{t+0.5}}}{\sum_{t=0}^{n}\frac{E_t}{(1+r)^{t+0.5}}
  }
      \label{eq:lcoe}
  \end{equation}

 where $t$ is the project year, $n$ is the evaluation year, $C_{\text{cap},t}$ and $C_{\text{op},t}$ are capital and operating expenditures in year $t$, $E_t$ is the energy delivered to the IT and cooling loads in year $t$ (zero before operations begin), and $r$ is the discount rate. This quantity provides us a direct point of comparison to evaluate the economic performance of different designs. Our base case discount rate is 7\%, and we discount mid-year. All values are in 2022 dollars. 
 
Note that LCOE has well-documented limitations when comparing technologies with different operational characteristics such as dispatchability \cite{joskow_comparing_2011, ueckerdt_system_2013}, because the metric does not completely capture how generation timing and reliability affect economic value. Our framework explicitly addresses this critique through scenario equivalence: all power systems face identical facility loads, reliability requirements, and geographic constraints, enabling direct comparison of the true cost to deliver equivalent energy services.

For PV+storage systems, capital and operating costs were derived by combining aggregate system projections from \cite{mirletz_annual_2024} with detailed cost decompositions from \cite{ramasamy_us_2023}. The NLR ATB provides bundled utility-scale PV+storage system costs of \$2,275/kW (2025, moderate case), which we disaggregated into component costs (PV modules, battery cells, and balance of system) using proportional shares from Ramasamy et al.'s detailed cost breakdowns. To allocate balance of system (BOS) costs between PV and storage components, we compared standalone PV systems to combined PV+storage systems in their data, attributing the full standalone PV infrastructure costs to the solar component and only incremental costs to storage.

For DC-coupled architectures, we estimated BOS cost by analyzing differences between coupling configurations in \cite{ramasamy_us_2021, ramasamy_us_2023}. While electrical system costs for DC coupled architectures showed marginal increases in \cite{ramasamy_us_2021} due to required DC-DC converters, those costs are exceeded by avoidance of the inverters. In \cite{gerber_cost_2023} the authors find that scale economies in power systems components result in DC-DC converters having lower capital costs compared to bidirectional AC-DC components. Based on this analysis, we derate DC-coupled solar BoS costs by 29.7\% with all savings attributed to reduced PV-side inverter infrastructure while battery-specific BOS costs remained constant. Battery module replacement at year 13 is treated as a capital expenditure using NLR ATB's projected costs for that future year. Land requirements are parameterized at 4.17 acres per MW of solar capacity \cite{ong_land-use_2013, bolinger_land_2022}.

Fixed operating and maintenance costs for PV systems were sourced from Ramasamy et al. \cite{ramasamy_us_2023}, using values for standalone PV systems to avoid double-counting battery replacement costs that are embedded in their combined system estimates. We note that these PV O\&M costs include inverter replacement, which DC-coupled architectures would avoid; however, as the reference does not specifically decompose costs, we retained the full values while acknowledging this represents a likely small overestimate for DC-coupled systems. Battery storage O\&M costs were obtained from the \cite{pacific_northwest_national_lab_esgc_2024} database entry for lithium iron phosphate systems. 

Natural gas plant costs vary by turbine class, plant configuration, and capacity. We source baseline capital and operating costs from established references: F-class and H-class single and combined cycle configurations from the NLR ATB \cite{mirletz_annual_2024}, and aeroderivative turbines from EIA's Annual Energy Outlook \cite{larose_capital_2024}. These reference plants provide cost benchmarks at specific capacities (e.g., 237 MW for F-class simple cycle, 702 MW for H-class combined cycle).

To estimate costs for configurations that differ from these reference plants, we apply empirical scaling relationships following the well-established ``six-tenths rule'' \cite{tribe_scale_1986}. After reviewing component-level scaling behaviors \cite{james_quality_2022}, we adopt scaling exponents of 0.67 for capital costs and 0.80 for fixed O\&M \footnote{We scale per-kW costs using exponents of -0.33 and -0.20, which is mathematically equivalent to scaling total costs with exponents 0.67 and 0.80 since per-kW cost $\times$ capacity $=$ total cost.}. The scaling relationships are:
\begin{equation}
\text{CapEx}_{\text{scaled}} = \text{CapEx}_{\text{ref}} \times \left(\frac{P_{\text{plant}}}{P_{\text{ref}}}\right)^{0.67} \times f_{\text{complexity}}
\end{equation}
\begin{equation}
\text{Fixed O\&M}_{\text{scaled}} = \text{Fixed O\&M}_{\text{ref}} \times \left(\frac{P_{\text{plant}}}{P_{\text{ref}}}\right)^{0.80} \times f_{\text{complexity}}
\end{equation}
where $P_{\text{plant}}$ and $P_{\text{ref}}$ are the candidate and reference plant capacities, and $f_{\text{complexity}}$ accounts for multi-unit configurations not represented in our reference dataset: 1.0 for 1-2 units, 1.3 for 3-4 units, and 1.8 for $>$4 units, reflecting increased balance-of-plant and coordination costs. Diesel system costs are linearly scaled according to system capacity from reference documentation \cite{ericson_comparison_2019}. We did not vary variable O\&M unit costs within a plant class. 

Annual fuel costs in year $t$ depend on the energy delivered by natural gas and diesel generators, multiplied by their respective fuel consumption rates. Natural gas generation is constrained by plant availability:
\begin{equation}
E_{NG,t} = \min(P_{plant,t} \cdot AF \cdot H_{year}, E_{required})
\end{equation}
where $P_{plant,t}$ is the nameplate capacity after degradation (MW), $AF$ is the availability factor incorporating forced and planned outages, $H_{year} = 8760$ hours represents annual operating hours, and $E_{required}$ is the facility's annual energy demand (MWh). Any shortfall is met by diesel generation:
\begin{equation}
E_{diesel,t} = \max(E_{required} - E_{NG,t}, E_{test})
\end{equation}
where $E_{test}$ accounts for mandatory testing requirements. The gas plant's operating efficiency depends on its operational load factor $\beta_{op} = P_{avg}/P_{plant}$, which represents the ratio of average facility demand to plant nameplate capacity. Operating efficiency in year $t$ is adjusted for both part-load penalties and calendar degradation:
\begin{equation}
\eta_{op,t} = \eta_{rated} \cdot f_{part-load}(\beta, c) \cdot \eta_{deg}(t)
\end{equation}
where $\eta_{rated}$ is the rated efficiency at full load, $f_{part-load}$ is the class-dependent part-load penalty function defined previously, and $\eta_{deg}(t)$ represents efficiency degradation over time.

Total fuel cost in year $t$ is calculated as:
\begin{equation}
C_{fuel,t} = E_{NG,t} \cdot \frac{\varrho}{\eta_{op,t}} \cdot p_{gas} + E_{diesel,t} \cdot \eta_{diesel} \cdot p_{diesel}
\end{equation}
where $\varrho = 3.412$ MMBtu/MWh is the thermal energy conversion factor, $p_{gas}$ is the natural gas price (\$/MMBtu), $\eta_{diesel}$ is diesel specific fuel consumption (gal/MWh), and $p_{diesel}$ is the diesel price (\$/gal). Diesel price were based on the USA national average according to the EIA circa July, 2025 \cite{noauthor_gasoline_2025}. We price natural gas at the state level, using an Olympic scored 5-year average of each state's industrial natural gas price \cite{noauthor_natural_2025}. We analogously assign each state an industrial electricity price based on the July industrial electricity prices \cite{noauthor_electric_2025} All prices are reported in $y_{0}$ dollars. 

Our LCOE calculation divides total system costs by delivered energy at the facility loads, not generated energy at the source. For solar+storage systems, $E_t$ represents the hourly facility demand successfully served after accounting for all conversion losses through the power distribution chain. Natural gas plant fuel consumption and equipment sizing reflect required generation at the source, yet we amortize costs only over the energy services delivered to the load, as with the PV based systems. This approach ensures consistent comparison across architectures: all systems are evaluated based on the energy that actually reaches IT and cooling equipment.

\subsubsection{Procurement, Construction, and Interconnection}

Beyond equipment and operating costs, project economics depend critically on implementation timelines that vary substantially across generation technologies and grid interconnection pathways. Economic value from any end-use is only unlocked when construction of generation and delivery assets completes. When we refer to construction, we mean the combined site permitting, engineering design, physical construction, testing, and commissioning required to begin plant operation. 

We applied the following capital outlay schedule across all architectures: For projects lasting one year or less the entire investment is assumed to occur at notice to proceed (year 0); for two- and three-year builds the schedule follows industry and modeling heuristics of 40 \%/60 \% and 30 \%/40 \%/30 \% in years 0–2 \cite{larose_capital_2024, schmitt_cost_2022}. Projects longer than three years are treated as being spent uniformly: the total cost is divided equally among the full years of design and construction and prorated for any remaining partial year.

For solar and storage systems, we began with the 2 year all-in timeline from \cite{baranko_fast_2024}. To validate this assumption, we consulted with relevant industry experts. An early stage grid storage expert estimated that the median time frame for component delivery of a large contract for LFP cathodes was 12 months using the existing mature supply chain. They estimated the time to bootstrap a new production base (for example in response to export restrictions) from scratch would be 18-36 months. A PV systems design expert estimated the average cumulative project timeline for an integrated utility scale PV+BESS project was 18 months, with 16-24 months as the overall range. Based on this feedback, we select 24 months as our cumulative project development timeline to bring solar+storage systems online. 

Natural gas plants face more varied timeline influences. We model construction time as differing between single and combined cycle plants, building off estimates in \cite{larose_capital_2024}, with the additional 2 months for combined cycle configurations reflecting increased plant size and complexity. Natural gas plant construction is not perfectly parallelizable, especially with regard to the keystone component of the turbine. We estimate that 30\% plant construction and validation must occur after turbine delivery, making turbine receipt the rate limiting factor. While reports of turbine delays can exceed five year \cite{gerke_nextera_2025}, we select value of three years for all turbine models. With base construction timelines of 40 months (simple cycle) and 42 months (combined cycle), total project duration is determined by the maximum of either the base construction time or turbine delivery plus post-delivery construction share.

As a point of comparison, we also estimate the time to interconnection for baseline grid-connected facilities: this includes the initial request processing, any associated studies, and construction of marginal transmission and distribution infrastructure. The process and complexity can differ between supply-side and demand-side interconnection requests, but we lack systematic data for demand side request duration. We adopt supply side interconnection queue data as our proxy for large load interconnection timeline, using raw data from \cite{rand_queued_2024}. We took the 2018-2023 median initial-request-to-commercial-operation span among completed projects reported for each ISO/RTO or non-ISO region, converted it to years, and assigned that value to every state in the region.  For the handful of states that straddle two regions we chose the higher of the two values. Although based on supply-side projects, these medians provide an empirically grounded proxy for the timeline a large load would expect to interconnect. We examine the sensitivity of our results to these timeline assumptions sensitivity analysis section of the SI.

\subsubsection{Design Optimization}

To identify cost-optimal configurations within each architecture class, we must search design spaces while maintaining the hard constraint that all systems achieve required facility uptime. The search strategies differ fundamentally between technologies with continuous versus discrete design variables.
Solar+storage optimization searches over continuous capacity ranges for both solar (typically 1-15× facility load) and battery power (typically 1-10× facility load). We employ two-stage optimization: sampling with relaxed evaluation (year-0 only) identifies feasible regions, then differential evolution refines the solution using full lifecycle analysis. Designs that cannot meet minimum uptime in $y_{0}$ will not meet minimum uptime at EoL. This approach efficiently handles the non-convex solution landscape created by the interplay between diurnal solar cycles, multi-day weather events, and battery state-of-charge dynamics

In our base case, we treat design increments in PV based systems as effectively continuous, the marginal solar cell or LFP module is minimal compared to the scale of the facility. We test sensitivity to this assumption by evaluating the affect on LCOE of jumping up to the nearest minimum PV and BESS design increment/step size. Sweeps across facility scales (10,000-100,000 GPUs), locations (Phoenix, Austin, Portland), and module sizes (1-10 MW) show cost penalties scale inversely with system size and are dominated by discrete factorization alignment rather than system performance. Median penalty across realistic configurations (2.5 MW PV, 1 MW BESS and below) is $<0.3\%$, though the combination of small facilities ($<25,000$ GPUs) and oversized step size ($>5$ MW) can cause the penalty to exceed 2\%. We observe the same phenomenon with commercially available power electronics. Taking a 4 MW utility inverter as our example, and based on component costs from our reference case, explicitly modeling inverter increment can increase LCOE on the order of .25\% over the continuous case at a 100k GPU facility, with penalties approaching 1\% at smaller facilities. Given the small absolute penalty of these industry representative step-sizes (\$0.001 / kWh on average) as compared to the continuous optimum, we accept continuous search as a simplifying assumption. 

The key computational challenge is avoiding full 25-year hourly simulation for each candidate design. We make two approximations. First, during search we minimize capital cost rather than LCOE for solar and storage systems —these are proportional for systems with identical operational profiles, making capital cost an exact proxy. Second, we evaluate only four anchor years that capture the system's degradation trajectory: fresh commissioning, maximum degradation before battery replacement (year 13), the discontinuity from replacement (year 14), and end-of-life (year 25). Between anchors, linear interpolation slightly underestimates the generated energy. This causes our method to overestimate LCOE by 0.014\% on average, acceptable given the 6-fold reduction in simulation calls.

Natural gas systems present a discrete combinatorial problem: select turbine model, determine unit count, and choose cycle configuration. The engineering filters described previously reduce the solution space sufficiently that exhaustive enumeration becomes tractable. We evaluate LCOE for each configuration passing the filters and select the minimum-cost option. Unlike the continuous solar+storage optimization requiring iterative search, natural gas selection is deterministic once the feasible set is enumerated.

\subsubsection{Value of Energized Compute}

Unlike conventional energy infrastructure that generates value only after commissioning, data centers house compute assets capable of generating revenue immediately if deployed elsewhere. Each month of construction represents not only capital carrying costs but also foregone computational output. This opportunity cost varies dramatically across power architectures: grid interconnection may require 4 years in congested regions, translating to 2 years of lost utility compared to faster-deploying alternatives. We term this differential the ``speed premium'', the economic value of earlier compute availability. This section develops a framework for quantifying these time-dependent costs and incorporating them into our technoeconomic comparison.

Proper comparison of total system economics requires accounting for both the levelized cost of energy and the opportunity cost of delayed operations. This opportunity cost represents the value of compute services that could have been delivered during the marginal construction period of a slower-to-build system. This necessarily raises the question: what would the value of those computations have been?

Within AI development, GPU-hours represent the fundamental unit of resource allocation, with utilization carefully tracked and optimized. The existence of liquid spot markets for H100 compute provides an empirical lower bound for this value: at \$2.40 per GPU-hour \cite{bloomberg_lp_sdh100_2025}, operators who cannot extract at least this value from their infrastructure would rationally rent capacity on the open market instead.

Infrastructure providers with internal AI development (such as major hyperscale cloud firms) must ration compute between internal workloads and external rental based on relative value. When internal workload value falls below rental rates, profit-maximizing firms should rent capacity rather than run lower-value jobs internally. The rental price thus represents the marginal decision threshold: the value of the last GPU-hour consumed internally.

We use the observed spot rental price for H100 capacity (\$2.40/GPU-hour) as our baseline opportunity cost. This represents a reasonable approximation for hyperscalers who continuously optimize across diverse workloads spanning frontier model training to routine inference and fine-tuning. This assumption likely provides a lower bound for specialized AI labs running exclusively research training workloads. See the sensitivity analysis and Figure 3 in the results for further discussion of how GPU-hour valuation affects results.

We model the value of lost compute during construction years as:
\begin{equation}
VLC_{t} = \sum_{h=1}^{H_{\text{idle},t}} N_{\text{GPU}} \times p_{\text{spot}}
\end{equation}
where $VLC_t$ represents the value of lost compute in year $t$, $H_{\text{idle},t}$ is the total idle compute hours in year $t$ prior to operations, $N_{\text{GPU}}$ is the total GPU count, and $p_{\text{spot}}$ is the hourly spot-market price per GPU-hour \cite{bloomberg_lp_sdh100_2025}. Only idle hours before the start of operations at $t_{\text{ops}}$ are considered. We incorporate $VLC_t$ into our total cost assessment as:
\begin{equation}
\text{LCOE}_{\text{adj}} = \frac{\sum_{t=0}^{n}\frac{(C_{\text{cap},t} + C_{\text{op},t} +
  \text{VLC}_t)}{(1+r)^{t+0.5}}}{\sum_{t=0}^{n}\frac{E_t}{(1+r)^{t+0.5}}}
\end{equation}
This formulation treats lost compute value as functionally an additional expense during construction years, ensuring architectures with longer deployment timelines bear the full economic burden of delayed operations. By default the accounting places foregone-compute cost from project initiation onward; alternative schedules can be modeled by adjusting when $VLC_t$ becomes non-zero.

Note that indexing $VLC_t$ to $y_0$ is a modeling convenience, not an economic claim: it does not assume a firm procures its GPUs at project start and leaves them idle through construction, a strategy no operator would adopt. The origin is simply the date from which foregone compute is counted, and the comparison is insensitive to it across the relevant range. The quantity that governs architectural preference is the difference in present-value cost between two options, and any opportunity cost accrued before the earliest energization (``go-live'') date is common to all architectures and cancels exactly in that difference, provided the origin falls on or before that earliest date. We could equally index to $y_2$, $y_{1.5}$, or any date up to the fastest go-live and recover the same ranking. 

Because $LCOE_{\text{adj}}$ divides by architecture-specific discounted delivered energy, this common term does not cancel exactly in the ratio, but the residual sensitivity of the ranking to the origin is small relative to the energization-date differential that drives it. Index choice can alter preference order only outside this regime---when chip delivery postdates the fastest go-live, so that the earliest-energizing architecture cannot fully capitalize on its head start. Accordingly, firms with known internal procurement and silicon-delivery timelines should index $VLC_t$ to their own schedule.

Figures~\ref{fig:queue_breakeven} and~\ref{fig:ng_breakeven} apply this framework geographically by inverting the comparison: they map how quickly a grid interconnection (Figure~\ref{fig:queue_breakeven}) or a natural-gas plant (Figure~\ref{fig:ng_breakeven}) would have to be delivered to match the speed-premium LCOE of DC-coupled solar at each location.

\subsection{Sensitivity Analysis}

We evaluate how parameter uncertainty affects both system costs and architectural selection through two complementary approaches: varying parameters for fixed designs to test cost estimate robustness, then re-optimizing with modified parameters to assess robustness. Within a fixed design, LCOE is linear in each cost parameter, so a one-at-a-time perturbation moves LCOE in proportion to that component's share of lifecycle cost. Re-optimizing under perturbed parameters can only lower costs with respect to the fixed design response, so the fixed design case upper bounds the re-optimized case. 

For our fixed design and cost structure sensitivity sweep, we selected a site in southeastern Colorado (37.5, -103.5) with a local gas price of \$7.70 / MMBtu, and tested at 99\% facility uptime. We chose this site for its moderately high insolation and moderate natural gas price.  Where a perturbation changes the resource or the energy required to deliver a given service (design weather year, network topology, optimization stability), we evaluate sensitivity on multi-site panels spanning the 14 continental U.S. DOE climate zones.

\subsubsection{Baseline LCOE sensitivity} Excluding idling, solar+storage base LCOE responds most strongly to the discount rate (27--28\% decrease at 3\%, 41--43\% increase at 12\%). Balance-of-system and year-0 battery cost are next, each shifting LCOE by 4--8\% under $\pm$20\% variation, followed by module cost ($\sim$3--4\%); construction timeline, O\&M, and replacement-battery cost are second-order ($\leq$5\% on the long-build side, and $\leq$1.6\% for O\&M and battery BOS). Land costs are negligible, requiring a 6$\times$ increase to $\sim$\$1M/km$^2$ to match even battery O\&M in cost influence.

Natural gas base LCOE is most sensitive to fuel price, with the continental U.S. range (Oklahoma to California) creating roughly \$0.08/kWh variation at the 100,000-GPU scale, larger for smaller turbines. Discount-rate effects are comparable (\$0.03--0.05/kWh across 3--12\%). Construction timeline and $\pm$20\% capital costs each affect LCOE by $\sim$\$0.01--0.02/kWh.

\subsubsection{Speed-premium sensitivity} Incorporating deployment speed dramatically amplifies temporal parameter influence. For solar+storage, discount rate and GPU opportunity cost (\$1.00--4.00/hr) each create \$0.55--0.60/kWh swings in $\text{LCOE}_{\text{adj}}$---larger than construction-timeline effects ($\sim$\$0.38/kWh). Natural gas shows even stronger sensitivity (GPU idling $\sim$\$1.36/kWh, discount rate $\sim$\$1.22/kWh, construction timeline $\sim$\$0.89/kWh) due to its longer baseline timelines. At current GPU rental index values, deployment speed dominates nameplate energy economics.

As a complementary check on the robustness of our estimates of which of our modeled architectures has the lowest $\text{LCOE}_{\text{adj}}$, we compute a pairwise indifference envelope across all 4,317 valid hexes at 100,000 GPUs: a hex is flagged as contested if its second-cheapest architecture (among AC-Coupled MV solar, DC-coupled MV solar, grid-connected, and NG) falls within 1~\textcent/kWh of the lowest cost option. We selected this value as an arbitrary reporting threshold rather than any data-derived breakpoint. Under the baseline 4-year gas build, 149 hexes (3.5\%) are contested, all at the solar--grid boundary; NG never enters the envelope. In the accelerated 3-year  NG case, the count of near indifferent tiles climbs to 348 hexes (8.1\%), with the new mass split roughly evenly between DC-solar-preferring and NG-preferring hexes (168 vs.\ 136 respectively; the remaining 64 are grid). In both cases contested hexes are a small minority, and the contested hexes do not skew systematically toward any particular technology as runner up. 

\subsubsection{Fuel-price volatility}

Natural-gas LCOE inherits delivered fuel-price volatility, but the magnitude of that exposure is fixed at design time by the plant a facility can economically build, which is in turn bounded by its scale. For a representative city in each continental DoE climate zone and at three facility scales, we size the cost-optimal gas plant at the state's baseline gas price, then recompute its LCOE under a set of alternative delivered fuel-price levels (Figure~\ref{fig:ng_elasticity}); the slope of LCOE against fuel price is the facility's committed fuel-price elasticity. No design response to price is permitted: each plant is a price-taker on its installed heat rate.

Elasticity is governed by the installed heat rate, and the accessible heat rate is gated by scale. Small and mid-scale facilities (15{,}000 and 75{,}000 chips, $\sim$20 and $\sim$100~MW) draw aeroderivative simple-cycle plants almost everywhere, with an elasticity near 0.0116~\$/kWh per \$/MMBtu that is nearly independent of location. Combined cycle, with its lower heat rate, becomes the cost-optimal choice only at sufficient scale (135{,}000 chips, $\sim$180~MW) and only where mean gas prices are high enough to amortize its larger capital cost. When selected it flattens the elasticity. Because combined cycle appears only at the largest scale in our sweep, scale and architecture are collinear.

Elasticity sets how LCOE moves per unit of fuel-price change; the realized spread also depends on how much prices actually move. To bound the latter we run a complementary analysis on the 48 contiguous states at three scales (10k/75k/300k chips), drawing plant-lifetime fuel-price paths by stationary block bootstrap (Politis--Romano, mean block 12 months; 2000 paths) from each state's post-2010 monthly industrial series, sizing the plant once at the bootstrap-mean price and running the fixed plant over every path. Historical fuel-price dispersion moves NG LCOE within a P5--P95 band of roughly \$0.005--\$0.024/kWh (median cell $\sim$\$0.011/kWh), with no shift in the central estimate---LCOE is linear in fuel price, and the bootstrap mean reproduces the deterministic mean-price LCOE to within 0.2\% in every cell. The band narrows with scale (median \$0.0139 $\to$ \$0.0110 $\to$ \$0.0084/kWh from 10k to 300k), consistent with the elasticity result: larger facilities draw lower-heat-rate plants, so each delivered kWh carries less fuel cost and less fuel-cost dispersion.

The practical consequence is that the design-time option to blunt fuel-price exposure by building a more efficient plant is available only to sufficiently large facilities; smaller deployments are locked into the more fuel-exposed simple-cycle regime. Two caveats. First, this is purely a design-time option: once built, every facility is a price-taker on its installed heat rate, the regime the figure depicts. Second, this cost-side sensitivity is generally second-order relative to the speed premium that dominates the cross-architecture comparison.

\subsubsection{Technical parameter stability sensitivity} 
Re-optimization with modified technical parameters shows robust architectural selection. For natural gas, 72 parameter variations across four scales triggered only five configuration changes, all from power electronics efficiency improvements reducing generation needs across turbine size thresholds. Technical parameters (part-load curves, temperature derating, conversion efficiencies) show 1--3\% LCOE impacts when relevant to selected configurations. Solar systems exhibit comparable stability, with battery round-trip efficiency largest (about $-2.6\%$ to $+3.2\%$ over the 0.85--0.95 range) and power conversion $\pm$1--3\%. Sizing to meet end-of-life availability requirements effectively eliminates design contingency sensitivity: under common random numbers the re-optimized design is unchanged across the 1.03--1.10 range, leaving no first-order cost impact. In five instances, the design optimization for natural gas systems switched to a different configuration when changing a power electronics value. In each case, high-end estimates for UPS or PSU efficiency enabled a lower-capacity design to pass the minimum load engineering filter.

Variation in the modeled battery replacement year (16 and 10 years vs.\ the 13-year baseline) shifts LCOE by $-0.61\%/+5.10\%$ for AC-coupled systems and $-1.14\%/+3.48\%$ for DC-coupled systems. These cases are re-optimized with the degradation schedule re-anchored to the replacement year, so both the discounted timing of the replacement capital outlay and a modest re-sizing of the battery contribute: the optimal battery capacity shifts by a few percent across the range, with a later replacement requiring somewhat more initial storage to hold reliability at the more-degraded pre-replacement year. 

The battery replacement vs augmentation strategy that likely best balances cost and operational objectives would be fixed period augmentation, replacing only faded capacity once or twice over the facility lifespan but continuing to operate existing batteries, with tests of such augmentation strategies reducing costs by 8\% compared to the fixed replacement strategy. Such savings are notable, and make theoretical sense given the high SoH of these BESS systems at modeled replacement year. We should tread lightly in our interpretation however, as our modeled system lifespan exceeds the operational history of utility scale LFP systems in wide commercial deployment. To remain within in the duty cycle distribution our fade model is built on, and in light of known industry reliability preference and operational caution, we hold replacement year (and system replacement rather than augmentation) fixed while acknowledging this conservative assumption raises baseline costs of our BESS based systems.

\subsubsection{Network topology and storage placement}

Beyond the component-efficiency sensitivities above, we assess how the power systems choices within the PV+BESS architecture (distribution voltage, where storage sits, and load conditioning). Unless noted, topology deltas are reported on base LCOE for a 100k-GPU MV facility; the medium- vs.\ low-voltage comparison spans all 4{,}317 hexes, while the placement, path-value, and conditioning analyses use a 14-site panel spanning the DOE climate / solar-resource range (best-of-five re-optimization per site). Table~\ref{tab:topology_sensitivity} orders these by magnitude: a single first-order lever (whether the full IT load is conditioned online) and a set of second-order siting choices whose AC cost reduces to whether storage rides the shared PV inverter. The text reports the panel ranges that establish robustness. In the MV DC case, siting nodes include within the data hall or off the main MV distribution feeder bus, or direct coupled to the PV. In the AC case, storage could be placed DC coupled with the PV behind the inverter, or within the data hall (or off the MV spine) with their own dedicated inverter. In each case, there is a dominated option: MV coupling strictly outperforms direct coupling to the PV in the DC architecture. In the AC architecture, data hall embedded is strictly preferred to MV spur. In our discussion of topological choices we will discount those nodes moving forward, see \ref{fig:lv_powerflow}. 

Across the full geographic sample, the MV architecture costs +3.7\% (AC) and +2.4\% (DC) more in median LCOE than the single-bus modular LV architecture. This delta is the added conversion loss of the MV design. Composition differs by architecture: for the all-DC system it is purely the two MVDC spine crossings, whereas for the AC system it also includes the discrete MPPT and battery DC-DC stages that accompany field DC-coupling of storage as well as marginal inverter requirements if hall embedded. The LV architecture is therefore modestly cheaper where physically realizable, but, as established in the Power Systems Architecture section, it is bounded by practical limits of fault current, bus and component ampacity limits, and PV gathering line losses. 

For each of the MV designs, we quantify the cost of siting all storage at each candidate node by holding the optimized baseline design fixed, varying the distribution of storage placed at different nodes at the network (e.g. 70\% in the data hall, 30\% behind the solar inverter, etc.) adjusting the total PV and BESS capacity to return exactly to the threshold uptime constraint, and measuring the change in total system costs. For the all-DC system the discharge-weighted pull is toward the load: across the 14-site panel, relative to the spine baseline, all-field storage costs \textbf{+0.7 to +1.4\%} and all-hall \textbf{$-$0.8 to $-$1.5\%} (medians: +1.0\% and $-$1.2\%), a tight band across the full climate range. The hall position is cheapest but ampacity-strained---charging the islanded battery fleet can draw 7--16$\times$ facility load while the hall LV buses admit only a few percent of it---so the realized benefit is negligible and, once the ampacity cap is applied, the spine fed centralized storage baseline is within $\sim$0.05\% of the cost optimal embedded LV BESS case. Because the all-DC system has no AC waveform to form, nothing forces storage to the hall, and DC placement is effectively closed.

For the AC system, relocating storage off the field forfeits the shared PV inverter; the resulting re-inversion cost is additive in the relocated share. We price it conservatively at the full avoided PV-side inverter bundle, \$253/kW (the AC$-$DC solar-BoS differential of the TEA cost model).  Provisioning the distributed hall increment---of order one facility-nameplate of storage, $\approx$10\% of the islanded fleet---then adds a re-inversion capital cost of +0.5 to +1.3\% of LCOE across the sweep (median +0.83\%). 

A complementary fixed-design analysis (improving one power-flow path by one percentage point and re-shrinking the fleet to achieve equivalent energy services) finds the battery discharge path is the dominant cost lever in every architecture and climate. Each percentage point improvement in cumulative path efficiency of battery-discharge-to-load is worth 0.7--0.9\% of system costs across the 14-site panel (median 0.78\%). This is roughly 2--4$\times$ the charge path and well above the solar-to-load path. This ordering is robust to supply strategy: walking the solar/battery ratio across a four-fold range along the 99\%-uptime frontier leaves discharge sensitivity essentially flat. PV-to-battery-charge sensitivity rises with battery-heaviness, from near-zero in solar-heavy, heavily-curtailed designs to $\sim$0.5\% in battery-heavy ones. The practical implication is that the cost exposure of the power system concentrates on the storage discharge chain (the battery converter, the inverter, and the switchgear) regardless of the optimal design being solar- or battery-driven.

The our baseline LV and MV AC architectures maintain power quality through grid forming and do not fully condition all IT load to manage training transients. We separately quantify the cost of full double-conversion conditioning: routing the entire IT (and optionally cooling) load through an in hall BESS uninterruptible power supply. We apply the double-conversion factor (0.94) to the AC bus-to-load path and re-optimizing. This is the most cost consequential choice in distribution architecture for costs: across the 14-site panel, full IT conditioning adds +5.6\% to +6.2\% to base LCOE (median +5.9\%). Conditioning IT and cooling together +6.1\% to +6.7\% (median +6.4\%). The penalty is climate-invariant (it varies by under a percentage point across resource regimes spanning all DOE climate zones) with at most a mild ($\sim$0.5~pp) tendency to be lower in cold, low-solar climates, near the run-to-run optimization noise. Whether such conditioning is a genuine requirement (and for IT alone or IT and cooling) is an operational decision \cite{chouksePowerStabilizationAI2025}, not a model output; we report its cost conditionally so that a reader requiring fully-conditioned power can locate the corresponding architecture and premium. 

Finally, we decompose the all-DC advantage itself into a path-efficiency component and a capital component. They contribute in roughly equal measure. The efficiency side reflects that the DC chain delivers more of the generated energy to the load: roughly 90\% of PV output reaches the IT load versus 83\% for the AC chain. The cumulative battery-charge-discharge-IT load sees comparable savings for the DC architecture over AC. Holding capital costs equal, these efficiency improvements are worth $\sim$7.9\% of LCOE. The capital side is the avoided PV-side inverter and AC collection equipment ($\sim$\$253/kW of solar balance-of-system), a further 7.5--9.3\%, larger where the optimal design runs more PV-heavy since the saving accrues per kW of PV. Composed at fixed design, these components predict a 15--16\% advantage, modestly below the 18.3\% mean across the re-optimized continental sample at the 100{,}000-GPU scale. The difference reflects the sizing response, as higher path efficiency also reduces the overbuild required to meet the reliability constraint. This decomposition makes the advantage robust to uncertainty in DC conversion-equipment pricing. Holding optimized designs fixed, the DC-side solar BoS must reach \$965--1{,}176/kW (1.6--2.0$\times$ our central estimate, and 13--38\% above the corresponding AC-system BoS) in our range of evaluated designs before the all-DC advantage vanishes entirely. This would require the DC-specific power-conversion equipment to cost more than the entire avoided inverter bundle.

\subsubsection{Optimization stability}

To characterize the robustness of our designs and the stability of the optimizer, we run 24 independent optimizations at each of 14 locations spanning the DOE climate zones, for the DC-coupled architecture serving 100{,}000 GPUs at 99\% uptime on the medium-voltage spine (design load $\approx$130~MW, varying modestly with cooling performance). Across locations the LCOE coefficient of variation ranges from 0.1\% to 4.7\% (median 1.0\%, mean 1.4\%) and varies with climate: high-solar-resource sites (Miami, Phoenix, Las Vegas, El~Paso) converge tightly (CV $<0.35\%$), while humid or variable-resource climates (Northern Virginia, Houston, Atlanta) are the loosest (2.7--4.7\%). Figure~\ref{fig:optimization_stability} shows four representative climates, with Albuquerque and Denver at the tight end and Baltimore and Chicago more dispersed.

Component sizing varies substantially more than LCOE. Across the 14 sites the solar-capacity CV has median 5.3\% (mean 5.5\%, up to 10\%) and the battery-capacity CV median 6.9\% (mean 6.3\%, up to 11\%), several-fold the LCOE CV. This disparity indicates a near-flat objective basin---configurations differing by 5--10\% in component sizing achieve nearly identical cost (within $\pm$2\% LCOE). This Pareto frontier-like behavior reflects substitutability between solar and battery capacity on the margin near the boundary of the uptime constraint.

Cross-location differences reflect search difficulty rather than a change in problem structure. Where solar resource is abundant and steady, the cost-optimal region is a large target that sampling-based optimization reliably finds; where resource is variable, that region is narrower or split into local basins, lowering the chance a random seed initializes near the global optimum. The figure shows this directly: easy locations cluster tightly along a clear trade-off curve, while harder locations mix dense clusters (converged) with interior points (suboptimal convergence) carrying 2--3 cents/kWh penalties

For challenging locations (CV $> 2\%$), running 3--5 independent optimizations and selecting the minimum provides more reliable results. Alternatively, increasing the differential evolution iteration budget from 25 to 50--100 generations improves convergence at modest computational cost for detailed investigation of specific candidate sites.

\subsubsection{Design-year weather sensitivity} 

Our main analysis sizes each design against a single typical meteorological year (TMY), which raises two questions: is a TMY a biased basis for design, and how much does a design's cost depend on the particular weather year chosen? We re-optimized the DC-coupled architecture against the TMY and against each of the fourteen historical years 2010--2023 (drawn from the NSRDB, our weather source throughout). We chose two cities in each of the fourteen DOE climate zone modelling 40,000 GPU facilities and 99\% uptime, taking the best-of-five seeds per design. We draw inference per location, Figure~\ref{fig:tmy_hist} pools the resulting single-year deviations across all location-years to show their shape.

Relative to the distribution of historical years, TMY is a mildly optimistic design basis, with the optimism concentrated in the tail rather than spread across the bulk. The pooled single-year deviations (Figure~\ref{fig:tmy_hist}) are right-skewed: most years sit close to the TMY value, but a thin upper tail of punishing years pulls the mean to $+\$0.0121$/kWh above TMY. Within a site, the across-year LCOE distribution is right-skewed at 23 of 28 locations (mean per-location skew $+0.68$, $p < 10^{-4}$), and across locations the historical-year mean exceeds TMY by a median of 4.5\% (mean 3.0\%); a location-clustered robust test places the mean TMY shortfall at \$0.012/kWh ($p = 0.024$), with TMY the cheaper basis at 20 of the 27 sites. Because the gap is a tail effect and not a uniform offset, TMY closely tracks the typical (median) historical year, and trimming design years beyond $\pm2$ standard deviations from the mean removes the significance of the bias: TMY misses the worst years rather than systematically diverging from the typical one. 

This tail-driven optimism nonetheless leaves TMY the preferable single-year basis for designing systems. A uniformly drawn historical year is unbiased for the across-year mean by construction, but its draw-to-draw scatter is large: the single-year LCOE has a per-location standard deviation around the mean of $11.5\%$ (median across sites). TMY, a deterministic construction near the typical year, carries essentially no sampling variance, so its error is the $\sim5\%$ bias alone. As point estimators of the across-year mean cost, a single random year is therefore roughly $2.2\times$ noisier (RMSE 11.5\% vs.\ 5.2\%) and outperforms TMY at only 2 of the 27 sites.

The mechanism is a solar-resource-timing convexity. The reliability constraint makes cost a convex function of resource adversity: the worst sustained cloudy stretch sizes storage nonlinearly, and a TMY, assembled from independently selected typical months, tends to smooth over exactly the punishing multi-day stretches that drive the worst design years. This single convexity explains both the per-year right skew and TMY's aggregate optimism.

A key consequence of this is that a system sized to a typical year may under-provision against a bad year and miss its uptime target. Site-specific, reliability-critical point designs should size against historical worst-stretch resource rather than a single representative year. Designing on a bootstrapped sample of historical years in sequence, or iteratively running the optimizer with the historical case year represent potential paths to address this.

\section{Additional Figures and Tables}
\clearpage

\begin{figure*}[p]
\centering
\includegraphics[width=0.95\textwidth]{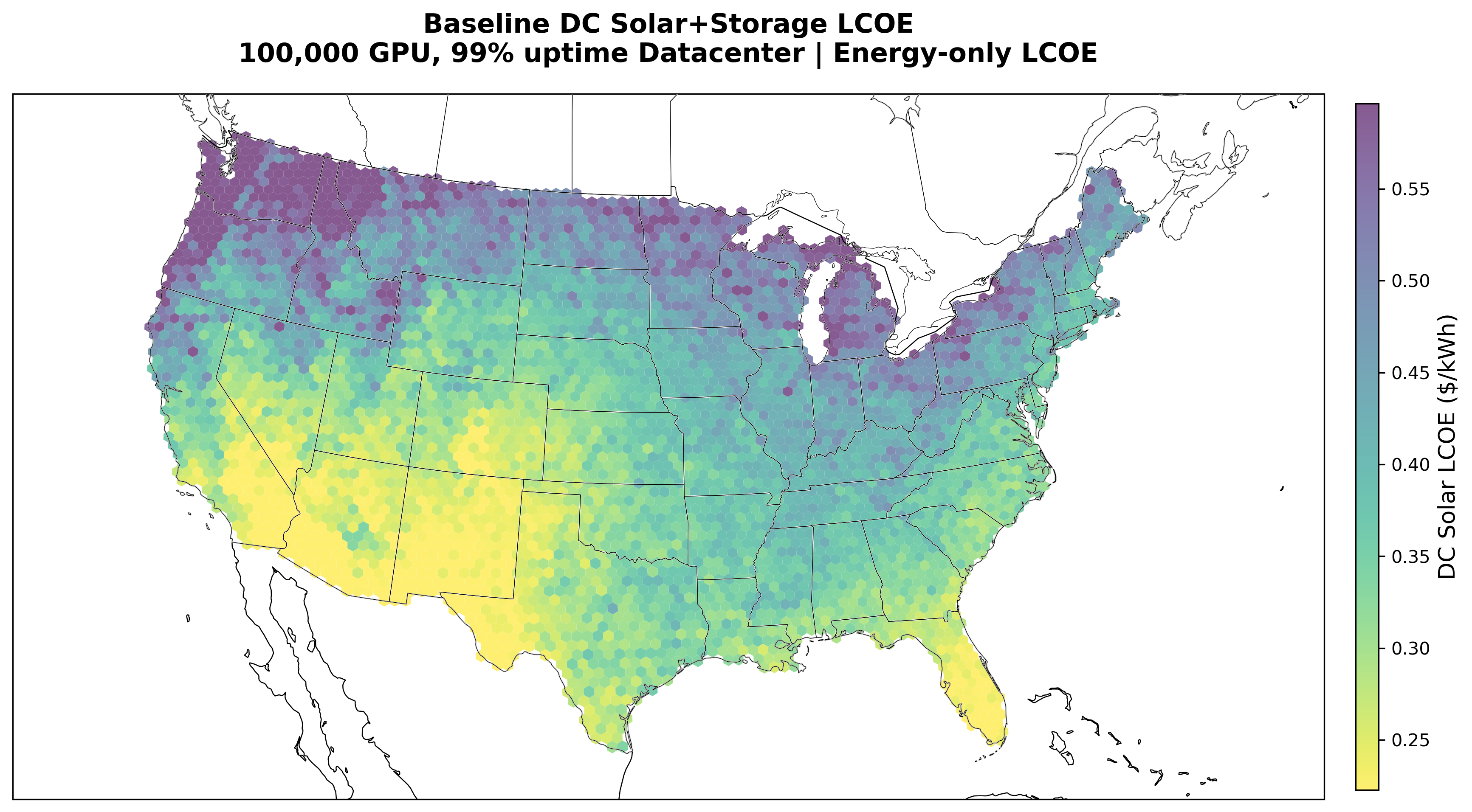}
\caption{\textbf{LCOE for all-DC microgrids across the continental United States.} Baseline LCOE ranges from \$0.15/kWh in the most consistently sunny regions to \$1.10/kWh in the Olympic peninsula of Washington. System cost is more sensitive to worst-case cloud cover than latitude driven expected irradiance.}
\label{fig:dc_heatmap}
\end{figure*}

\begin{figure*}[p]
\centering
\includegraphics[width=0.95\textwidth]{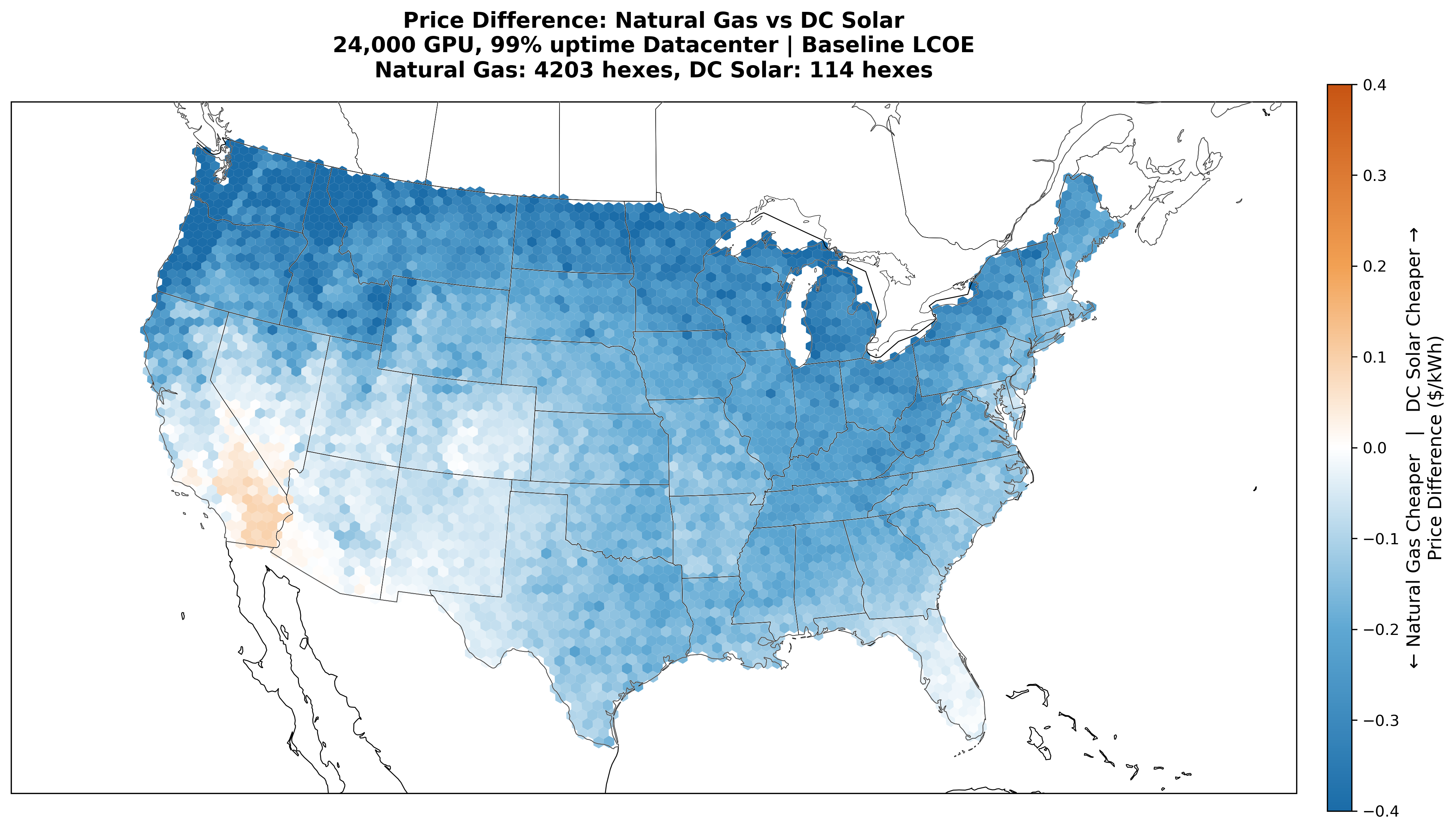}
\caption{\textbf{Comparative LCOE of natural gas and DC Solar microgrid supplying a 24,000 GPU data center across the United States.} DC solar achieves or nears cost-parity in high-insolation regions spanning southern California, Arizona, Nevada, New Mexico, and Colorado. In solar poor regions of the pacific northwest and upper midwest, natural gas significantly outperforms DC solar systems. Note the skew in performance geographically: the worst performing locations for solar are disproportionately expensive, while the spread of gas costs between the best and worst tiles is more modest.  }
\label{fig:ng_dc_comp}
\end{figure*}

\clearpage

\begin{figure*}[p]
\centering
\includegraphics[width=0.95\textwidth]{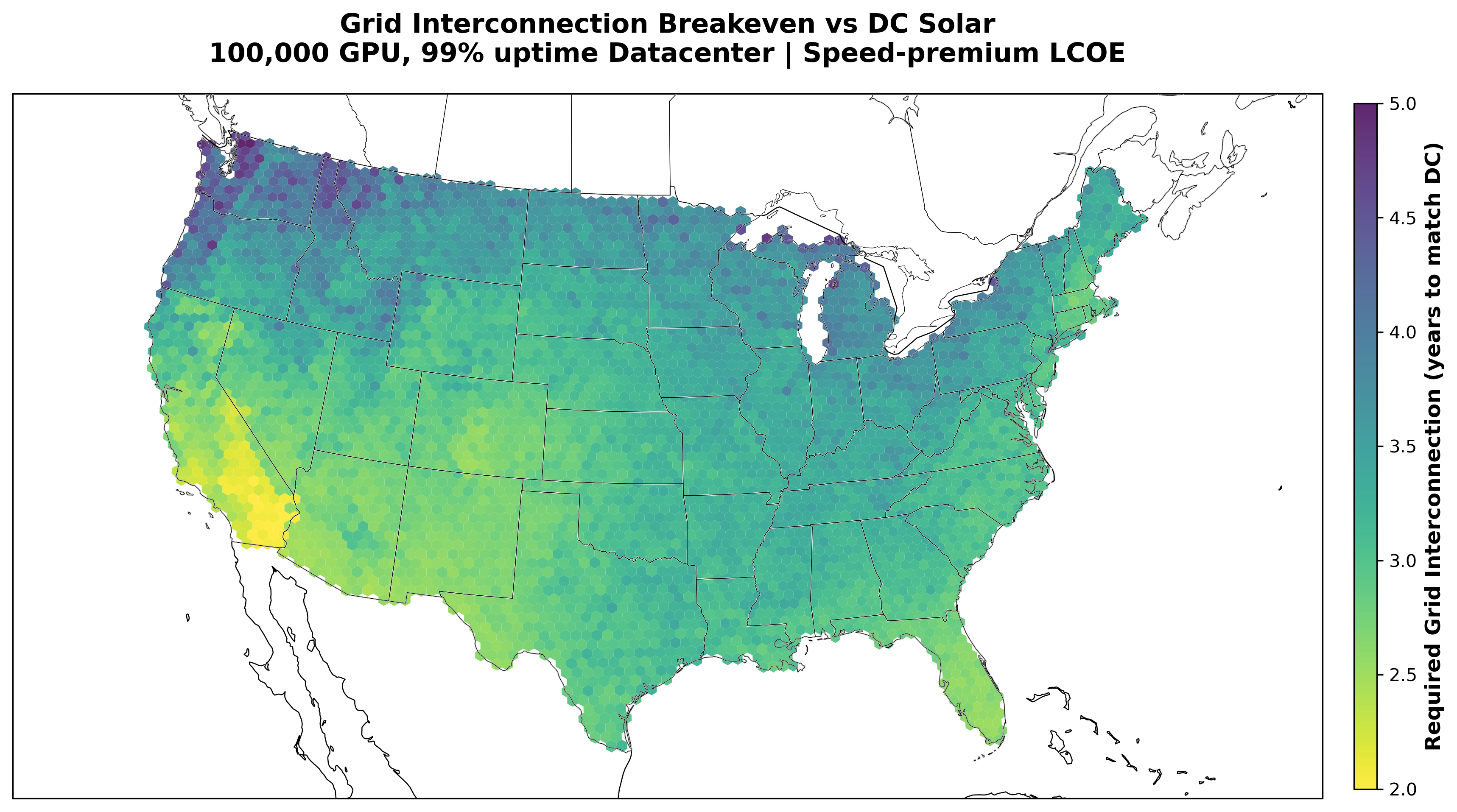}
\caption{\textbf{Required grid interconnection timeline for grid and DC-coupled solar microgrids to achieve equal speed-premium LCOE.} For a 100,000 GPU datacenter, holding GPU opportunity cost constant at \$2.40/hr. Note this plot displays only the absolute interconnection time required for indifference and is not relative to regional interconnection estimates.}
\label{fig:queue_breakeven}
\end{figure*}

\clearpage

\begin{figure*}[p]
\centering
\includegraphics[width=0.95\textwidth]{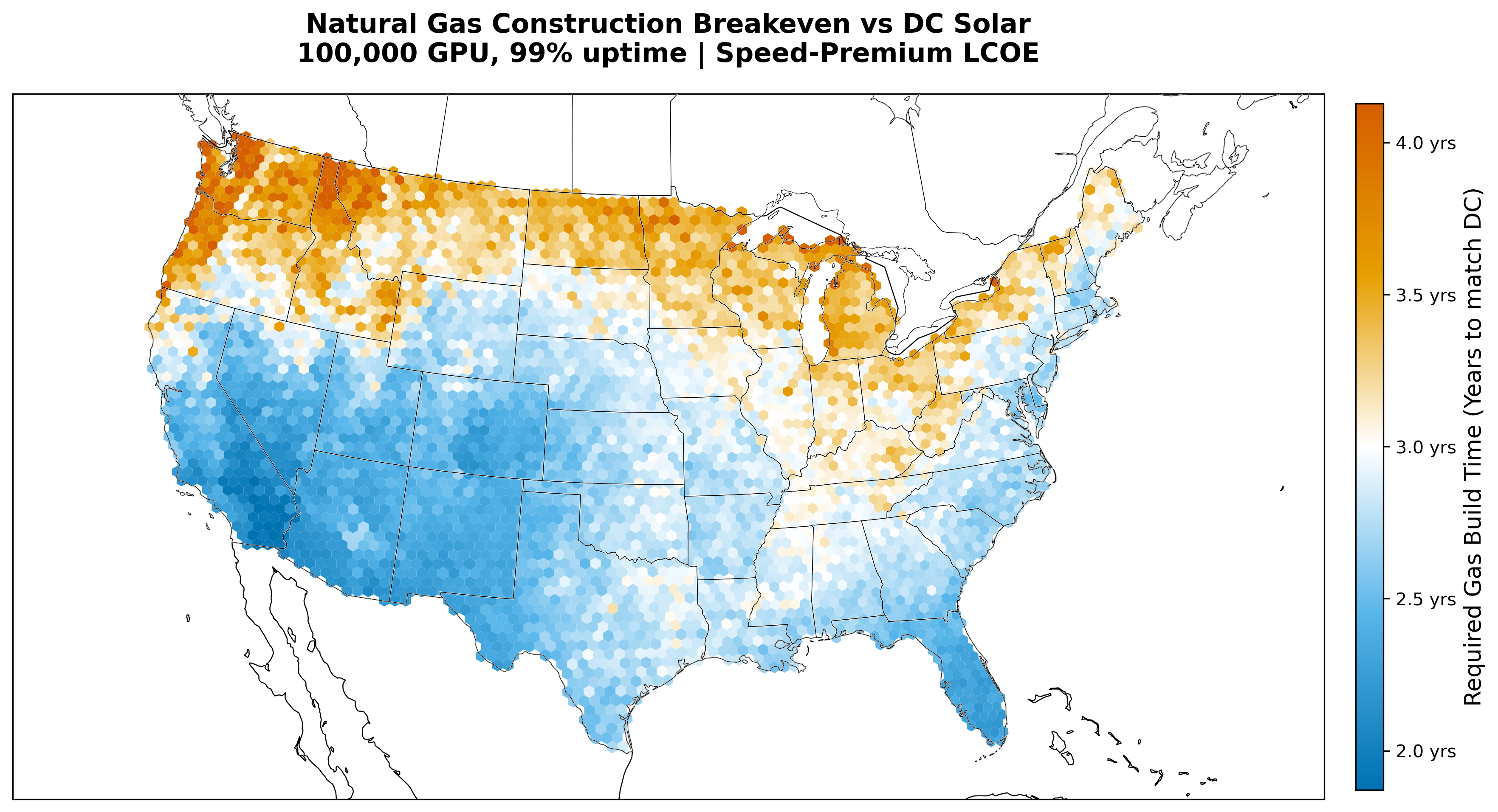}
\caption{\textbf{Required natural-gas construction time to match DC-coupled solar on speed-premium LCOE.} For each hexagon we solve for the gas-plant construction duration at which the gas system's speed-premium-adjusted LCOE equals that of the optimized DC-coupled solar design (100{,}000 GPUs, 99\% uptime; GPU opportunity cost \$2.40/hr; DC solar built in 2.0 years). White marks the 3.0-year accelerated-gas target. Blue regions---2{,}801 of 4{,}317 hexes---require gas to be built \emph{faster} than this accelerated pace merely to match solar, and concentrate in the high-insolation Southwest where solar is cheapest; orange regions can be built at the standard $\sim$4-year pace and still win, reflecting solar's higher cost in the cloudy North. For a facility of this scale NG can match DC solar at some build speed in every hexagon; the map shows how fast it must be built to do so.}
\label{fig:ng_breakeven}
\end{figure*}

\clearpage

\begin{figure*}[p]
\centering
\includegraphics[width=\textwidth]{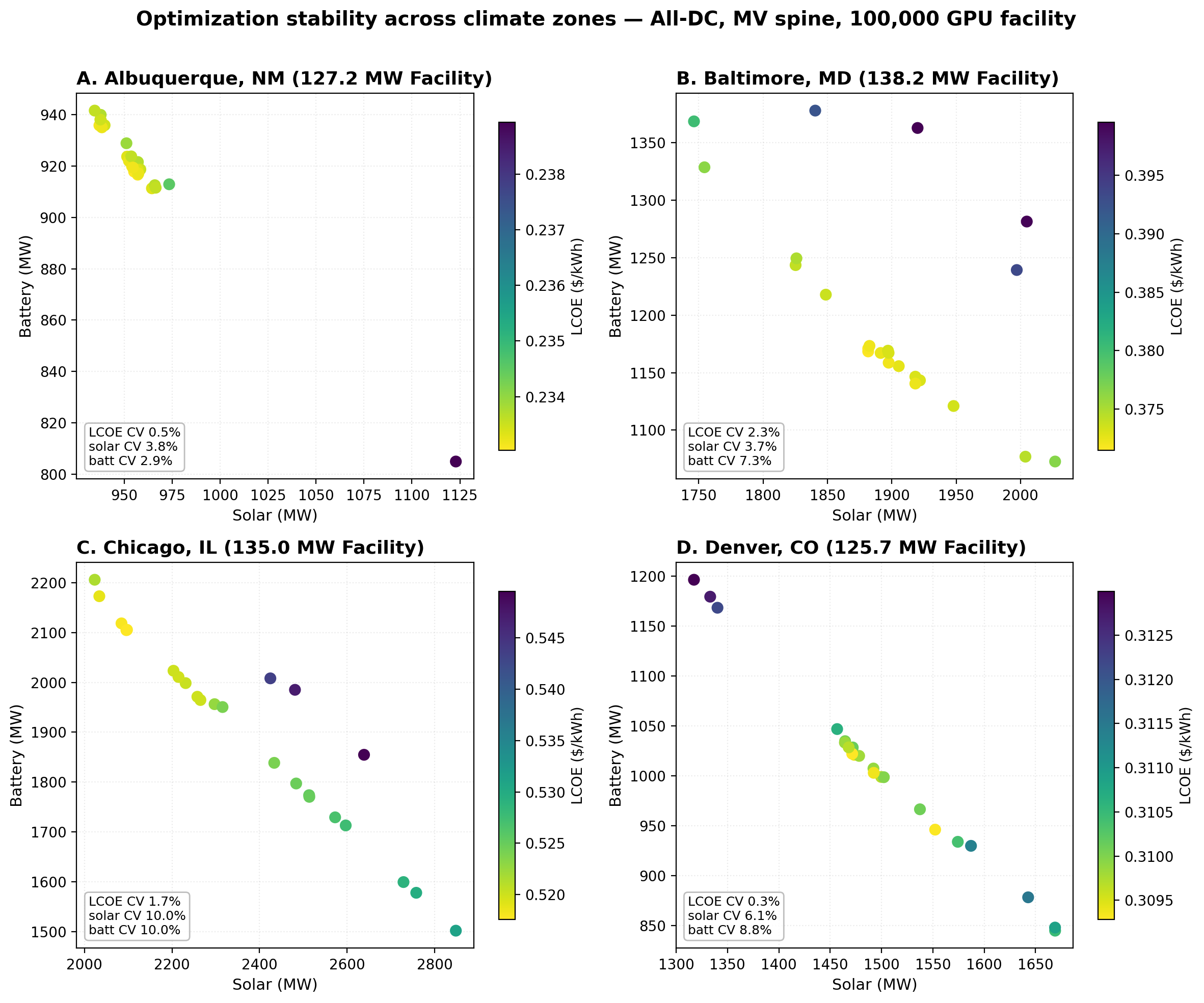}

\caption{\textbf{Optimization stability across representative climate zones.} DC-coupled, 100{,}000 GPUs at 99\% uptime, medium-voltage spine, 24 independent optimizations (one per seed) per panel. Each point is one optimization, colored by LCOE (\$/kWh). Interior solutions indicate outlier runs with suboptimal convergence ($> 2\%$ LCOE penalty). Component sizing variation (solar CV 3.7--10\%, battery CV 2.9--10\% across these four sites) substantially exceeds LCOE variation (0.3--2.3\%), indicating near-flat objective function regions where many configurations achieve nearly identical economic performance. Note the regional variation in optimal design cluster between comparatively battery favoring and compartively PV favoring. Interior solutions represent convergence failure by the optimizer to identify the performant design basin.}
\label{fig:optimization_stability}
\end{figure*}
\clearpage

\begin{figure}[t]
\centering
\includegraphics[width=\linewidth]{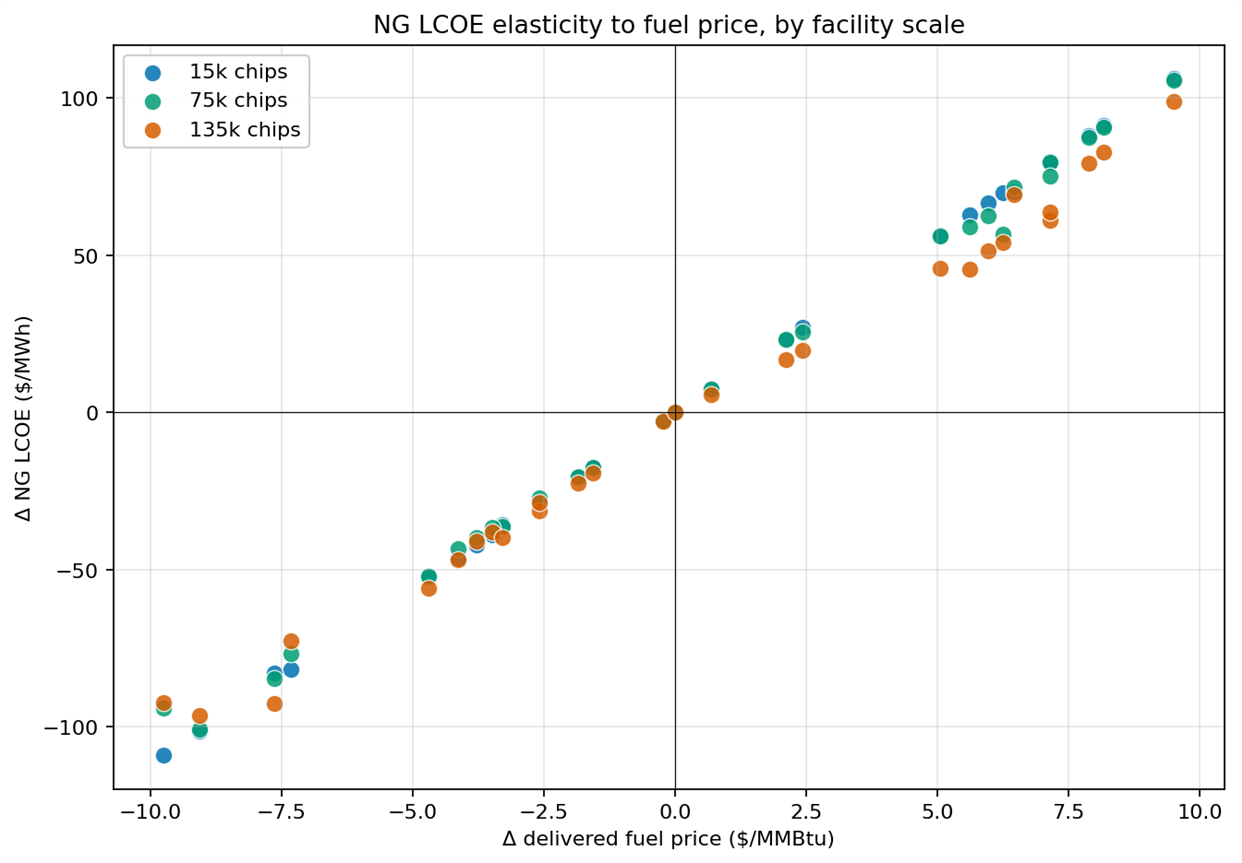}
\caption{\textbf{Committed fuel-price elasticity of natural-gas LCOE, by facility scale.} Each point is one (city, scale) plant, sized once at its state's baseline gas price; the plant is then held fixed and its LCOE recomputed at a set of alternative delivered fuel-price levels. Axes show deviations from each plant's baseline point ($\Delta$ delivered fuel price, $\Delta$ LCOE). The common near-linear trend is the simple-cycle heat-rate regime that prevails for the smaller facilities; combined-cycle plants, selected only in high-gas price locations, sit off-trend at large fuel-price deviations, reflecting their lower heat rate and flatter elasticity.}
\label{fig:ng_elasticity}
\end{figure}
\clearpage

\begin{figure}[t]
\centering
\includegraphics[width=\linewidth]{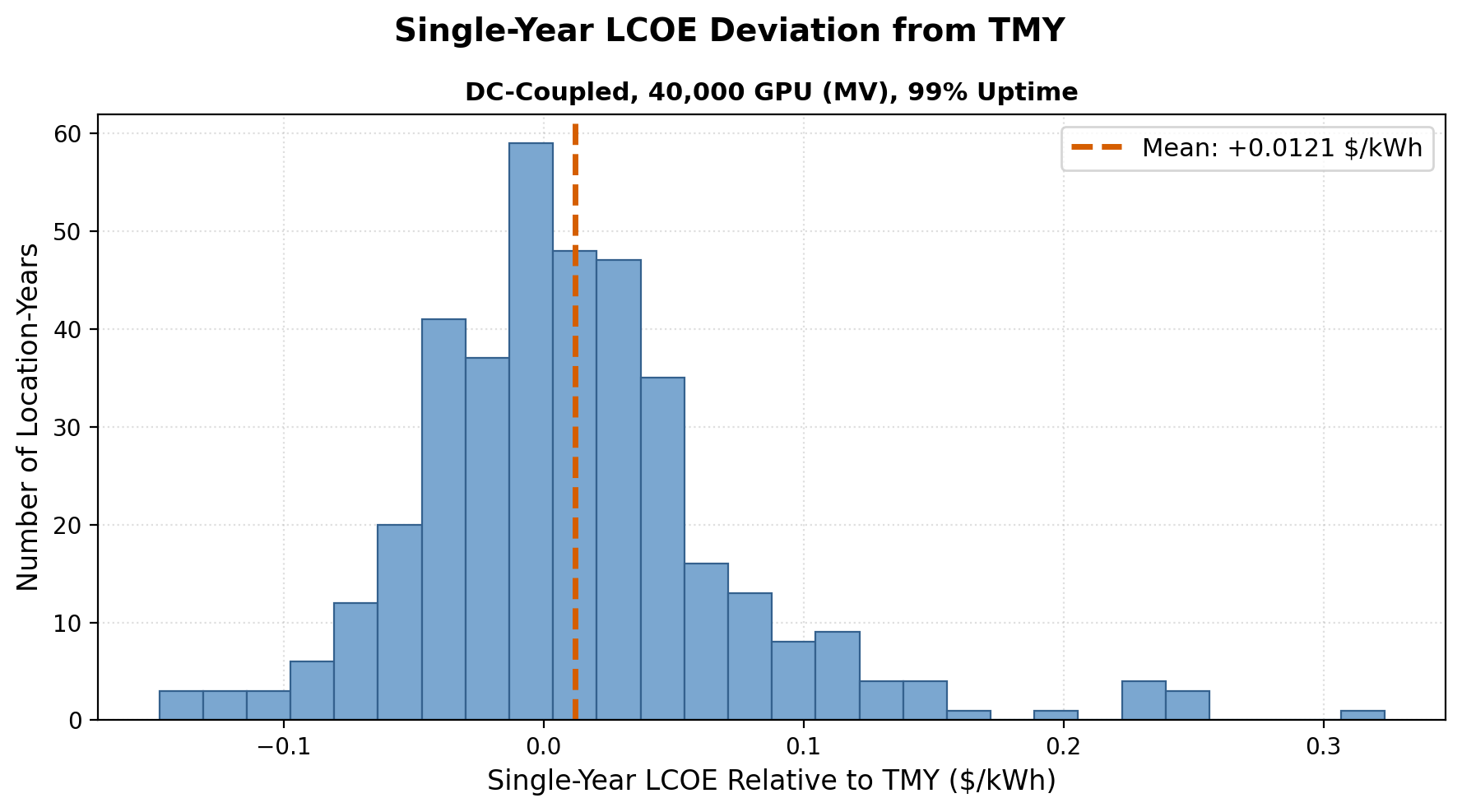}
\caption{\textbf{Distribution of single-year LCOE deviations from the TMY design basis.} Pooled across 28 locations (two per DOE climate zone) and the fourteen historical years 2010--2023, for the DC-coupled architecture at 40,000 GPUs and 99\% uptime (392 location-years). Each observation is one historical year's LCOE minus the TMY-based LCOE at the same site. The distribution is right-skewed---most years cluster near the TMY value while a thin tail of adverse years exceeds $+\$0.2$/kWh---pulling the mean to $+\$0.0121$/kWh (dashed line), the basis-wide optimism of TMY.}
\label{fig:tmy_hist}
\end{figure}
\clearpage

\begin{figure}[t]
\centering
\includegraphics[width=\linewidth]{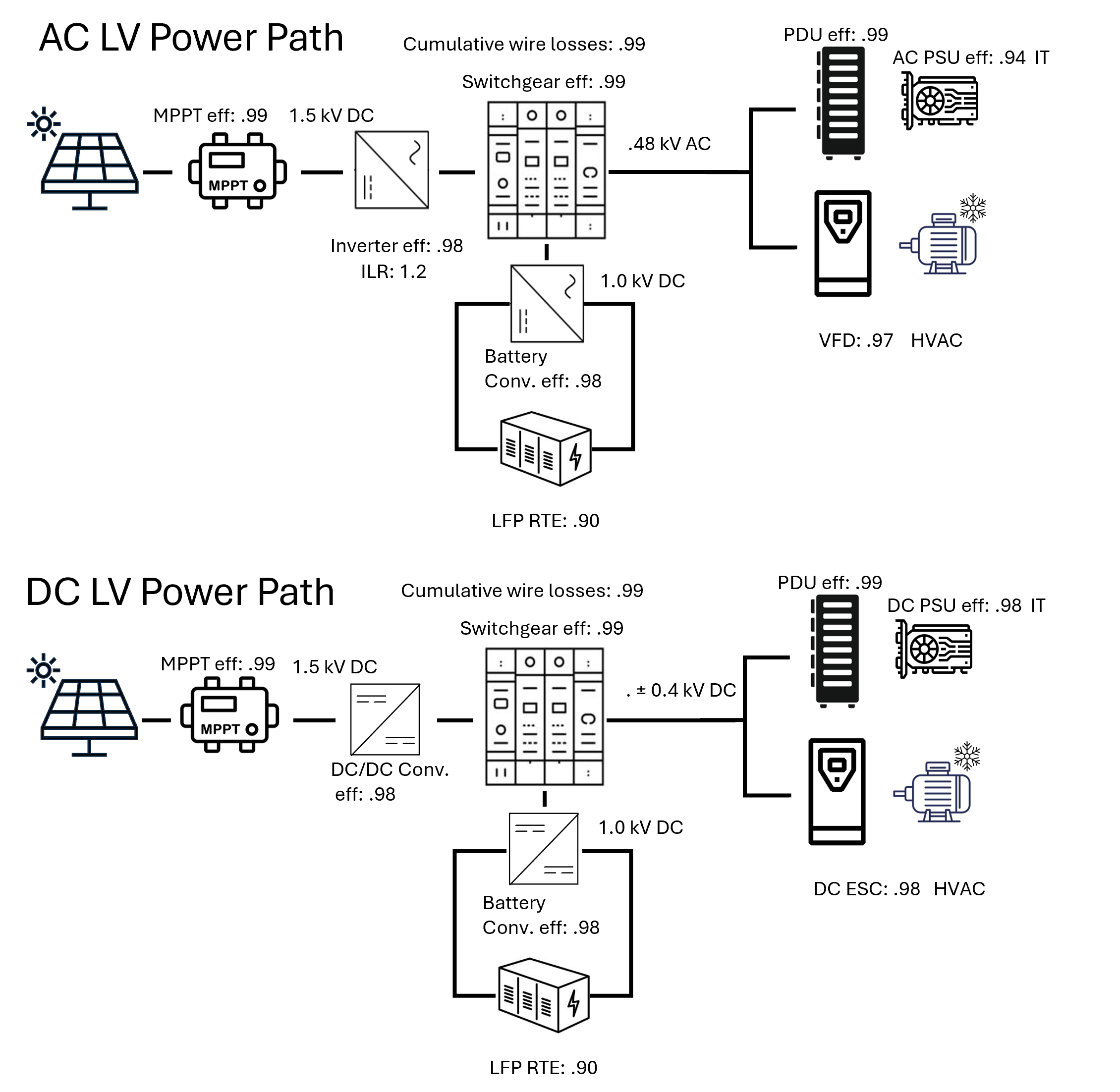}
\caption{\textbf{Modular Low Voltage PV+BESS Power Flow Architectures.} Candidate low voltage power topologies. Power flows are idealized, without batching or correspondence to any particular facility scale. Note that in the AC case, grid forming and power quality services are provided by the inverter. For cases with fully conditioned power, there would be a double conversion storage system embedded between the switchgear and the loads, analogous to the UPS placement in maintext figure 6.}
\label{fig:lv_powerflow}
\end{figure}
\clearpage

\begin{table}[t!]
\centering
\caption{Base LCOE sensitivity analysis across power system architectures (excluding GPU idling costs). Percent changes are relative to each architecture’s baseline LCOE (shown above result columns). Solar+storage and natural gas results are shown for the 100,000-GPU facility (solar+storage MV-coupled); the per-row percentage sensitivities of solar+storage are nearly scale- and topology-invariant, differing by less than one percentage point at the 24,000-GPU LV-coupled scale. Baseline assumptions: Solar construction = 2.0 yr; NG construction = 4.0 yr; discount rate = 7\%; NG price = \$7.70/MMBtu.}
\label{tbl:base_lcoe_sensitivity}
\begin{tabular}{lcccc}
\toprule
Parameter & Low & High & $\Delta\%$ Low & $\Delta\%$ High \\
          & (\$/kWh) & (\$/kWh) & (\%) & (\%) \\
\midrule
\multicolumn{5}{c}{\textbf{AC-Coupled Solar+Storage}} \\ 
\multicolumn{5}{c}{\textit{Baseline LCOE: \$0.3743/kWh}} \\
\midrule
Discount Rate (3--12\%) & 0.2685 & 0.5339 & -28.26 & +42.61 \\
Solar BOS (×0.8--1.2) & 0.3448 & 0.4039 & -7.89 & +7.89 \\
Battery Cost Y0 (×0.8--1.2) & 0.3581 & 0.3906 & -4.33 & +4.33 \\
Solar Module (×0.8--1.2) & 0.3604 & 0.3883 & -3.72 & +3.72 \\
Construction Time (1.5--3.0 yr) & 0.3757 & 0.3933 & +0.35 & +5.05 \\
Solar Fixed O\&M (×0.8--1.2) & 0.3678 & 0.3809 & -1.74 & +1.74 \\
Battery BOS (×0.8--1.2) & 0.3705 & 0.3782 & -1.03 & +1.03 \\
Battery Cost Y15 (×0.8--1.2) & 0.3706 & 0.3781 & -1.01 & +1.01 \\
Storage Fixed O\&M (×0.8--1.2) & 0.3734 & 0.3753 & -0.26 & +0.26 \\
Land Cost (×0.5--2.0) & 0.3741 & 0.3748 & -0.07 & +0.13 \\
\midrule
\multicolumn{5}{c}{\textbf{DC-Coupled Solar+Storage}} \\ 
\multicolumn{5}{c}{\textit{Baseline LCOE: \$0.2880/kWh}} \\
\midrule
Discount Rate (3--12\%) & 0.2100 & 0.4056 & -27.08 & +40.82 \\
Battery Cost Y0 (×0.8--1.2) & 0.2691 & 0.3070 & -6.57 & +6.57 \\
Solar BOS (×0.8--1.2) & 0.2736 & 0.3024 & -5.00 & +5.00 \\
Solar Module (×0.8--1.2) & 0.2784 & 0.2977 & -3.36 & +3.36 \\
Construction Time (1.5--3.0 yr) & 0.2896 & 0.3023 & +0.53 & +4.97 \\
Solar Fixed O\&M (×0.8--1.2) & 0.2835 & 0.2926 & -1.57 & +1.57 \\
Battery BOS (×0.8--1.2) & 0.2836 & 0.2925 & -1.56 & +1.56 \\
Battery Cost Y15 (×0.8--1.2) & 0.2836 & 0.2924 & -1.52 & +1.52 \\
Storage Fixed O\&M (×0.8--1.2) & 0.2869 & 0.2892 & -0.39 & +0.39 \\
Land Cost (×0.5--2.0) & 0.2878 & 0.2884 & -0.06 & +0.13 \\
\midrule
\multicolumn{5}{c}{\textbf{Natural Gas (100k GPU)}} \\ 
\multicolumn{5}{c}{\textit{Baseline LCOE: \$0.1729/kWh}} \\
\midrule
Natural Gas Price (\$3.01--12.75/MMBtu) & 0.1322 & 0.2168 & -23.53 & +25.34 \\
Discount Rate (3--12\%) & 0.1428 & 0.2245 & -17.45 & +29.81 \\
Construction Timeline (2.0--4.8 yr) & 0.1623 & 0.1768 & -6.13 & +2.23 \\
NG CapEx (×0.8--1.2) & 0.1599 & 0.1860 & -7.55 & +7.55 \\
NG Fixed O\&M (×0.8--1.2) & 0.1705 & 0.1754 & -1.39 & +1.39 \\
NG Variable O\&M (×0.8--1.2) & 0.1724 & 0.1735 & -0.29 & +0.29 \\
\bottomrule
\end{tabular}
\end{table}

\clearpage

\begin{table}[t!]
\centering
\caption{$\mathrm{LCOE_{adj}}$ sensitivity analysis. Percent changes are relative to each architecture’s baseline (shown above result columns). All results are shown for the 100,000-GPU facility (solar+storage MV-coupled); solar+storage percentage sensitivities are nearly scale- and topology-invariant. Baseline assumptions: Solar construction = 2.0 yr; NG construction = 4.0 yr; discount rate = 7\%; NG price = \$7.70/MMBtu; GPU spot price = \$2.40/chip-hr.}
\label{tbl:total_lcoe_sensitivity}
\begin{tabular}{lcccc}
\toprule
Parameter & Low & High & $\Delta\%$ Low & $\Delta\%$ High \\
          & (\$/kWh) & (\$/kWh) & (\%) & (\%) \\
\midrule
\multicolumn{5}{c}{\textbf{AC-Coupled Solar+Storage}} \\ 
\multicolumn{5}{c}{\textit{Baseline $\mathrm{LCOE_{adj}}$: \$0.8188/kWh}} \\
\midrule
Discount Rate (3--12\%) & 0.5548 & 1.2256 & -32.24 & +49.69 \\
GPU Idling (\$1--4/chip-hr) & 0.5595 & 1.1151 & -31.66 & +36.19 \\
Construction Time (1.5--3.0 yr) & 0.7128 & 1.0943 & -12.95 & +33.65 \\
Solar BOS (×0.8--1.2) & 0.7893 & 0.8483 & -3.61 & +3.61 \\
Battery Cost Y0 (×0.8--1.2) & 0.8026 & 0.8350 & -1.98 & +1.98 \\
Solar Module (×0.8--1.2) & 0.8049 & 0.8327 & -1.70 & +1.70 \\
Solar Fixed O\&M (×0.8--1.2) & 0.8123 & 0.8253 & -0.80 & +0.80 \\
Battery BOS (×0.8--1.2) & 0.8150 & 0.8226 & -0.47 & +0.47 \\
Battery Cost Y15 (×0.8--1.2) & 0.8150 & 0.8226 & -0.46 & +0.46 \\
Storage Fixed O\&M (×0.8--1.2) & 0.8178 & 0.8198 & -0.12 & +0.12 \\
Land Cost (×0.5--2.0) & 0.8185 & 0.8193 & -0.03 & +0.06 \\
\midrule
\multicolumn{5}{c}{\textbf{DC-Coupled Solar+Storage}} \\ 
\multicolumn{5}{c}{\textit{Baseline $\mathrm{LCOE_{adj}}$: \$0.7310/kWh}} \\
\midrule
GPU Idling (\$1--4/chip-hr) & 0.4726 & 1.0264 & -35.35 & +40.40 \\
Discount Rate (3--12\%) & 0.4953 & 1.0952 & -32.24 & +49.81 \\ 
Construction Time (1.5--3.0 yr) & 0.6256 & 1.0012 & -14.43 & +36.96 \\
Battery Cost Y0 (×0.8--1.2)  & 0.7121 & 0.7499 & -2.59 & +2.59 \\
Solar BOS (×0.8--1.2) & 0.7166 & 0.7454 & -1.97 & +1.97 \\
Solar Module (×0.8--1.2) & 0.7214 & 0.7407 & -1.32 & +1.32 \\
Solar Fixed O\&M (×0.8--1.2) & 0.7265 & 0.7355 & -0.62 & +0.62 \\
Battery BOS (×0.8--1.2) & 0.7265 & 0.7355 & -0.61 & +0.61 \\
Battery Cost Y15 (×0.8--1.2) & 0.7266 & 0.7354 & -0.60 & +0.60 \\
Storage Fixed O\&M (×0.8--1.2) & 0.7299 & 0.7322 & -0.16 & +0.16 \\
Land Cost (×0.5--2.0) & 0.7308 & 0.7314 & -0.03 & +0.05 \\
\midrule
\multicolumn{5}{c}{\textbf{Natural Gas}} \\ 
\multicolumn{5}{c}{\textit{Baseline $\mathrm{LCOE_{adj}}$: \$1.2603/kWh}} \\
\midrule
GPU Idling (\$1--4/chip-hr) & 0.6260 & 1.9852 & -50.33 & +57.52 \\
Discount Rate (3--12\%) & 0.8143 & 2.0390 & -35.39 & +61.79 \\
Construction Timeline (2.0--4.8 yr) & 0.6505 & 1.5414 & -48.38 & +22.30 \\
Natural Gas Price (\$3.01--12.75/MMBtu) & 1.2196 & 1.3041 & -3.23 & +3.48 \\
NG CapEx (×0.8--1.2) & 1.2472 & 1.2733 & -1.04 & +1.04 \\
NG Fixed O\&M (×0.8--1.2) & 1.2579 & 1.2627 & -0.19 & +0.19 \\
NG Variable O\&M (×0.8--1.2) & 1.2598 & 1.2608 & -0.04 & +0.04 \\
\bottomrule
\end{tabular}
\end{table}
\clearpage

\begin{table}[ht]
\centering
\caption{Technical parameter sensitivity across power system architectures for the 100,000-GPU facility (solar+storage MV-coupled; natural gas F-class) in southeastern Colorado. Values show the LCOE impact range (absolute \$/kWh and percentage change) for each parameter variation, computed by re-optimization under the perturbed parameter using common RNG seed. Sensitivity at 24,000 GPUs (aeroderivative) and 500,000 GPUs (H-class) similar in magnitude. The replacement-year row re-optimizes with the degradation schedule re-anchored to the replacement year: a later replacement (year 16) requires modestly more initial storage to hold reliability at the more-degraded pre-replacement point, and battery sizing moves by a few percent across the range. Parameters that do not bind the energy-limited solar design—inverter load ratio, design contingency, and the double-conversion UPS stage return the identical re-optimized design and hence no first-order LCOE effect. Inverter load ratio matters only in the rare hours when clipped DC output coincides with excess bus demand, which the large power overbuild renders negligible.}
\label{tab:technical_sensitivity}
\small
\begin{tabular}{@{}p{3.5cm}rrr@{}}
\toprule
\textbf{Parameter (Range)} & \textbf{AC Solar} & \textbf{DC Solar} & \textbf{Natural Gas} \\
\midrule
& \textit{Baseline: \$0.3743/kWh} & \textit{Baseline: \$0.2880/kWh} & \textit{Baseline: \$0.1729/kWh} \\
\midrule
\multicolumn{4}{@{}l}{\textbf{Battery Performance}} \\
\midrule
Battery RTE & $-\$0.0093$ to $+\$0.0119$ & $-\$0.0075$ to $+\$0.0073$ & --- \\
(0.85 -- 0.95) & $-2.48\%$ to $+3.19\%$ & $-2.61\%$ to $+2.54\%$ \\
\addlinespace
Replacement Year & $-\$0.0023$ to $+\$0.0191$ & $-\$0.0033$ to $+\$0.0100$ & --- \\
(Year 16 -- Year 10) & $-0.61\%$ to $+5.10\%$ & $-1.14\%$ to $+3.48\%$ \\
\midrule
\multicolumn{4}{@{}l}{\textbf{Power Electronics}} \\
\midrule
PSU Efficiency & $-\$0.0074$ to $+\$0.0102$ & $-\$0.0032$ to $+\$0.0053$ & $-\$0.0019$ to $+\$0.0016$ \\
(AC: 0.92--0.96; DC: 0.96--0.99) & $-1.98\%$ to $+2.73\%$ & $-1.12\%$ to $+1.85\%$ & $-1.08\%$ to $+0.93\%$ \\
\addlinespace
UPS Efficiency & --- & --- & $-\$0.0020$ to $+\$0.0017$ \\
(0.92 -- 0.96) & & & $-1.15\%$ to $+1.01\%$ \\
\addlinespace
Inverter Efficiency & $-\$0.0018$ to $+\$0.0020$ & --- & --- \\
(0.94 -- 0.98) & $-0.47\%$ to $+0.52\%$ & & \\
\addlinespace
Inverter Load Ratio & $\sim\$0$ & --- & --- \\
(1.1 -- 1.4) & $<0.05\%$ & & \\
\midrule
\multicolumn{4}{@{}l}{\textbf{Turbine Performance}} \\
\midrule
Part-Load Efficiency & --- & --- & $-\$0.0019$ to $+\$0.0020$ \\
(F-class: 0.10--0.20) & & & $-1.09\%$ to $+1.15\%$ \\
\addlinespace
Temperature Derating & --- & --- & $+\$0.00$ to $+\$0.00$ \\
(F-class: 0.004--0.012 /°C) & & &  \\
\midrule
\multicolumn{4}{@{}l}{\textbf{Design Parameters}} \\
\midrule
Design Contingency & $\sim\$0$ & $\sim\$0$ & $\sim\$0$ \\
(1.03 -- 1.10) & $\approx 0\%$ & $\approx 0\%$ & $\approx 0\%$ \\
\bottomrule
\end{tabular}
\end{table}

\begin{table*}[t]
\centering
\caption{Cost levers in PV+BESS topology, ordered by magnitude. Penalties relative to base LCOE modeled for an illustrative high-resource site (SE Colorado, 100k GPUs, 99\% uptime).}
\label{tab:topology_sensitivity}
\scriptsize
\begin{tabular}{@{}llll@{}}
\toprule
\textbf{Decision} & \textbf{Preferred Siting} & \textbf{Driver for Costlier Choice} & \textbf{Penalty ($\Delta$ LCOE)} \\
\midrule
\multicolumn{4}{@{}l}{\textit{First-Order Levers}} \\
\midrule
Full IT Load Conditioning & BESS for transients & Imposed PQ / full-conditioning mandate & $+5.9\%$ ($+6.4\%$ w/ cooling) \\
\midrule
\multicolumn{4}{@{}l}{\textit{Second-Order Levers (Distribution \& Storage Siting)}} \\
\midrule
Distribution Voltage & LV modular (single bus) & Scaling $>2$~MW (ampacity, O\&M complexity) & AC: $+3.7\%$ / DC: $+2.4\%$ \\
\addlinespace
AC Storage Siting & Behind shared PV inverter & Gathering ceiling / local hall services & $\approx +1\%$ \\
\addlinespace
DC Storage Siting & Data hall (short discharge) & LV-bus ampacity constraints & $<1\%$ \\
\bottomrule
\end{tabular}
\end{table*}

\begin{table}[t!]
\centering
\caption{Natural Gas baseline $\mathrm{LCOE}$ parameter sensitivity for the 24,000 and 500,000 GPU cases Modeled facilities are in southeastern Colorado. Percent changes are relative to each scale's baseline. Baseline assumptions: Discount rate = 7\%; NG price = \$7.70/MMBtu.}
\label{tbl:ng_lcoe_sensitivity_baseline}
\begin{tabular}{lcccc}
\toprule
Parameter & Low & High & $\Delta\%$ Low & $\Delta\%$ High \\
          & (\$/kWh) & (\$/kWh) & (\%) & (\%) \\
\midrule
\multicolumn{5}{c}{\textbf{24,000 GPUs }} \\ 
\multicolumn{5}{c}{\textit{Baseline $\mathrm{LCOE}$: \$0.2082/kWh}} \\
\midrule
Natural Gas Price (\$3.01--12.75/MMBtu) & 0.1549 & 0.2657 & -25.60 & +27.57 \\
Discount Rate (3--12\%) & 0.1790 & 0.2557 & -14.03 & +22.78 \\
NG CapEx (×0.8--1.2) & 0.1967 & 0.2198 & -5.56 & +5.56 \\
Turbine Lead Time (×0.7--1.3) & 0.2097 & 0.2220 & +0.70 & +6.59 \\
NG Construction Time (×0.5--1.2) & 0.1997 & 0.2199 & -4.12 & +5.61 \\
NG Fixed O\&M (×0.8--1.2) & 0.2060 & 0.2105 & -1.09 & +1.09 \\
NG Variable O\&M (×0.8--1.2) & 0.2063 & 0.2102 & -0.93 & +0.93 \\
\midrule
\multicolumn{5}{c}{\textbf{500,000 GPUs }} \\ 
\multicolumn{5}{c}{\textit{Baseline $\mathrm{LCOE}$: \$0.1430/kWh}} \\
\midrule
Natural Gas Price (\$3.01--12.75/MMBtu) & 0.1043 & 0.1846 & -27.02 & +29.10 \\
Discount Rate (3--12\%) & 0.1216 & 0.1796 & -14.98 & +25.62 \\
NG Construction Time (×0.5--1.2) & 0.1355 & 0.1458 & -5.26 & +1.94 \\
NG CapEx (×0.8--1.2) & 0.1346 & 0.1514 & -5.87 & +5.87 \\
Turbine Lead Time (×0.7--1.3) & 0.1392 & 0.1470 & -2.67 & +2.84 \\
NG Fixed O\&M (×0.8--1.2) & 0.1411 & 0.1448 & -1.30 & +1.30 \\
NG Variable O\&M (×0.8--1.2) & 0.1425 & 0.1435 & -0.36 & +0.36 \\
\bottomrule
\end{tabular}
\end{table}

\begin{table}[t!]
\centering
\caption{Natural Gas $\mathrm{LCOE_{adj}}$ parameter sensitivity for the 24,000 and 500,000 GPU cases. Modeled facilities are in southeastern Colorado. Percent changes are relative to each scale's baseline. Baseline assumptions: Discount rate = 7\%; NG price = \$7.70/MMBtu.}
\label{tbl:ng_lcoe_sensitivity_scale}
\begin{tabular}{lcccc}
\toprule
Parameter & Low & High & $\Delta\%$ Low & $\Delta\%$ High \\
          & (\$/kWh) & (\$/kWh) & (\%) & (\%) \\
\midrule
\multicolumn{5}{c}{\textbf{24,000 GPUs}} \\ 
\multicolumn{5}{c}{\textit{Baseline $\mathrm{LCOE_{adj}}$: \$1.1909/kWh}} \\
\midrule
GPU Idling (\$1--4/chip-hr) & 0.6177 & 1.8460 & -48.13 & +55.01 \\
Discount Rate (3--12\%) & 0.8018 & 1.8361 & -32.67 & +54.18 \\
NG Construction Time (×0.5--1.2) & 0.6428 & 1.5626 & -46.02 & +31.21 \\
Turbine Lead Time (×0.7--1.3) & 0.9112 & 1.7104 & -23.48 & +43.62 \\
Natural Gas Price (\$3.01--12.75/MMBtu) & 1.1376 & 1.2483 & -4.48 & +4.82 \\
NG CapEx (×0.8--1.2) & 1.1793 & 1.2025 & -0.97 & +0.97 \\
NG Fixed O\&M (×0.8--1.2) & 1.1886 & 1.1931 & -0.19 & +0.19 \\
NG Variable O\&M (×0.8--1.2) & 1.1889 & 1.1928 & -0.16 & +0.16 \\
\midrule
\multicolumn{5}{c}{\textbf{500,000 GPUs}} \\ 
\multicolumn{5}{c}{\textit{Baseline $\mathrm{LCOE_{adj}}$: \$1.2303/kWh}} \\
\midrule
GPU Idling (\$1--4/chip-hr) & 0.5960 & 1.9552 & -51.55 & +58.92 \\
Discount Rate (3--12\%) & 0.7931 & 1.9941 & -35.54 & +62.08 \\
NG Construction Time (×0.5--1.2) & 0.6237 & 1.5103 & -49.31 & +22.76 \\
Turbine Lead Time (×0.7--1.3) & 0.8510 & 1.6593 & -30.83 & +34.87 \\
Natural Gas Price (\$3.01--12.75/MMBtu) & 1.1917 & 1.2719 & -3.14 & +3.38 \\
NG CapEx (×0.8--1.2) & 1.2219 & 1.2387 & -0.68 & +0.68 \\
NG Fixed O\&M (×0.8--1.2) & 1.2285 & 1.2322 & -0.15 & +0.15 \\
NG Variable O\&M (×0.8--1.2) & 1.2298 & 1.2308 & -0.04 & +0.04 \\
\bottomrule
\end{tabular}
\end{table}

\begin{table}[t!]
\centering
\caption{Economic, Hardware, and Solar PV System Parameters}
\label{tbl:economic_hardware_solar}
\small
\begin{tabular}{lll}
\toprule
\textbf{Parameter} & \textbf{Value} & \textbf{Source} \\
\midrule
\multicolumn{3}{c}{\textbf{COMPUTE HARDWARE}} \\
\midrule
GPUs per node & 8 & DGX H100 standard \cite{nvidia_nvidia_nodate} \\
Node power (average) & 7.3 kW & \cite{newkirk_empirically-calibrated_2025} \\
Node power (maximum) & 8.5 kW & \cite{newkirk_empirically-calibrated_2025} \\
Design contingency & 1.05 & 5\% safety margin \\
\midrule
\multicolumn{3}{c}{\textbf{ECONOMIC ASSUMPTIONS}} \\
\midrule
GPU spot price & \$2.40/GPU-hr & \cite{bloomberg_lp_sdh100_2025} \\
Discount rate & 7\% & Base case \\
Evaluation period & conclusion of year 27 & Solar EOL \\
\midrule
\multicolumn{3}{c}{\textbf{SOLAR PV SYSTEM}} \\
\midrule
\multicolumn{3}{l}{\textit{Capital Costs (Year 0)}} \\
Solar modules & \$402/kW$_{\text{DC}}$ & NLR ATB 2024 \cite{mirletz_annual_2024} \\
Solar BOS (AC-coupled) & \$661.95/kW$_{\text{DC}}$ & \cite{ramasamy_us_2023} \\
Solar BOS (DC-coupled) & \$501.98/kW$_{\text{DC}}$ & DC coupled share of \cite{ramasamy_us_2023}  \\
Land & \$150,000/km$^2$ & Estimated average costs \\
\multicolumn{3}{l}{\textit{Operations \& Maintenance}} \\
Solar fixed O\&M & \$16.58/kW$_{\text{DC}}$-yr & \cite{ramasamy_us_2023} \\
Land intensity & 4.17 acres/MW$_{\text{DC}}$ & \cite{ong_land-use_2013, bolinger_land_2022} \\
\multicolumn{3}{l}{\textit{Degradation}} \\
First year loss & 1.0\%/yr & \cite{yang_comprehensive_2025} \\
Subsequent years & 0.55\%/yr & \cite{yang_comprehensive_2025} \\
\midrule
\multicolumn{3}{c}{\textbf{CONSTRUCTION TIMELINES}} \\
\midrule
Solar + storage & 24 months & \cite{baranko_fast_2024} \\
NG simple cycle & 36 months & \cite{larose_capital_2024} \\
NG combined cycle & 42 months & \cite{larose_capital_2024} \\
Turbine procurement & 36 months & OEM estimate \\
Grid interconnection & State-specific & \cite{rand_queued_2024} \\
\bottomrule
\end{tabular}
\end{table}

\begin{table}[t!]
\centering
\caption{Battery Storage System Parameters}
\label{tbl:battery_conversion}
\small
\begin{tabular}{lll}
\toprule
\textbf{Parameter} & \textbf{Value} & \textbf{Source} \\
\midrule
\multicolumn{3}{c}{\textbf{BATTERY STORAGE SYSTEM}} \\
\midrule
\multicolumn{3}{l}{\textit{Capital Costs}} \\
Battery cells (Year 0) & \$826/kW & \cite{mirletz_annual_2024}  \\
Battery cells (Year 13) & \$508.12/kW & \cite{mirletz_annual_2024} moderate case \\
Battery BOS (AC-coupled) & \$385.31/kW & \cite{ramasamy_us_2023} \\
Battery BOS (DC-coupled) & \$292.20/kW & Proportional reduction \\
\multicolumn{3}{l}{\textit{Operations \& Maintenance}} \\
Battery fixed O\&M & \$4.37/kW-yr & PNNL \cite{pacific_northwest_national_lab_esgc_2024} \\
\multicolumn{3}{l}{\textit{Performance \& Degradation}} \\
Battery duration & 4 hours & \cite{mirletz_annual_2024}   \\
Round-trip efficiency (RTE) & 90\% & \cite{rehm_comparing_2025} \\
Replacement year & 13 & Project mid life \\
$E_{\text{cal}}$ (calendar) & $-5.21$ kJ/mol & \cite{smith_lithium-ion_2021} \\
$E_{\text{cyc}}$ (cycling) & $+2.19$ kJ/mol & Inverse Arrhenius \cite{smith_lithium-ion_2021} \\
$\beta_0$ (calendar exponent) & 0.526 & BLAST-Lite distillation \cite{smith_lithium-ion_2021} \\
$\alpha_0$ (cycle exponent) & 0.828 & BLAST-Lite distillation \cite{smith_lithium-ion_2021} \\
\midrule

\end{tabular}
\end{table}

\begin{table}[t!]
\centering
\caption{Natural Gas, Diesel, and Gas Turbine Performance Parameters}
\label{tbl:gas_diesel}
\small
\begin{tabular}{lll}
\toprule
\textbf{Parameter} & \textbf{Value} & \textbf{Source} \\
\midrule
\multicolumn{3}{c}{\textbf{NATURAL GAS TURBINES}} \\
\midrule
\multicolumn{3}{l}{\textit{Simple Cycle Reference Plants}} \\
Aero SC (211 MW) CapEx & \$1,510/kW & EIA \cite{larose_capital_2024} \\
Aero SC Fixed O\&M & \$35/kW-yr & \cite{larose_capital_2024} \\
Aero SC Variable O\&M & \$8/MWh & \cite{larose_capital_2024} \\
F-class SC (233 MW) CapEx & \$1,319/kW & NLR ATB \cite{mirletz_annual_2024} \\
F-class SC Fixed O\&M & \$25.70/kW-yr & \cite{mirletz_annual_2024} \\
F-class SC Variable O\&M & \$6.94/MWh & \cite{mirletz_annual_2024} \\
\multicolumn{3}{l}{\textit{Combined Cycle Reference Plants}} \\
F-class CC (727 MW) CapEx & \$1,455/kW & \cite{mirletz_annual_2024} \\
F-class CC Fixed O\&M & \$33/kW-yr & \cite{mirletz_annual_2024} \\
F-class CC Variable O\&M & \$2.10/MWh & \cite{mirletz_annual_2024}\\
H-class CC (800 MW) CapEx & \$1,600/kW & \cite{mirletz_annual_2024} \\
H-class CC Fixed O\&M & \$36/kW-yr & \cite{mirletz_annual_2024} \\
H-class CC Variable O\&M & \$2.10/MWh & \cite{mirletz_annual_2024} \\
\multicolumn{3}{l}{\textit{Cost Scaling}} \\
CapEx exponent & 0.67 & \cite{tribe_scale_1986, james_quality_2022} \\
Fixed O\&M exponent & 0.80 & \cite{larose_capital_2024} \\

\midrule
\multicolumn{3}{c}{\textbf{TURBINE DEGRADATION \& MAINTENANCE}} \\
\midrule
\multicolumn{3}{l}{\textit{Annual Degradation}} \\
Aero capacity loss & 0.20\% & \cite{kurz_degradation_2000} \\
Aero efficiency loss & 0.15\% & \cite{kurz_degradation_2000}\\
F-class capacity loss & 0.15\% & \cite{kurz_degradation_2000} \\
F-class efficiency loss & 0.13\% & \cite{kurz_degradation_2000} \\
H-class capacity loss & 0.13\% & \cite{kurz_degradation_2000} \\
H-class efficiency loss & 0.10\% & \cite{kurz_degradation_2000} \\
\multicolumn{3}{l}{\textit{Part-Load Efficiency Multiplier}} \\
Aero & $1.0 - 0.10(1-\beta)$ & \cite{munoz_de_escalona_part-load_2012} \\
F-class & $1.0 - 0.15(1-\beta)$ & \cite{munoz_de_escalona_part-load_2012} \\
H-class & $1.0 - 0.20(1-\beta)$ & \cite{munoz_de_escalona_part-load_2012}\\
\multicolumn{3}{l}{\textit{Temperature Derating (per °C above 15°C)}} \\
Aero & 1.0\% & \cite{de_sa_gas_2011} \\
F-class & 0.8\% & \cite{de_sa_gas_2011} \\
H-class & 0.7\% & \cite{de_sa_gas_2011} \\
\midrule
\multicolumn{3}{c}{\textbf{DIESEL BACKUP SYSTEM}} \\
\midrule
Genset capital cost & \$800/kW & \cite{ericson_comparison_2019} \\
Fuel storage cost & \$5/gallon & \cite{mott_best_2021} \\
Diesel fuel cost & \$3.40/gallon & EIA \cite{noauthor_gasoline_2025} \\
Diesel specific fuel consumption & 13.3 gal/MWh & Manufacturer data avg. \\
Annual testing requirement & 24 hr/yr & Monthly loaded tests \\
\bottomrule
\end{tabular}
\end{table}

\begin{table}[t!]
\centering
\caption{Gas Turbine Models Available for System Configuration. Aeroderivatives optimized for simple cycle (SC); F-class suitable for both SC and combined cycle (CC); H-class preferred for CC configurations.}
\label{tbl:turbine_library}
\small
\begin{tabular}{lccc}
\toprule
\textbf{Turbine Model} & \textbf{Capacity (MW)} & \textbf{Efficiency} & \textbf{Availability} \\
\midrule
\multicolumn{4}{c}{\textit{Aeroderivatives (Simple Cycle)}} \\
\midrule
Siemens SGT-A05 KB7S & 5.8 & 0.323 & 0.92 \\
GE LM2500+G4 & 35.0 & 0.390 & 0.95 \\
Siemens SGT-A45 & 44.0 & 0.404 & 0.94 \\
GE LM6000PC & 51.0 & 0.397 & 0.94 \\
Siemens SGT-800 & 62.0 & 0.411 & 0.92 \\
P\&W FT4000 SwiftPac 70 & 71.0 & 0.410 & 0.95 \\
GE LMS100 & 100.0 & 0.395 & 0.93 \\
\midrule
\multicolumn{4}{c}{\textit{F-Class Industrial (Simple \& Combined Cycle)}} \\
\midrule
GE 6F.03 & 88.0 & 0.368 & 0.93 \\
Mitsubishi M501F & 185.4 & 0.370 & 0.91 \\
GE 7F.04 & 202.0 & 0.375 & 0.93 \\
Siemens SGT6-5000F & 260.0 & 0.400 & 0.93 \\
\midrule
\multicolumn{4}{c}{\textit{H-Class (Combined Cycle Preferred)}} \\
\midrule
GE 7HA.01 & 290.0 & 0.420 & 0.92 \\
Siemens SGT-8000H & 310.0 & 0.404 & 0.91 \\
GE 7HA.02 & 384.0 & 0.426 & 0.91 \\
GE 9HA.01 & 448.0 & 0.429 & 0.91 \\
GE 9HA.02 & 571.0 & 0.440 & 0.91 \\
\bottomrule
\end{tabular}
\end{table}

\begin{table}[h]
\centering
\caption{State-Level Grid and Fuel Parameters (2022 USD). Electricity and gas prices from EIA 2022 industrial rates \cite{noauthor_electric_2025, noauthor_natural_2025}. Interconnection queue data from LBNL 2018--2023 median \cite{rand_queued_2024}.}
\label{tab:state_params}
\scriptsize
\begin{tabular}{@{}llcccp{3.5cm}@{}}
\toprule
\textbf{State} & \textbf{ISO/RTO} & \textbf{Queue (yr)} & \textbf{Elec (¢/kWh)} & \textbf{Gas (\$/MMBtu)} & \textbf{Region} \\
\midrule
Connecticut & ISO-NE & 2.5 & 15.20 & 7.87 & New England \\
Maine & ISO-NE & 2.5 & 12.01 & 9.47 & \\
Massachusetts & ISO-NE & 2.5 & 16.79 & 12.04 & \\
New Hampshire & ISO-NE & 2.5 & 15.39 & 12.02 & \\
Rhode Island & ISO-NE & 2.5 & 16.42 & 11.57 & \\
Vermont & ISO-NE & 2.5 & 11.02 & 5.16 & \\
\midrule
New Jersey & PJM & 3.75 & 11.03 & 8.35 & Mid-Atlantic \\
New York & NYISO & 4.5 & 7.82 & 9.43 & \\
Pennsylvania & PJM & 3.75 & 7.26 & 10.67 & \\
\midrule
Illinois & PJM/MISO & 3.75 & 7.88 & 6.79 & E N Central \\
Indiana & MISO & 3.75 & 7.49 & 6.17 & \\
Michigan & MISO & 3.75 & 7.52 & 7.70 & \\
Ohio & PJM & 3.75 & 6.39 & 7.02 & \\
Wisconsin & MISO & 3.75 & 7.53 & 6.22 & \\
\midrule
Iowa & MISO & 3.75 & 5.63 & 6.40 & W N Central \\
Kansas & SPP & 4.0 & 7.02 & 5.04 & \\
Minnesota & MISO & 3.75 & 8.46 & 5.60 & \\
Missouri & SPP/MISO & 4.0 & 7.10 & 8.69 & \\
Nebraska & SPP & 4.0 & 6.67 & 5.54 & \\
North Dakota & MISO/SPP & 4.0 & 6.98 & 2.98 & \\
South Dakota & SPP & 4.0 & 7.63 & 5.57 & \\
\midrule
Delaware & PJM & 3.75 & 7.92 & 12.43 & S Atlantic \\
Florida & Non-ISO & 3.0 & 8.09 & 6.50 & \\
Georgia & Non-ISO & 3.0 & 5.41 & 4.86 & \\
Maryland & PJM & 3.75 & 8.75 & 12.07 & \\
North Carolina & PJM/Non-ISO & 3.5 & 7.26 & 6.08 & \\
South Carolina & Non-ISO & 3.0 & 6.01 & 4.68 & \\
Virginia & PJM & 3.75 & 8.26 & 4.58 & \\
West Virginia & PJM & 3.75 & 7.19 & 3.85 & \\
\midrule
Alabama & Non-ISO & 3.0 & 6.73 & 4.11 & E S Central \\
Kentucky & PJM/MISO & 3.75 & 6.12 & 4.07 & \\
Mississippi & MISO & 3.75 & 6.38 & 5.04 & \\
Tennessee & Non-ISO/MISO & 3.5 & 5.99 & 4.99 & \\
\midrule
Arkansas & SPP/MISO & 4.0 & 5.85 & 8.57 & W S Central \\
Louisiana & MISO/SPP & 4.0 & 5.98 & 3.49 & \\
Oklahoma & SPP & 4.0 & 4.68 & 3.01 & \\
Texas & ERCOT & 3.0 & 5.48 & 3.24 & \\
\midrule
Arizona & Non-ISO & 4.0 & 6.84 & 6.29 & Mountain \\
Colorado & Non-ISO & 4.0 & 7.75 & 7.70 & \\
Idaho & Non-ISO & 4.0 & 6.15 & 6.03 & \\
Montana & Non-ISO & 4.0 & 6.67 & 7.14 & \\
Nevada & CAISO/Non-ISO & 4.5 & 7.45 & 10.64 & \\
New Mexico & SPP/Non-ISO & 4.0 & 5.39 & 4.58 & \\
Utah & Non-ISO & 4.0 & 6.11 & 9.06 & \\
Wyoming & Non-ISO & 4.0 & 7.26 & 5.92 & \\
\midrule
California & CAISO & 4.75 & 18.50 & 12.75 & Pacific \\
Oregon & Non-ISO & 4.0 & 7.18 & 7.23 & \\
Washington & Non-ISO & 4.0 & 5.87 & 10.32 & \\
Alaska & Non-ISO & 4.0 & 20.09 & 5.92 & \\
Hawaii & Non-ISO & 4.0 & 33.71 & 25.82 & \\
\midrule
US Average &  & 4.0 & 7.25 & 4.11 & National Average \\
\bottomrule
\end{tabular}
\end{table}

\FloatBarrier

\end{document}